\documentclass[10pt,twocolumn,letterpaper]{article}

\usepackage[pagenumbers]{cvpr} 

\definecolor{DarkRed}{rgb}{0.8, 0.0, 0.4}
\definecolor{DarkGreen}{rgb}{0.0, 0.6, 0.4}
\definecolor{MetaBlue}{rgb}{0.0, 0.51, 0.98}

\definecolor{cvprblue}{rgb}{0.21,0.49,0.74}
\usepackage[pagebackref,breaklinks,colorlinks,allcolors=cvprblue]{hyperref}

\def\paperID{11935} 
\def\confName{CVPR}
\def\confYear{2025}

\def\rvx{{\mathbf{x}}}

\usepackage{adjustbox}
\usepackage{multirow}
\usepackage{graphicx}
\usepackage{subcaption}
\usepackage{wrapfig}
\usepackage{colortbl}
\usepackage{float}
\usepackage{dblfloatfix}

\usepackage[linesnumbered,ruled,vlined]{algorithm2e}
\SetKwInput{KwInput}{Input}
\SetKwInput{KwOutput}{Output}

\title{Autoregressive Distillation of Diffusion Transformers}

\author{$\text{Yeongmin Kim}^{\dagger, \ddagger, *}$ \quad $\text{Sotiris Anagnostidis}^{\dagger, \S}$ \quad $\text{Yuming Du}^{\dagger}$ \quad $\text{Edgar Schönfeld}^{\dagger}$ \quad $\text{Jonas Kohler}^{\dagger}$ \quad\\ $\text{Markos Georgopoulos}^{\dagger}$  \quad $\text{Albert Pumarola}^{\dagger}$ \quad $\text{Ali Thabet}^{\dagger}$  \quad $\text{Artsiom Sanakoyeu}^{\dagger}$}

\newcommand\blfootnote[1]{%
  \begingroup
  \renewcommand\thefootnote{}\footnote{#1}%
  \addtocounter{footnote}{-1}%
  \endgroup
}

\usepackage[accsupp]{axessibility}  

\begin{document}

\twocolumn[{
    \maketitle
    \begin{center}
        \vspace{-8mm}
            \includegraphics[width=0.77\linewidth]{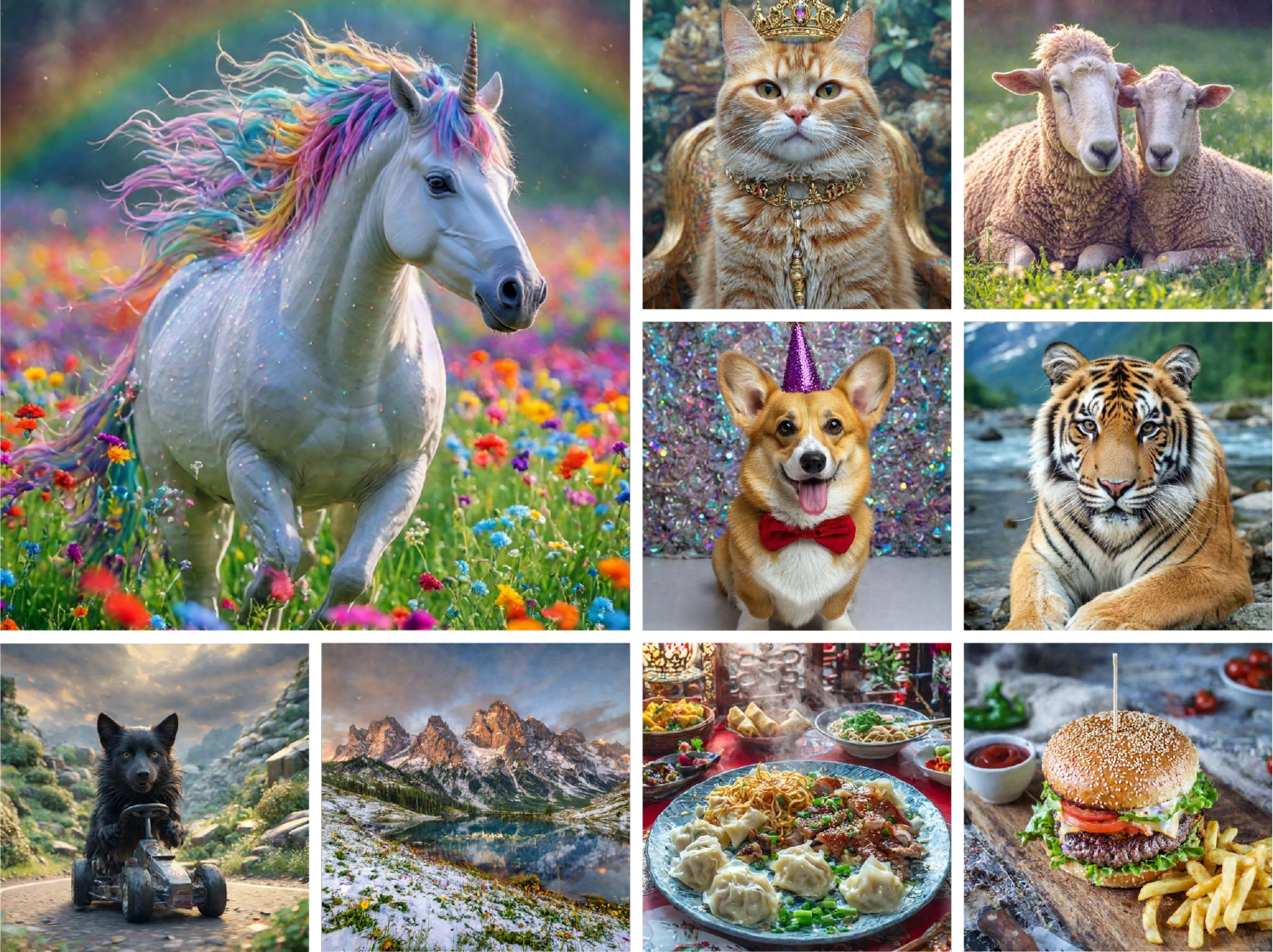}
            \captionof{figure}{Samples ($1024\times1024$) generated by our 3-step ARD model, distilled from a 1.7B Emu.}
            \label{fig:1}
            
    \end{center}
}]

\blfootnote{\thanks{${}^*$Work done during an internship at Meta GenAI. ${}^{\dagger}$Meta GenAI. ${}^{\S}$ETH Zürich. $\ddagger$KAIST. Correspondence to: \texttt{alsdudrla10@kaist.ac.kr}}}

\begin{abstract}
\begin{figure*}
     \begin{minipage}[h]{0.36\textwidth}
     \centering
         \begin{subfigure}{1.0\textwidth}
             \centering
             \includegraphics[width=\textwidth, height=0.60in]{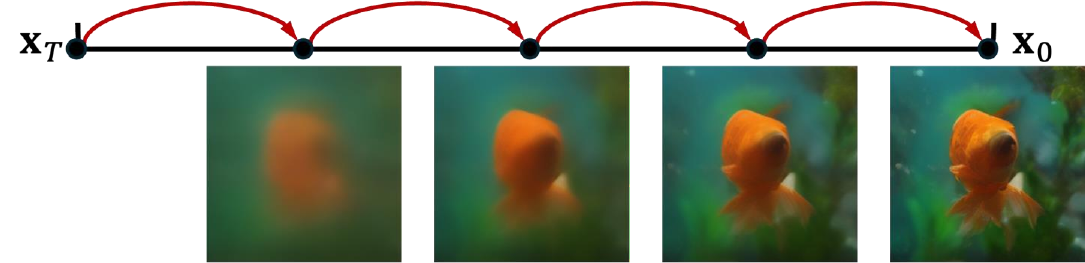}
             \caption{Step Distillation (baseline)}
             \label{fig:2_a}
         \end{subfigure}
         \begin{subfigure}{1.0\textwidth}
             \centering
             \includegraphics[width=\textwidth, height=0.70in]{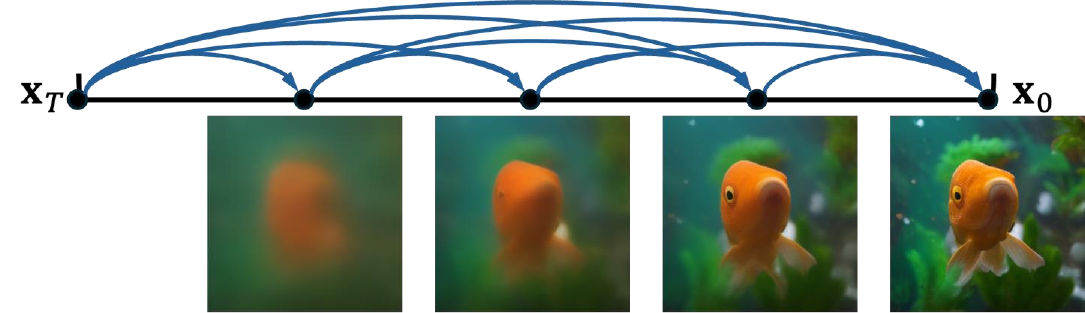}
             \caption{AutoRegressive Distillation (ours)} 
              \label{fig:2_b}
         \end{subfigure}     
    \end{minipage}
    \begin{minipage}[h]{0.32\textwidth}
        \begin{subfigure}{1.0\textwidth}
                 \centering
                 \includegraphics[width=\textwidth, height=1.5in]{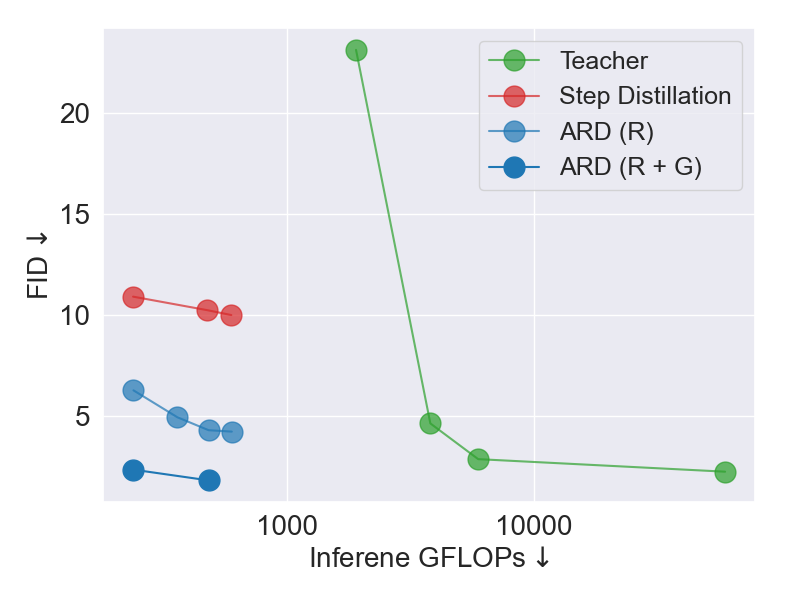}
                 \caption{A Comparison of distillation methods.} 
                  \label{fig:2_c}
             \end{subfigure}
    \end{minipage}
    \begin{minipage}[h]{0.32\textwidth}
        \begin{subfigure}{1.0\textwidth}
                 \centering
                 \includegraphics[width=\textwidth, height=1.5in]{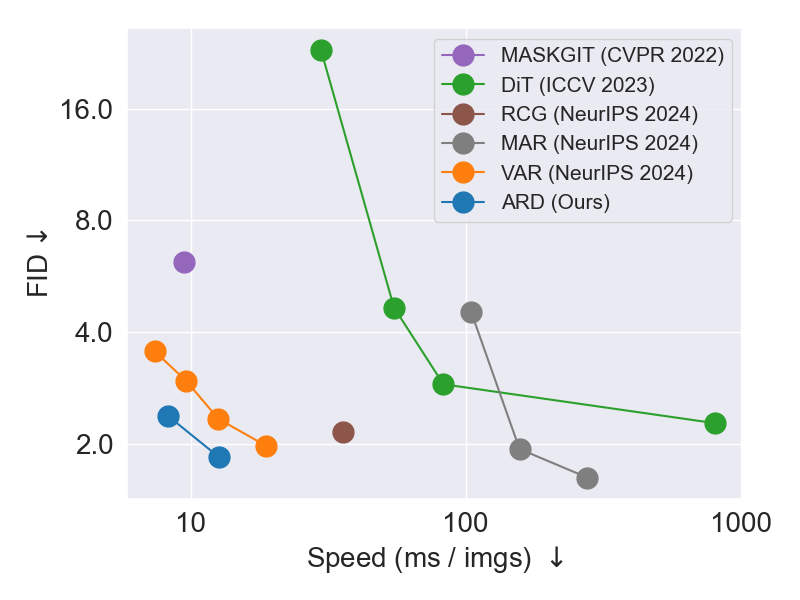}
                 \caption{A comparison of recent generative models.} 
                  \label{fig:2_d}
             \end{subfigure}
    \end{minipage}
     \vspace{-1.5mm}
    \caption{(a, b) Overall scheme of the baseline and proposed distillation methods. The training trajectory is given by the teacher ODE. (c, d) Comparison of the efficiency-performance trade-offs of the distillation methods and public generative models on ImageNet 256p.}
    \label{fig:2}
     \vspace{-3.5mm}
\end{figure*}
Diffusion models with transformer architectures have demonstrated promising capabilities in generating high-fidelity images and scalability for high resolution. However, iterative sampling process required for synthesis is very resource-intensive. A line of work has focused on distilling solutions to probability flow ODEs into few-step student models. Nevertheless, existing methods have been limited by their reliance on the most recent denoised samples as input, rendering them susceptible to exposure bias. To address this limitation, we propose AutoRegressive Distillation (ARD), a novel approach that leverages the historical trajectory of the ODE to predict future steps. ARD offers two key benefits: 1) it mitigates exposure bias by utilizing a predicted historical trajectory that is less susceptible to accumulated errors, and 2) it leverages the previous history of the ODE trajectory as a more effective source of coarse-grained information. ARD modifies the teacher transformer architecture by adding token-wise time embedding to mark each input from the trajectory history and employs a block-wise causal attention mask for training. Furthermore, incorporating historical inputs only in lower transformer layers enhances performance and efficiency. We validate the effectiveness of ARD in a class-conditioned generation on ImageNet and T2I synthesis. Our model achieves a $5\times$ reduction in FID degradation compared to the baseline methods while requiring only 1.1\% extra FLOPs on ImageNet-256. Moreover, ARD reaches FID of 1.84 on ImageNet-256 in merely 4 steps and outperforms the publicly available 1024p text-to-image distilled models in prompt adherence score with a minimal drop in FID compared to the teacher. Project page: \url{https://github.com/alsdudrla10/ARD}.
\end{abstract}     
\vspace{-10mm}
\section{Introduction}
\label{sec:intro}

Diffusion models currently dominate image synthesis landscape due to their striking generalization capabilities and unprecedented visual quality~\cite{dhariwal2021diffusion,rombach2022high,podell2024sdxl,dai2023emu}. Unlike generative adversarial networks (GANs)~\cite{goodfellow2014generative}, the stable training of DMs facilitates their expansion to high-resolution image generation. Recently, models based on Diffusion Transformers (DiT)~\cite{peebles2023scalable} architecture gained significant popularity due to their excellent scaling properties and ability to generate high-resolution images~\cite{chenpixart,chen2024pixart_sigma}. However, sampling from DMs requires repeated neural network evaluations~\cite{lu2022dpm}, which makes the high-resolution image synthesis slow and resource-intensive.

DMs generate samples by solving the denoising process numerically. The denoising process has a probability flow ordinary differential equation (ODE) formulation~\cite{songdenoising,songscore}, which provides deterministic coupling between noise and samples. To reduce sampling costs, a series of distillation models~\cite{luhman2021knowledge,salimans2022progressive,zheng2023fast,song2023consistency,kim2024consistency,gu2024datafree,liu2023flow} have been developed that learn to predict ODE solution with fewer steps. However, few-step student models suffer from exposure bias~\cite{ranzato2016sequence,ning2024elucidating} because the student's intermediate prediction often deviates from the teacher's ODE due to estimation errors. The errors accumulate during iterative sampling, causing the prediction to become more erroneous as we approach a solution.

To address exposure bias in few-step distillation models, we propose an AutoRegressive Distillation (ARD) method for diffusion transformers. 
ARD predicts the next sample $\rvx_{\tau_{s-1}}$ based on both the current estimate $\rvx_{\tau_{s}}$ and the entire historical trajectory, which is more informative.
This approach offers two benefits: it reduces accumulated errors and provides a better source of coarse-grained information which is contained in the historical trajectory.
Incorporating the historical trajectory in the lower layers further introduces an inductive bias to handle coarse-grained information. We find that when distilling based on the whole historical trajectory, the FID degradation from the teacher is five times lower than that of the baselines on ImageNet 256p, with only 1.1\% more computation required. Our approach also scales well and can be used to distill 1024p text-to-image diffusion transformers, which outperform public distillation approaches in text-image alignment metrics.

\begin{figure*}[t!]
\centering
    \begin{minipage}[h]{0.625\textwidth}
         \centering
             \begin{subfigure}{1.0\textwidth}
                 \centering
                 \includegraphics[width=\textwidth, height=2.5in]{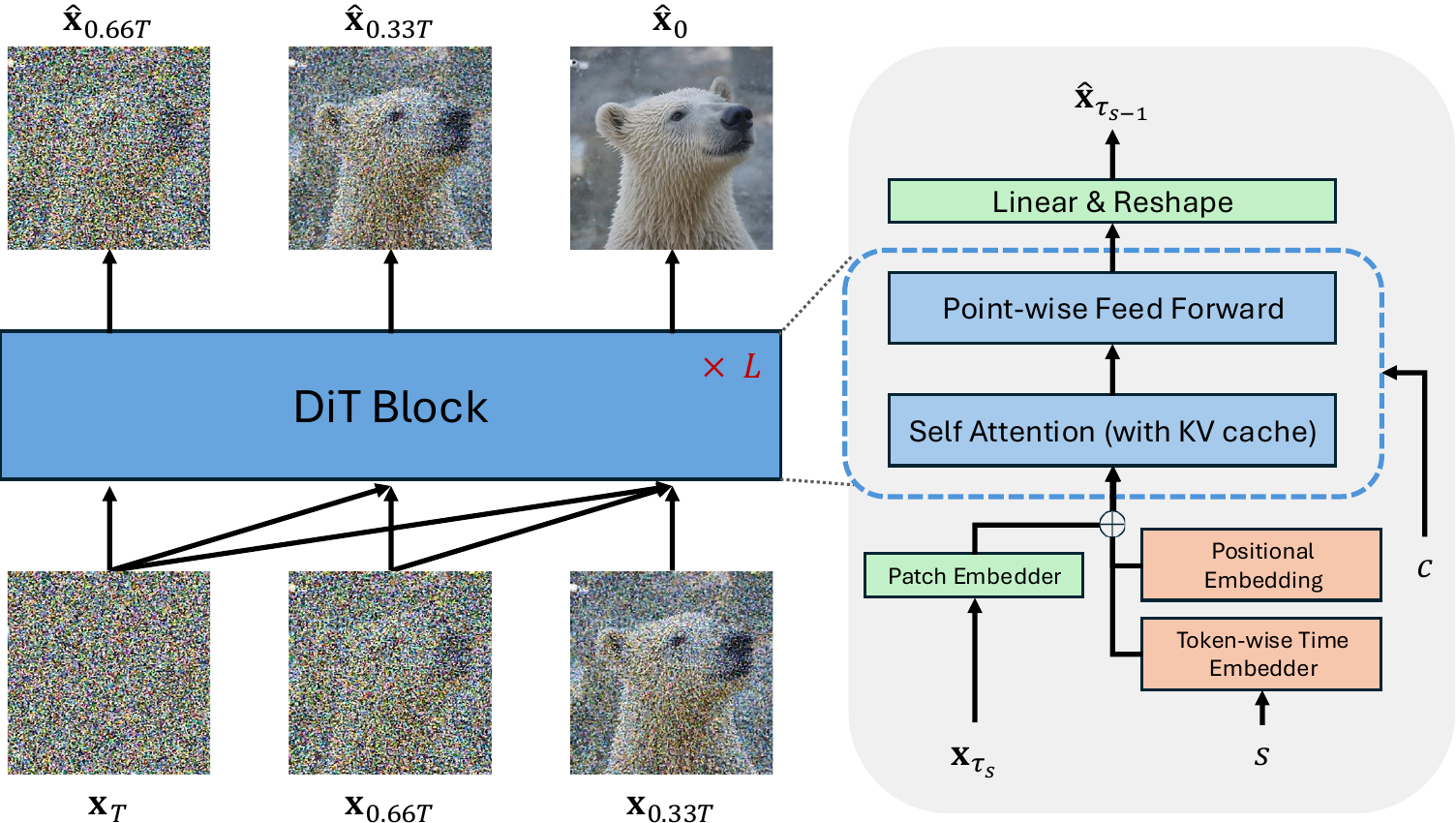}
                 \caption{A transformer architecture of ARD. The DiT parameters are shared across each input $\rvx_{\tau_{s}}$. }
                 \label{fig:3_a}
             \end{subfigure}
        \end{minipage}
    \begin{minipage}[h]{0.33\textwidth}
         \centering
             \begin{subfigure}{1.0\textwidth}
                 \centering
                 \includegraphics[width=\textwidth, height=2.5in]{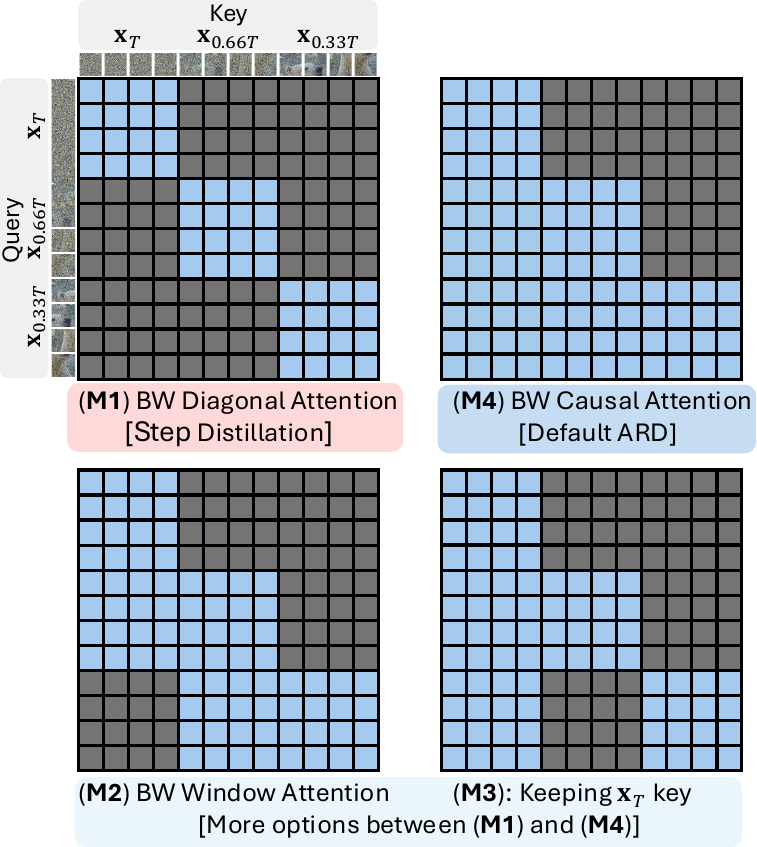}
                 \caption{Attention mask options. (BW: Block-Wise)}
                 \label{fig:3_b}
             \end{subfigure}
        \end{minipage}
        \vspace{-1.5mm}
        \caption{(a) The proposed transformer architecture for ARD. (b) The visualization of generalized mask options used during training: M1 represents step distillation, while M4 is the default setting of ARD. M2 and M3 are intermediate options between M1 and M4.}
            \label{fig:3}
            \vspace{-3.5mm}
\end{figure*}

\section{Preliminary}
\label{sec:preliminary}

\subsection{Diffusion models}
Diffusion models define a forward process and a corresponding reverse process with Stochastic Differential Equations (SDEs). The forward process in \cref{eq:fwd} maps from the data $\rvx_{0} \sim p_{\text{data}}(\rvx_0)$ to a noise $\rvx_T$.
\begin{align}
d\rvx_t = \mathbf{f}(\rvx_t,t)dt + g(t)d\mathbf{w}_t,\label{eq:fwd}
\end{align}
where $\mathbf{f}:\mathbb{R}^d\times[0,T] \rightarrow \mathbb{R}$ is a drift term, $g:[0,T]\rightarrow \mathbb{R}$ is a diffusion term, and $\mathbf{w}_t$ is a Wiener process. The forward process is often set to variance-preserving~\cite{ho2020denoising} or variance-exploding~\cite{songscore} SDEs to closely resemble a Gaussian distribution at $t=T$. Diffusion models generate the data from the noise $\rvx_T \sim p_{\text{prior}}(\rvx_T)$ through a reverse process~\citep{anderson1982reverse,songscore}. There exists a probability flow ODE (PF-ODE), which is a deterministic counterpart of the reverse process: 
\begin{align}
d\rvx_t = [\mathbf{f}(\rvx_t,t)-\frac{1}{2}g(t)^2\nabla_{\rvx_t}\log{p_t(\rvx_t)}]dt.\label{eq:bwd}
\end{align}
Here $p_t(\rvx_t)$ is the marginal distribution defined by the forward process in \cref{eq:fwd}. PF-ODE has the same marginal distribution as the reverse SDE while providing deterministic coupling between the noise $\rvx_T$ and the sample $\rvx_0$. Since the score function $\nabla_{\rvx_t}\log{p_t(\rvx_t)}$ is intractable, it is estimated by a neural network $\nabla_{\rvx_t}\log{p_t(\rvx_t)} \approx \nabla_{\rvx_t}\log{p^{\boldsymbol{\phi}}_t(\rvx_t)}$ with a score matching objective~\cite{song2019generative,vincent2011connection}.

\subsection{Step distillation models}
\label{sec:2.2}
The solution of an ODE in \cref{eq:bwd} is obtained by $\rvx_T + \int_{T}^{0}{\frac{d\rvx_t}{dt}}dt$; however, it requires a sufficient number of steps to reduce discretization error~\cite{de2021diffusion,lu2022dpm}. In order to compute $\frac{d\rvx_t}{dt}$ at each step, we need to evaluate the learned neural score function $\nabla_{\rvx_t}\log{p^{\boldsymbol{\phi^{*}}}_t(\rvx_t)}$, leading to high computational costs. To make inference efficient, step distillation~\cite{salimans2022progressive,meng2023distillation} defines intermediate times $\tau_{s}:=T\times\frac{s}{S}$  with $S$ as the total number of student steps and $s \in \{0,1,\ldots,S\}$. These intermediate times define a trajectory $\boldsymbol{\mu}_{\boldsymbol{\phi^{*}}} :=[\rvx_{\tau_S}, \rvx_{\tau_{S-1}},\ldots,\rvx_{\tau_{1}},\rvx_{\tau_{0}}]$ within the teacher ODE starting from an initial noise $\rvx_{\tau_{S}}=\rvx_T$ and ending with a clean sample $\rvx_{\tau_{0}}=\rvx_0$. The student model learns a joint probability $p(\boldsymbol{\mu}_{\boldsymbol{\phi^{*}}})$ defined as: 
\begin{align}
p(\boldsymbol{\mu}_{\boldsymbol{\phi^{*}}})= p_{\text{prior}}(\rvx_{\tau_{S}}) \times \prod_{s=1}^{S} p(\rvx_{\tau_{s-1}}|\rvx_{\tau_{s}})  \label{eq:step}
\end{align}
By the deterministic nature of PF-ODE, each conditional probability $p(\rvx_{\tau_{s-1}}|\rvx_{\tau_{s}})$ is a Dirac delta distribution, so it can be modeled by the deterministic mapping function; $\rvx_{\tau_{s-1}}= G(\rvx_{\tau_{s}},s):=\rvx_{\tau_{s}} + \int_{\tau_{s}}^{\tau_{s-1}}\frac{d\rvx_t}{dt}dt$. The student model $G_{\boldsymbol{\theta}}(\rvx_{\tau_{s}},s) \approx G(\rvx_{\tau_{s}}, s)$ learns to mimic the ground truth ODE integrations. Progressive distillation~\cite{salimans2022progressive,meng2023distillation} proposes a progressive algorithm for step distillation. However, such algorithm suffers from a significant drawback: the accumulation of errors during its iterative training phases when the student becomes the teacher again. 
Training a few-step student model directly from the teacher using $\mathcal{L}_{\text{step}}$ mitigates the accumulated errors brought the iterative progressive distillation procedure. We build our method on top of step distillation, where we directly learning from the teacher: 
\begin{align}
\mathcal{L}_{\text{step}}:=\mathbb{E}_{\boldsymbol{\mu}_{\boldsymbol{\phi^{*}}}}\left[ \sum_{s=1}^{S}||G_{\boldsymbol{\theta}}(\rvx_{\tau_{s}},s) - \rvx_{\tau_{s-1}}||_2^2\right].
\label{eq:step_loss}
\end{align}
\paragraph{Exposure bias} During inference, the generation starts from $\rvx_{\tau_{S}}\sim p_{\text{prior}}(\rvx_{\tau_{S}})$. At each step, the student model predicts $\hat{\rvx}_{\tau_{s-1}} = G_{\boldsymbol{\theta}}(\hat{\rvx}_{\tau_{s}},s)$ based only on the current sample $\hat{\rvx}_{\tau_{s}}$. If $\hat{\rvx}_{\tau_{s}}$ deviates from the teacher ODE, the student model $G_{\boldsymbol{\theta}}$ infers based on an unseen sample that was not encountered during training. Consider, for example, the intermediate samples depicted in \cref{fig:2_a}, where a fish is shown without eyes, despite such samples did not appear in the training data. 
This unforeseen input propagates through the sampling process, culminating in a final sample $\rvx_{\tau_{0}}$ that also lacks eyes.
This exposure bias is an inherent limitation of the iterative procedure~\cite{ranzato2016sequence,ning2024elucidating}, unless perfect optimization is achieved.
The errors accumulate as the iterative sampling process progresses.

\subsection{Autoregressive models}
\label{sec:2.3}

Autoregressive models~\cite{larochelle2011neural} represent the joint probability distribution of a multivariate random variable $\rvx:=[x_{S},x_{S-1},\ldots,x_0]$ by decomposing it into a product of conditional probabilities  
$p(\rvx)= p(x_S) \times \prod_{s=1}^{S} p(x_{s-1}|\rvx_{S:s})$,
where $\rvx_{S:s}=[x_{S},x_{S-1},\ldots,x_{s}]$. 
This formulation, as depicted above, does not rely on any specific assumptions. Each component $p(x_{s-1}|\rvx_{S:s})$ of the decomposition incorporates the information of all preceding variables.

\section{Method}
\begin{figure*}[t]
    \centering
    \begin{minipage}[h]{0.27\textwidth}
        \begin{subfigure}{1.0\textwidth}
                 \centering
                 \includegraphics[width=\textwidth, height=1.8in]{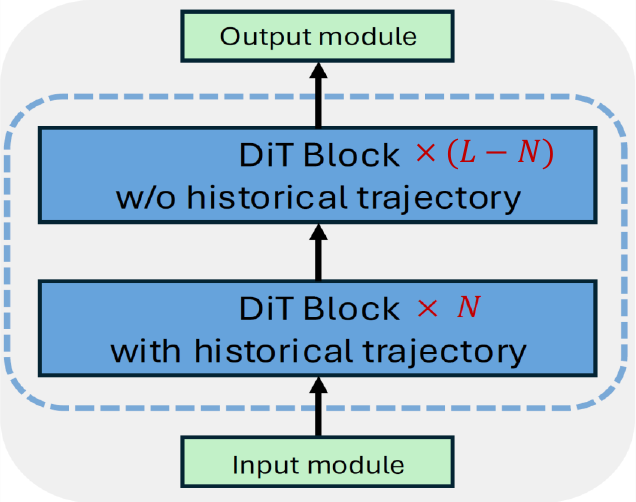}
                 \caption{Using historical trajectory in $N$ layers.} 
                  \label{fig:4a}
             \end{subfigure}
    \end{minipage}
     \begin{minipage}[h]{0.245\textwidth}
     \centering
         \begin{subfigure}{1.0\textwidth}
             \centering
             \includegraphics[width=\textwidth, height=1.05in]{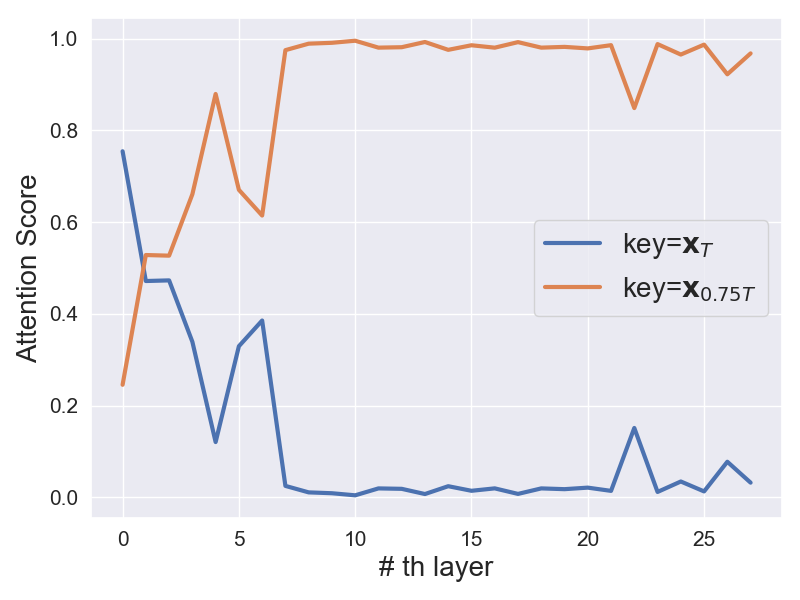}
             \caption{$N=28$ / Query: $\rvx_{0.75T}$}
             \label{fig:4b}
         \end{subfigure}
         \begin{subfigure}{1.0\textwidth}
             \centering
             \includegraphics[width=\textwidth, height=1.05in]{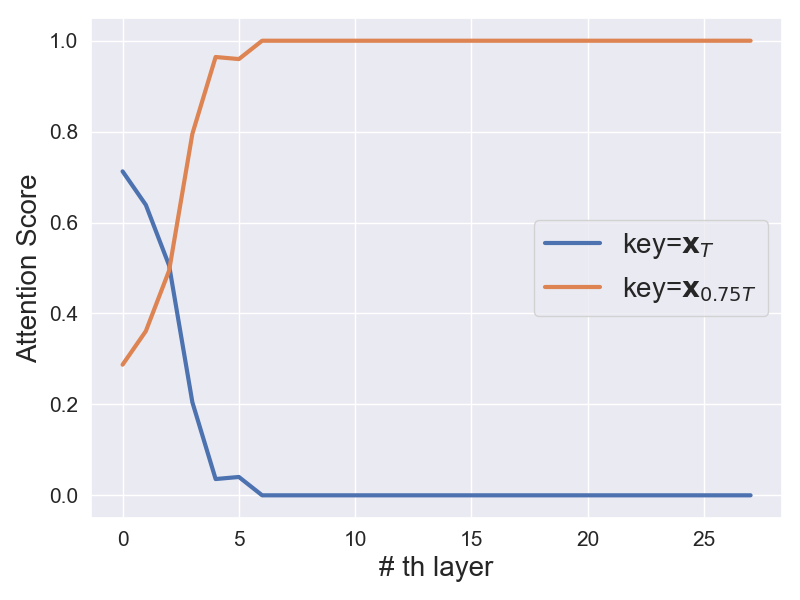}
             \caption{$N=6$ / Query: $\rvx_{0.75T}$} 
              \label{fig:4c}
         \end{subfigure}
    \end{minipage}
    \begin{minipage}[h]{0.23\textwidth}
     \centering
         \begin{subfigure}{1.0\textwidth}
             \centering
             \includegraphics[width=\textwidth, height=1.05in]{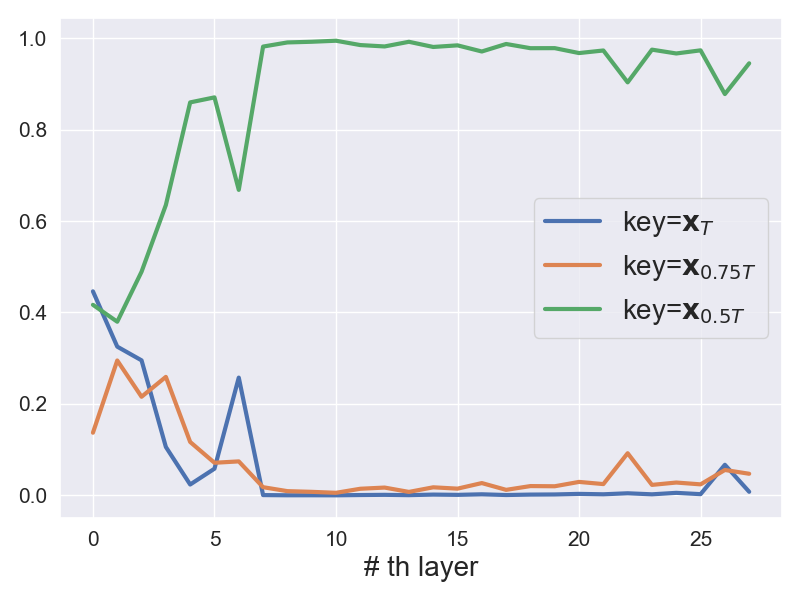}
             \caption{$N=28$ / Query: $\rvx_{0.5T}$}
             \label{fig:4d}
         \end{subfigure}
         \begin{subfigure}{1.0\textwidth}
             \centering
             \includegraphics[width=\textwidth, height=1.05in]{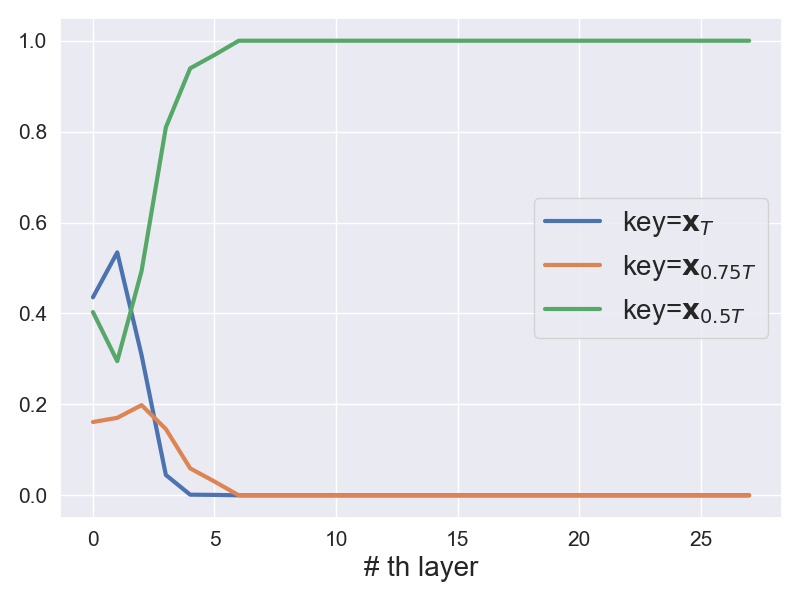}
    
             \caption{$N=6$ / Query: $\rvx_{0.5T}$} 
              \label{fig:4e}
         \end{subfigure}
    \end{minipage}
    \begin{minipage}[h]{0.23\textwidth}
     \centering
         \begin{subfigure}{1.0\textwidth}
             \centering
             \includegraphics[width=\textwidth, height=1.05in]{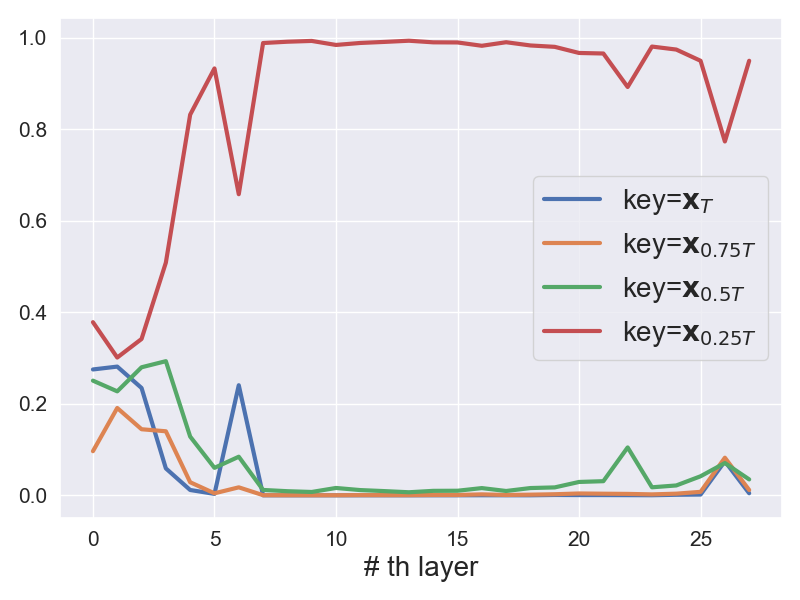}
             \caption{$N=28$ / Query: $\rvx_{0.25T}$}
             \label{fig:4f}
         \end{subfigure}
         \begin{subfigure}{1.0\textwidth}
             \centering
             \includegraphics[width=\textwidth, height=1.05in]{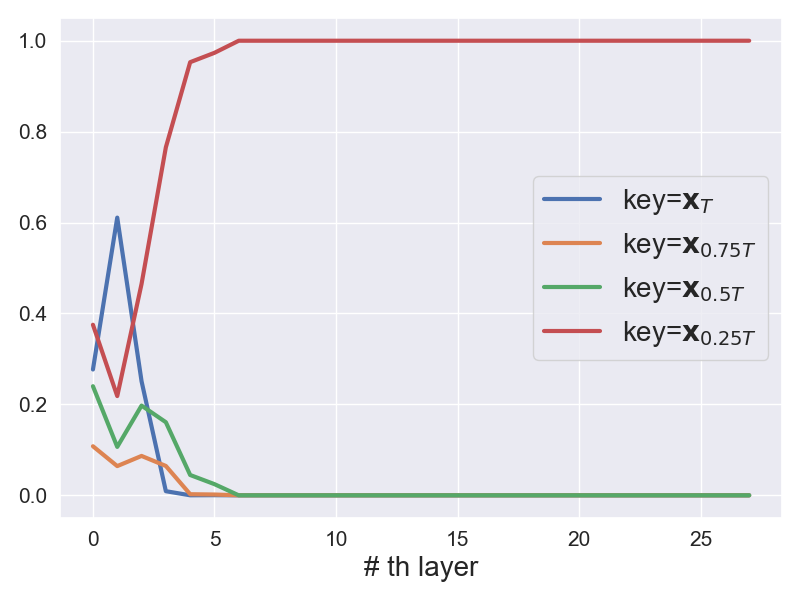}
             \caption{$N=6$ / Query: $\rvx_{0.25T}$} 
              \label{fig:4g}
         \end{subfigure}
    \end{minipage}
     \vspace{-1.5mm}
    \caption{(a) shows an additional inductive bias that we impose by using the historical trajectory in lower layers only. (b, d, f) show the attention scores for each history input (key tokens) during the 2$^{\text{nd}}$, 3$^{\text{rd}}$, 4$^{\text{th}}$ steps when $N=L$. (c, e, g) show the same but with $N=6$. The attention score on input $\rvx_{\tau_{s'}}$ is the sum of attention weights for all key tokens in $\rvx_{\tau_{s'}}$, indicating the portion of $\rvx_{\tau_{s'}}$.} 
            \label{fig:4}
            \vspace{-3.5mm}
\end{figure*}
In this section we introduce the AutoRegressive Distillation (ARD) of diffusion transformers (DiT). \Cref{fig:2_b} provides an overview of the ARD process. 
We'll break down the probabilistic formulations of distillation in \Cref{sec3.1}, then move on to the transformer architecture design for our student model in \Cref{sec3.2}. Lastly, we'll cover training and inference in \Cref{sec3.3}.

\subsection{Autoregressive distillation }
\label{sec3.1}
This section generalizes the step distillation formulation in \cref{eq:step} to ARD. The decomposition in \cref{eq:step} is valid without whole historical trajectory information under perfect distillation. However, when each probability $p(\rvx_{\tau_{s-1}}|\rvx_{\tau_{s}})$ is approximated by $\hat{\rvx}_{\tau_{s-1}}=G_{\boldsymbol{\theta}}(\rvx_{\tau_{s}},s)$, the discrepancy with the ground truth is inevitable due to estimation error, leading to the exposure bias problem discussed in \cref{sec:2.2}.

To mitigate this problem, we extend the formulation of \cref{eq:step} in an autoregressive manner motivated by \cref{sec:2.3}:
\begin{align}
p(\boldsymbol{\mu}_{\boldsymbol{\phi}^{*}})=  p_{\text{prior}}(\rvx_{\tau_{S}}) \times \prod_{s=1}^{S} p(\rvx_{\tau_{s-1}}|\rvx_{\tau_{S}:\tau_{s}}), \label{eq:ours}
\end{align}
where $\rvx_{\tau_{S}:\tau_{s}}=[\rvx_{\tau_{S}},\rvx_{\tau_{S-1}}, \ldots ,\rvx_{\tau_{s}}]$ denotes the historical trajectory. This formulation has two benefits: (i) Every step includes the ground truth initial noise $\rvx_{\tau_{S}}$ as input, which has a deterministic coupling with the prediction target $\rvx_{\tau_{s-1}}$. Furthermore, the historical trajectory predictions from $\hat{\rvx}_{\tau_{S-1}}$ to $\hat{\rvx}_{\tau_{s+1}}$ are more accurate compared to the recent sample $\hat{\rvx}_{\tau_{s}}$ because for them the error had less chances to accumulate during inference. In contrast, the input in \cref{eq:step} is merely the current sample $\hat{\rvx}_{\tau_{s}}$, making it vulnerable to exposure bias. (ii) To predict $\rvx_{\tau_{s-1}}$ at every step, the model needs to generate both coarse-grained and fine-grained information. The recent denoised sample $\rvx_{\tau_{s}}$ is the best source for fine-grained information, but the historical trajectory close to $\rvx_{\tau_{S}}$ is a better source for coarse-grained information~\cite{rissanen2023generative,dieleman2024spectral}. 

For the modified student formulation, we aim to estimate $p(\rvx_{\tau_{s-1}}|\rvx_{\tau_{S}:\tau_{s}})$, which is still a Dirac delta distribution. To achieve this we define a new mapping function $\rvx_{\tau_{s-1}} = G(\rvx_{\tau_{S}:\tau_{s}},s) := \rvx_{\tau_{s}}+ \int_{\tau_{s}}^{\tau_{s-1}}\frac{d\rvx_t}{dt}dt$.
This function is then approximated by a student neural network $G_{\boldsymbol{\theta}}(\rvx_{\tau_{S}:\tau_{s}},s)$.

\subsection{Transformer design}
\label{sec3.2}
 The design of our mapping function $G_{\boldsymbol{\theta}}(\rvx_{\tau_{S}:\tau_{s}},s)$ defined in \cref{sec3.1} is not trivial because the input size varies depending on the denoising step $s$. To overcome this, we modify the teacher DiT backbone to accommodate multiple inputs.

\paragraph{Architecture}
To handle the historical trajectory, we design transformer-based autoregressive model as shown in \cref{fig:3_a}. Each input $\rvx_{\tau_{s}}$ is tokenized into a sequence of tokens using a shared patch embedder. Since each input $\rvx_{\tau_{s}}$ has the same spatial structure as a 2D grid, positional embeddings are shared across the inputs. The transformer blocks need to identify the order of each token in the inputs sequence $\rvx_{\tau_{S}}, \ldots, \rvx_{\tau_{s}}$. To this end, we add an extra time-step embeddings to each token similar to the level embedding in VAR~\cite{tian2024visual}.\footnote{The original DiT backbone uses time embedding with adaLN~\cite{perez2018film} because the tokens in the DiT teacher are always from the same input, so it does not need to identify time on a token-wise basis. On the other hand, our student model needs to be modified to identify the origin of each token.} The recent denoised sample $\rvx_{\tau_{s}}$ becomes the query tokens, and the history sequence $\rvx_{\tau_{S}:\tau_{s}}$ becomes the key-value tokens in the self-attention blocks.  After passing through $L$ stacked transformer blocks, the tokens are linearly transformed and de-tokenized to obtain a sample $\rvx_{\tau_{s-1}}$.

\begin{table*}[b!]
\small
 \vspace{-0mm}
    \centering
    \caption{Comparison with distillation methods from the same teacher. The teacher uses DiT-XL/2 architecture trained on ImageNet 256p. Loss R denotes the use of regression loss for distillation, and R+D denotes using additional discriminator loss. The FLOPs and latency are measured during the denoising process. Latency refers to the time required to generate one image measured on an H100.} 
          \begin{tabular}{l|cc|ccc|cccc}
        \toprule
             Model  &Loss&Mask& Steps ($S$) $\downarrow$ & GFLOPs $\downarrow$ & 
             Latency (ms) $\downarrow$& FID $\downarrow$ & IS $\uparrow$ & Prec $\uparrow$ & Rec $\uparrow$ \\
            \midrule
            \midrule
          DiT/XL-2&  - &-& 250 &59300&4935&2.27&278.24&0.830&0.570 \\
          (target teacher) &  -&- & 25 &5930&493.5&2.89&230.22&0.797&0.572 \\
          \midrule
          \midrule
          KD~\cite{luhman2021knowledge}   &R&-& 1 &118.6&17.01&11.88&148.61&0.665&0.565 \\
          \midrule
          Step Distill.~\cite{salimans2022progressive} ($N=0$)&R&M1&2&237.2 &33.05&10.92&167.08&0.681&0.518 \\
          \rowcolor{gray!25}ARD ($N=6$)&R&M4&2&238.1& 34.05&\textbf{6.29}&\textbf{188.05}&\textbf{0.737}&0.564 \\
          \rowcolor{gray!25}ARD ($N=28$)&R&M4&2&241.5 &34.90&6.54&186.18&0.734&\textbf{0.569} \\
          \midrule
          Step Distill.~\cite{salimans2022progressive} ($N=0$)&R&M1&4&474.4&64.80&10.25&181.58&0.704&0.474 \\
          \rowcolor{gray!25}ARD ($N=6$)&R&M2&4&477.1&65.57&4.75&203.58&0.768&0.572 \\
          \rowcolor{gray!25}ARD ($N=6$)&R&M3&4&477.1&65.57&4.45&206.93&0.773&0.572 \\
          \rowcolor{gray!25}ARD ($N=6$)&R&M4&4&479.9 &66.34&\textbf{4.32}&\textbf{209.03}&\textbf{0.770}&\textbf{0.574} \\
          \rowcolor{gray!25}ARD ($N=28$)&R&M4&4&500.2&67.85&4.80&201.15&0.761&0.566 \\
          \midrule
          Step Distill.~\cite{salimans2022progressive} ($N=0$)&R+D&M1&4&474.4&64.80&3.84&221.16&0.785&0.557 \\
          \rowcolor{gray!25}ARD ($N=6$)&R+D&M4&4&479.9 &66.34&\textbf{1.84}&\textbf{235.84}&\textbf{0.797}&\textbf{0.615} \\
        \bottomrule
    \end{tabular}
    \label{tab:main1}
\end{table*}
\begin{figure*}[b!]
\centering
     \begin{minipage}[h]{0.48\textwidth}
     \centering
         \begin{subfigure}{1.0\textwidth}
             \centering
             \includegraphics[width=\textwidth, height=0.7in]{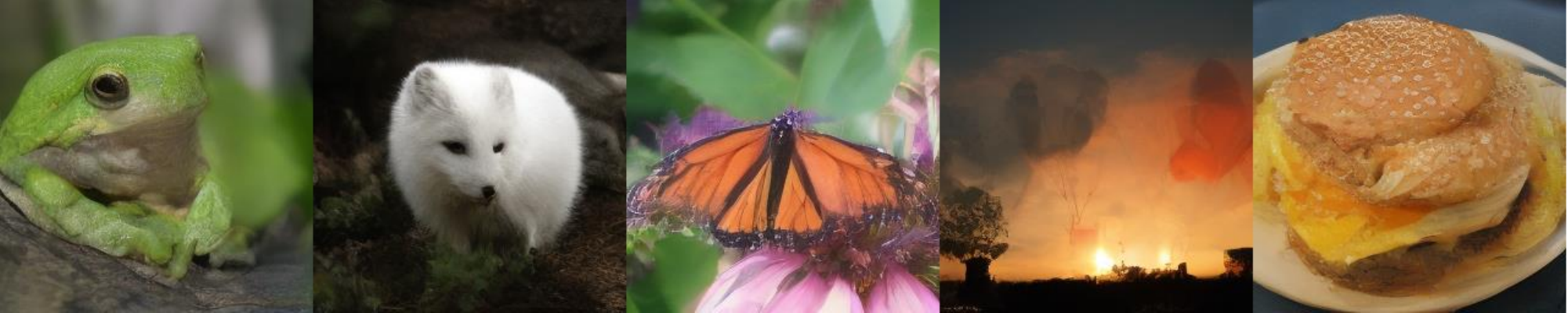}
             \caption{Step Distillation (R) / FID: 10.25}
             \label{fig:6_a}
         \end{subfigure}
         \begin{subfigure}{1.0\textwidth}
             \centering
             \includegraphics[width=\textwidth, height=0.7in]{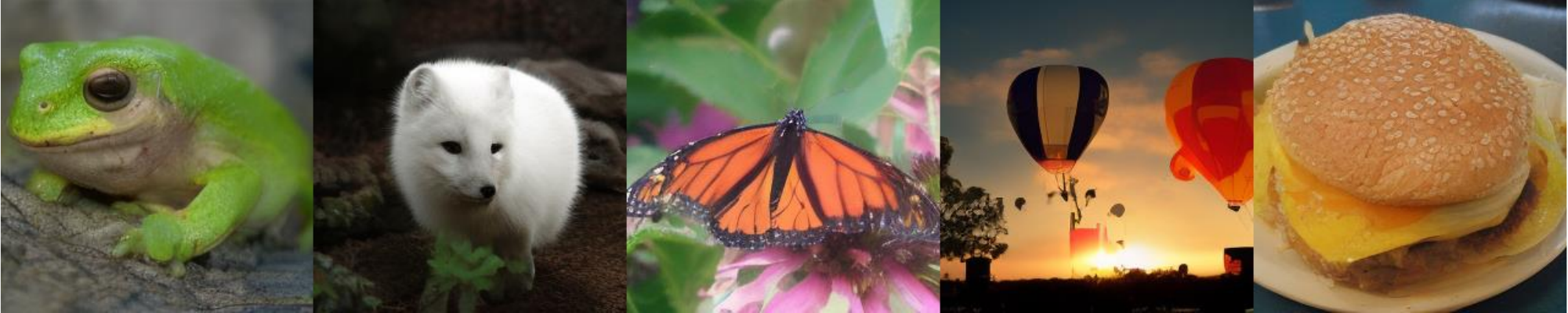}
             \caption{ARD (R) / FID: 4.32} 
              \label{fig:6_b}
         \end{subfigure}
    \end{minipage}
    \begin{minipage}[h]{0.48\textwidth}
        \begin{subfigure}{1.0\textwidth}
             \centering
             \includegraphics[width=\textwidth, height=0.7in]{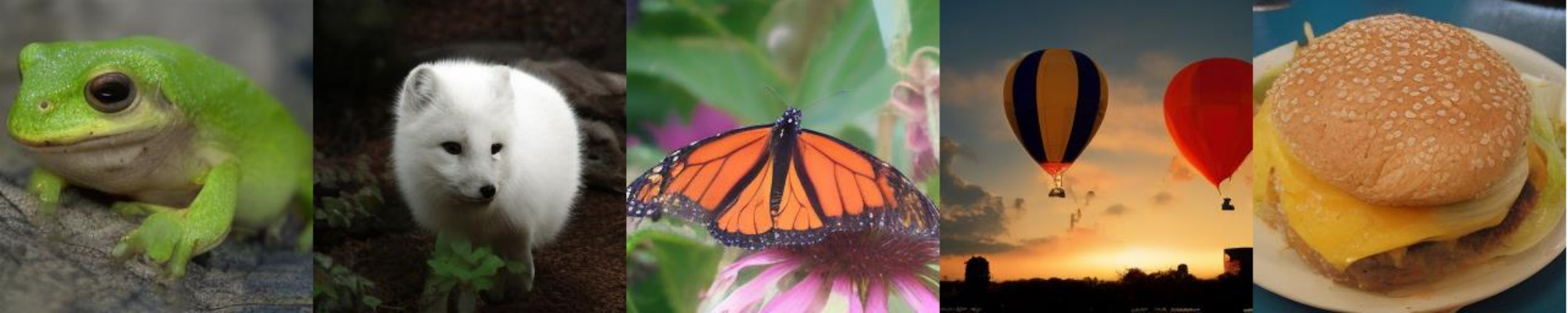}
             \caption{Teacher (25 steps) / FID: 2.89}
             \label{fig:6_c}
         \end{subfigure}
         \begin{subfigure}{1.0\textwidth}
             \centering
             \includegraphics[width=\textwidth, height=0.7in]{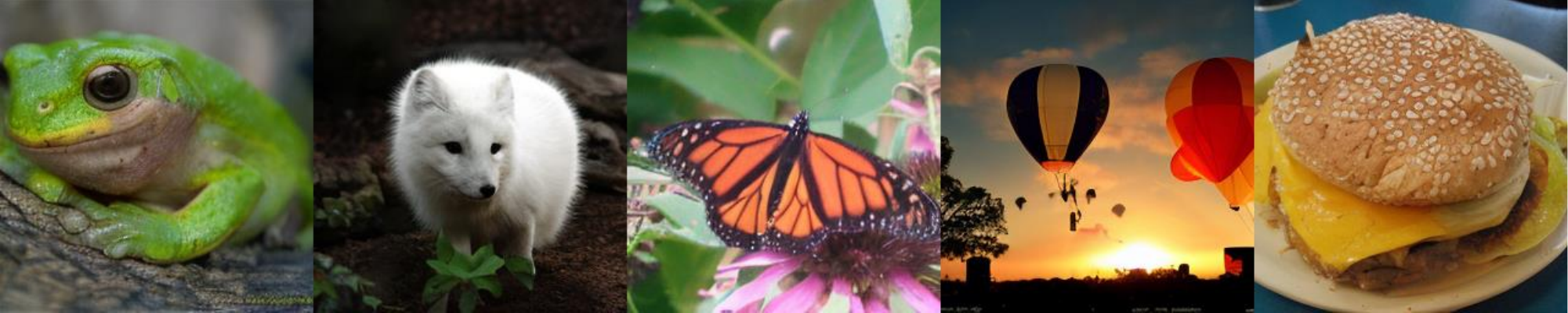}
             \caption{ARD (R+D) / FID: 1.84} 
              \label{fig:6_d}
         \end{subfigure}
    \end{minipage}
    \caption{Generated ImageNet 256p samples from same initial noise $\rvx_{\tau_{S}}$. All distilled models are 4-step models.}
            \label{fig:6}
            \vspace{-0mm}
\end{figure*}

\paragraph{Historical trajectory only in lower $N$ layers.}
\Cref{fig:4b,fig:4d,fig:4f} show the attention scores of each input in (2$^{\text{nd}}$, 3$^{\text{rd}}$, 4$^{\text{th}}$) steps at each $L$ transformer layers. The recent denoised sample $\rvx_{\tau_{s}}$ is most activated as key tokens in the higher layers, while the historical trajectory $\rvx_{\tau_{S}:\tau_{s+1}}$ is activated in the lower layers. The lower layers in DiT blocks are known to consider coarse-grained information, while the higher layers in DiT blocks are considered fine-grained information~\cite{hatamizadeh2025diffit}. This attention portion validates that the historical trajectory is useful and serves as a better source of coarse-grained information. However, the historical tokens still slightly fluctuate in the higher layers in \cref{fig:4b,fig:4d,fig:4f}, possibly due to imperfect optimization. We propose additional design choices in transformer layers as shown in \cref{fig:4a}; using the historical trajectory only in the lower $N$ layers. This inductive bias enhances the use of the historical trajectory in the lower layers as shown in \cref{fig:4c,fig:4e,fig:4g}.

\begin{figure*}[t]
    \centering
    \begin{minipage}[h]{1.0\textwidth}
     \begin{subfigure}[b]{0.34\textwidth}
         \centering
         \includegraphics[width=\textwidth, height=1.6in]{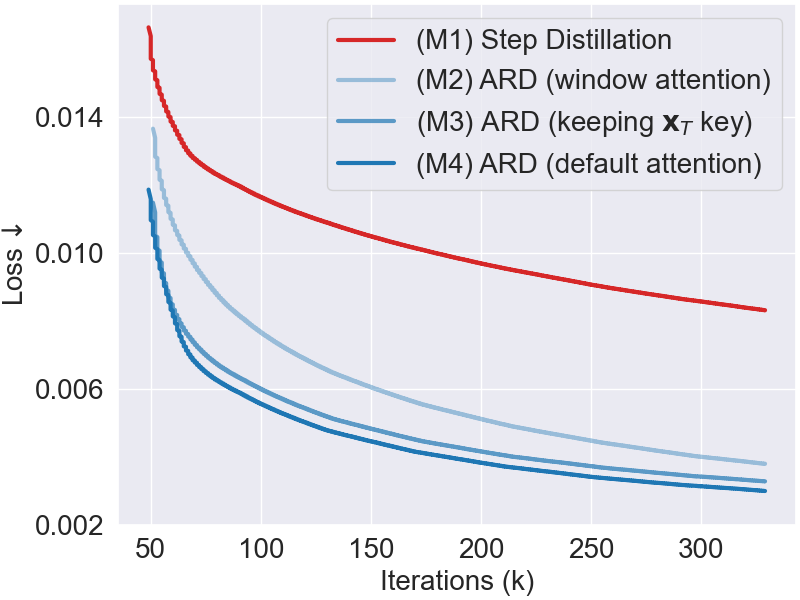}
         \caption{Training curve on 4-step models.} 
          \label{fig:5_a}
     \end{subfigure}
     \begin{subfigure}[b]{0.32\textwidth}
         \centering
         \includegraphics[width=\textwidth, height=1.6in]{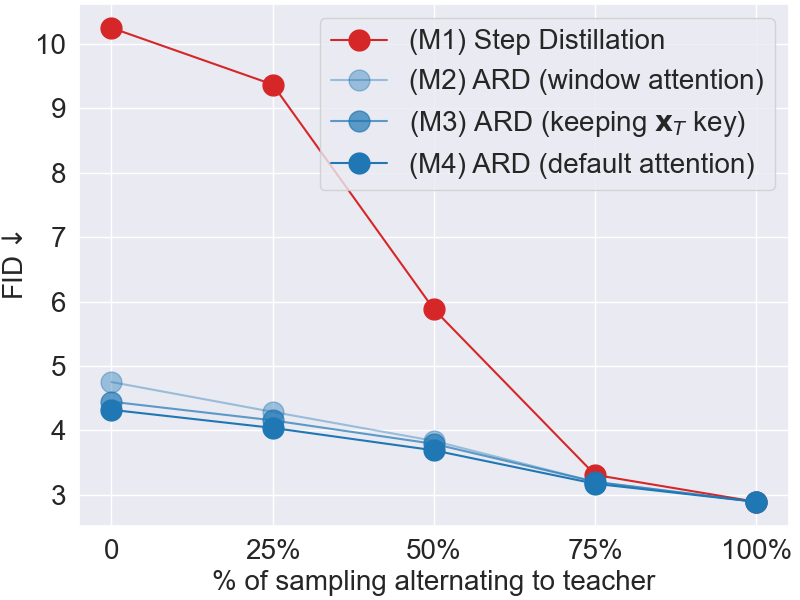}
         \caption{Error in each step in 4-step models}
         \label{fig:5_b}
     \end{subfigure}
    \begin{subfigure}[b]{0.32\textwidth}
         \centering
         \includegraphics[width=\textwidth, height=1.6in]{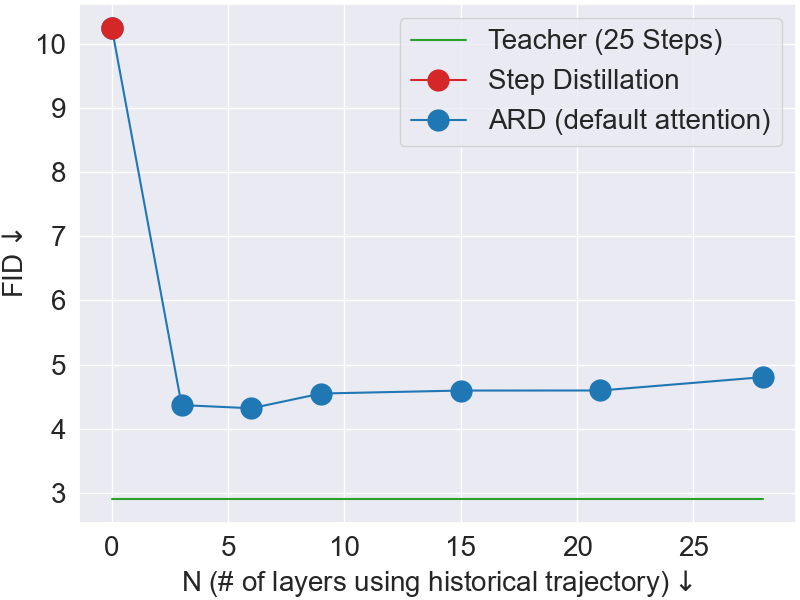}
         \caption{$N$ ablations on 4-step models.}
          \label{fig:5_c}
     \end{subfigure}
    \end{minipage}
    \vspace{-2.0mm}
    \caption{The analysis on design choices (attention mask options \& $N$) for 4-step distillation methods.} 
    \vspace{-4.0mm}
    \label{fig:5}
\end{figure*}

\subsection{Training and inference procedure}
\label{sec3.3}
The default training objective of ARD is a regression loss $\mathcal{L}_{\text{ARD}}$ in \cref{eq:ours_loss}, and it is optimized with respect to $\boldsymbol{\theta}$. The transformer architecture in \cref{fig:3_a} allows computing $\hat{\rvx}_{\tau_{s-1}}=G_{\boldsymbol{\theta}}(\rvx_{\tau_{S}:\tau_{s}},s)$ for all $s\in \{1,\ldots,S\}$ simultaneously by using an attention mask. We can generalize our framework by designing the attention mask as shown in \cref{fig:3_b}. Block-wise causal attention in option M4 is the most flexible, as it uses the entire trajectory history. Option M1 represents step distillation, which only uses the current sample ${\rvx}_{\tau_{s}}$ as input. Options M2 and M3 are intermediate choices between M1 and M4. The windowed attention in M2 uses only the current and the previous sample from the trajectory history. The attention mask in M3 uses the most recent denoised sample and the initial noise $\rvx_{\tau_{S}}$, which helps to consistently preserve the ground truth signal. 
Our framework can also benefit from an additional discriminator loss for the final prediction
$\hat{\rvx}_{\tau_{0}} = G_{\boldsymbol{\theta}}(\rvx_{\tau_{S}:\tau_{1}},1)$, similar to \cite{kim2024consistency}. 
By using real data in this loss, we can further improve the high-frequency details in the student generations, even outperforming the teacher.
\begin{align}
\mathcal{L}_{\text{ARD}}:=\mathbb{E}_{\boldsymbol{\mu}_{\boldsymbol{\phi^{*}}}}\left[ \sum_{s=1}^{S}||G_{\boldsymbol{\theta}}(\rvx_{\tau_{S}:\tau_{s}},s) - \rvx_{\tau_{s-1}}||_2^2\right]\label{eq:ours_loss}
\end{align}
During inference, the generation starts from $\rvx_{\tau_{S}}\sim p_{\text{prior}}(\rvx_{\tau_{S}})$. At each step, the student model predicts $\hat{\rvx}_{\tau_{s-1}}=G_{\boldsymbol{\theta}}(\hat{\rvx}_{\tau_{S}:\tau_{s}},s)$ based on the entire historical predictions $\hat{\rvx}_{\tau_{S}:\tau_{s}}=[\rvx_{\tau_{S}},\hat{\rvx}_{\tau_{S-1}},\ldots,\hat{\rvx}_{\tau_{s}}]$. The information of $[\rvx_{\tau_{S}},\hat{\rvx}_{\tau_{S-1}},\ldots,\hat{\rvx}_{\tau_{s+1}}]$ is stored as kv-cache in the previous steps for fast inference. No attention mask is required during inference.

\section{Experiments}

This section empirically validates the effectiveness of ARD. \Cref{sec:4.1} explains the results of class-conditional image synthesis on ImageNet~\cite{deng2009imagenet}. \Cref{sec:4.2} presents the experimental results for text-conditional image synthesis.

\subsection{Class-conditional image generation}
\label{sec:4.1}

We use a DiT/XL-2 latent diffusion transformer architecture following \cite{peebles2023scalable}, and employ it as a teacher architecture. The teacher ($\boldsymbol{\phi}$) is trained on ImageNet 256p. We construct an ODE trajectory $\boldsymbol{\mu}_{\boldsymbol{\phi^{*}}}$ by running the teacher with 25 steps and a classifier-free guidance scale \cite{ho2022classifier} of 1.5. In total, we pre-compute and store $2.56$M ODE trajectories for the distillation.

\paragraph{Evaluation metrics}
To evaluate sample fidelity and diversity, we measure FID~\cite{heusel2017gans}, IS~\cite{salimans2016improved}, Precision, and Recall~\cite{kynkaanniemi2019improved} following the protocol of ADM~\cite{dhariwal2021diffusion} with a pre-trained Inception-V3 Network~\cite{szegedy2015going}. FID and IS quantify both sample quality and diversity. Precision measures sample fidelity, while Recall measures sample diversity by quantifying the manifold overlap region between real and generated samples in the feature space.

\paragraph{Performance gain from ARD} 

The proposed ARD is a generalization over previous methods~\cite{luhman2021knowledge,salimans2022progressive}: When the number of steps $S$ is 1, ARD becomes Knowledge Distillation (KD)~\cite{luhman2021knowledge}. When ARD is used with attention mask option M1 in \cref{fig:3_b}, it becomes step distillation~\cite{salimans2022progressive}. \Cref{tab:main1} and \cref{fig:2_c} show the performance gain from the extended design of ARD. 
Increasing the number of steps from 2 to 4 for our method results in better performance in FID:  $6.29\xrightarrow{}4.32$. 
However, the FID gain in step distillation is marginal ($10.92\xrightarrow{}10.25$) even when the number of steps $S$ increases from 2 to 4, moreover Recall decreases significantly. 
In \Cref{fig:6_a,fig:6_c} we can see that step distillation fails to preserve the global structure (e.g., the orientation of a frog), indicating that the diversity of the teacher's solution is not maintained. This happens because as the number of steps increases in step distillation, 
it becomes harder to preserve the deterministic coupling between the initial noise $\rvx_{\tau_{S}}$ and the sample $\rvx_{\tau_{0}}$ provided by the teacher due to increased exposure bias (see \Cref{sec:2.2}).

\begin{table}[t]
 \vspace{-0mm}
    \centering
    \caption{Comparison with public models on ImageNet 256p}
         \vspace{-2.5mm}
    \footnotesize
          \begin{tabular}{clcc|ccc}
        \toprule
             Type&Model & Params $\downarrow$ & Steps $\downarrow$ & FID $\downarrow$  &  Rec $\uparrow$  \\
            \midrule
            \midrule
            GAN&BigGAN~\cite{brock2018large}&112M&1 &6.95&0.38 \\
            GAN&StyleGAN-XL~\cite{sauer2022stylegan}&166M&1 &2.30&0.53 \\
            GAN&GigaGAN~\cite{kang2023scaling}&569M& 1 &3.45&0.61 \\
            \midrule
            DM &ADM~\cite{dhariwal2021diffusion}& 554M & 250&4.59&0.52 \\
           DM &LDM~\cite{rombach2022high}& 400M & 250&3.60&0.48 \\
          DM&DiT~\cite{peebles2023scalable}& 675M & 250&2.27&0.57 \\
          DM&VDM++~\cite{kingma2024understanding}& 2.0B & 250&2.40&- \\
         \midrule
         NAT &MaskGIT~\cite{chang2022maskgit}&227M&8 &6.18&0.51 \\
          NAT &RCG~\cite{li2024return}&502M&20 &2.15&0.53 \\
          NAT &AutoNAT~\cite{ni2024revisiting}&194M&8 &2.68&- \\
          \midrule
          AR&VQVAE2~\cite{razavi2019generating}&13.5B&5120 &31.11&0.57 \\
          AR&VQGAN~\cite{esser2021taming}&227M&256 &18.65&0.26 \\
          AR&RQTran.~\cite{lee2022autoregressive}&3.8B&68 &7.55&- \\
          AR&VAR~\cite{tian2024visual}&2.0B&10 &1.97&0.59 \\
          AR&MAR~\cite{li2024autoregressive}&943M&32 &1.93&- \\
          \rowcolor{gray!25}AR&ARD (Ours) &675M&4 &\textbf{1.84}&\textbf{0.62} \\
        \bottomrule
    \end{tabular}
    \label{tab:main2}
     \vspace{-5.5mm}
\end{table}
\begin{table*}[b!]
 \vspace{-0mm}
    \centering
    \caption{Text-image alignment scores on CompBench for public high-resolution ($\geq 768$p) distillation models. Loss R denotes the use of regression loss, and D denotes the use of discriminator loss. The best and second-best results are highlighted in \textbf{bold} and \underline{underline}.}
     \footnotesize
          \begin{tabular}{lccccc|cccccc|c}
        \toprule
         &  &  &  & & &\multicolumn{5}{c}{\textbf{\quad \quad \quad \quad \quad CompBench $\uparrow$ (\%)}}&& \\
             Model & Params $\downarrow$ &Steps $\downarrow$ & Res.& CFG & Loss& Color &Shape&Texture&Spatial&Non-spatial&Complex&AVG $\uparrow$ \\
            \midrule
          Step distill.~\cite{salimans2022progressive}   & 2.7B & 4&768&6&R&44.1&26.9&40.0&48.7&54.3&39.6&42.3 \\
          ADD~\cite{sauer2025adversarial}   & 2.7B & 3&768&6&D&43.3&32.9&44.5&42.6&56.2&36.5&42.6 \\
          Imagine Flash~\cite{kohler2024imagine}   & 2.7B & 3&768&6&R+D&42.7&36.2&47.9&57.4&62.9&42.8&48.3 \\
         \midrule
          Lightning~\cite{lin2024sdxl}   & 2.6B& 4& 1024&6&D&57.1&46.5&53.1&62.0&61.4&41.9&53.7 \\
          DMD2~\cite{yin2024improved}& 2.6B& 4& 1024&8&R+D&64.5&\underline{47.8}&60.4&\textbf{68.9}&\textbf{66.3}&\textbf{49.1}&59.5 \\
          \midrule
          Pixart-delta~\cite{chen2024pixart}&0.6B&3 &1024&4.5&R&38.7&33.2&40.8&56.1&60.1&39.5&44.7\\
          LCM-LoRA~\cite{luo2023lcm}&1.4B&3 &1024&7.5&R&49.3&35.2&45.0&54.5&57.0&39.9&46.8 \\
          LCM-LoRA~\cite{luo2023lcm}&2.8B&3 &1024&7.5&R&59.5&45.3&50.2&55.3&60.8&44.6&52.6 \\
          \rowcolor{gray!25}ARD&1.7B&3 &1024&3&R&\underline{64.6}&46.6&\underline{63.5}&60.0&57.8&\underline{44.8}&56.2 \\
          \rowcolor{gray!25}ARD&1.7B&3&1024 &7.5&R&\textbf{71.1}&\textbf{53.1}&\textbf{65.7}&\underline{64.4}&\underline{61.2}&44.0&\textbf{59.9} \\
        \bottomrule
    \end{tabular}
    \vspace{-1.5mm}
    \label{tab:main3}
\end{table*}
\begin{figure*}[b!]
\centering
     \begin{minipage}[h]{0.49\textwidth}
     \centering
         \begin{subfigure}{1.0\textwidth}
             \centering
             \includegraphics[width=\textwidth, height=1.1in]{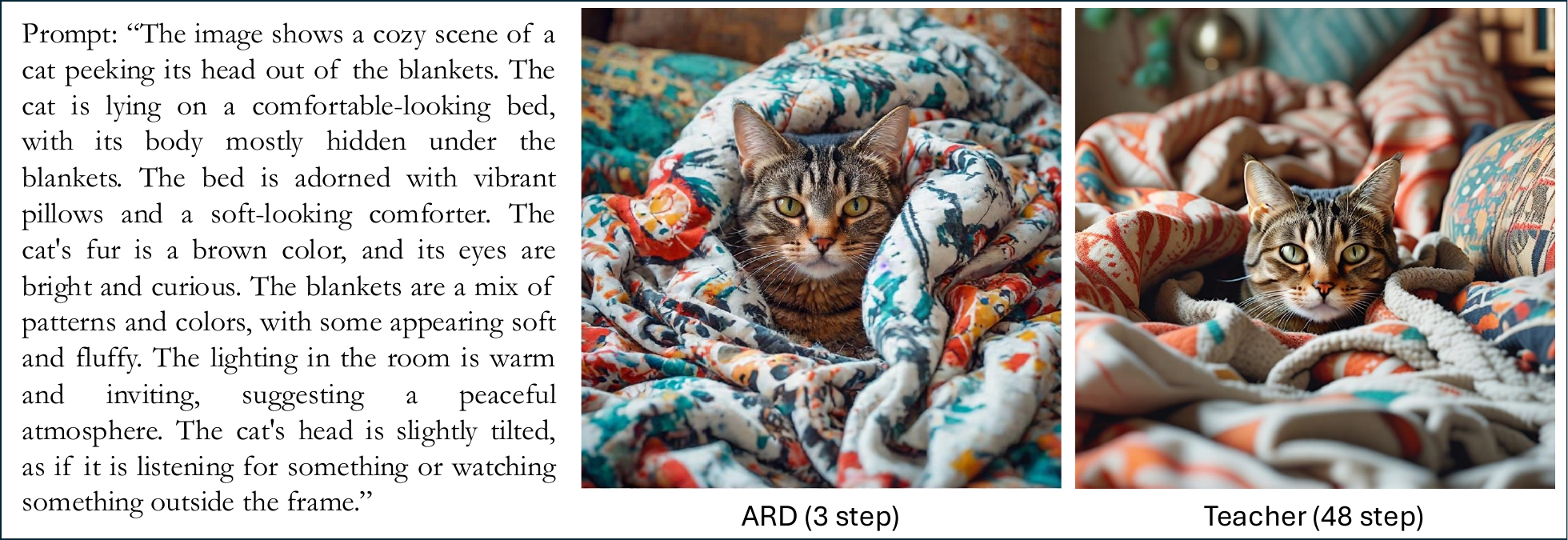}
              \label{fig:7_a}
         \end{subfigure}
    \end{minipage}
    \begin{minipage}[h]{0.49\textwidth}
         \begin{subfigure}{1.0\textwidth}
             \centering
             \includegraphics[width=\textwidth, height=1.1in]{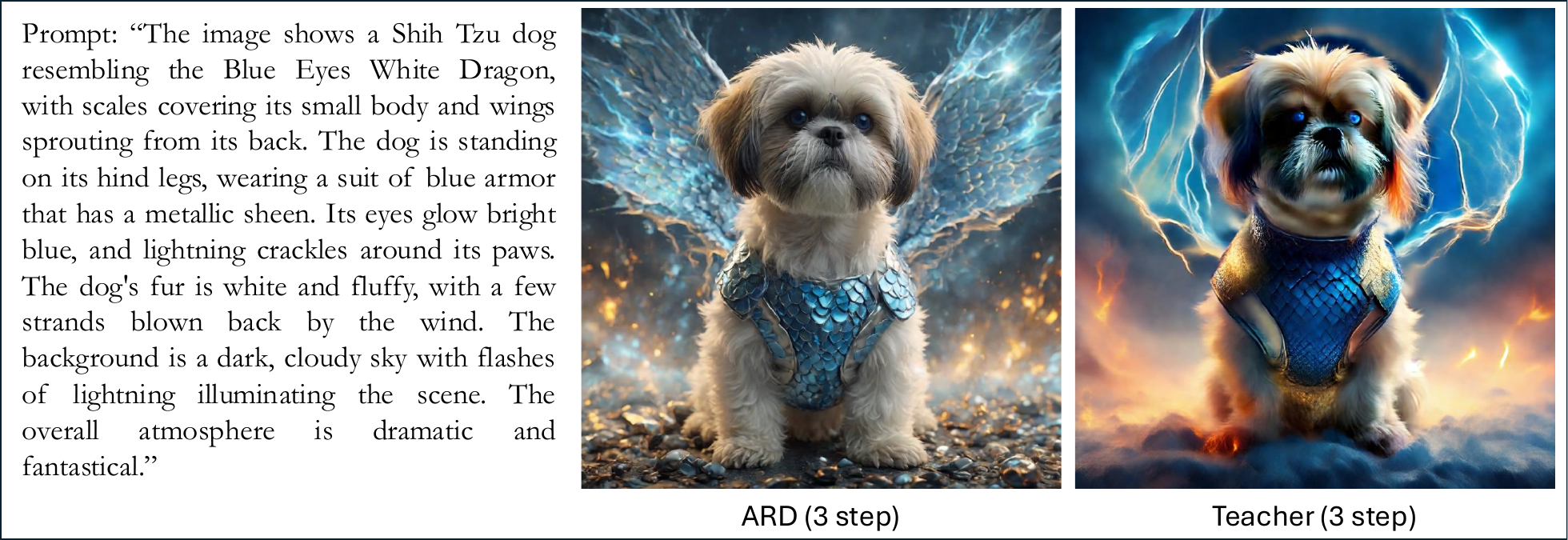}
              \label{fig:7_b}
         \end{subfigure}
    \end{minipage}
    \vspace{-7mm}
    \caption{Sample comparison between ARD and the Emu teacher on long (realistic/unrealistic) prompts.}
            \label{fig:7}
\end{figure*}

ARD with a block-wise causal mask M4 (using the entire trajectory history) outperforms step distillation in all metrics for $2$ and $4$ steps. Unlike step distillation, ARD maintains its Recall as the number of steps increases from 2 to 4. \Cref{fig:6_b,fig:6_c} show that ARD preserves the global structure of the teacher's solution. As a result, ARD improves significantly as the number of steps $S$ increases. 
\Cref{fig:5_a} shows that ARD variants (\cref{eq:ours_loss}) converge more effectively than vanilla regression loss (\cref{eq:step_loss}) during training, demonstrating the benefits of our autoregressive design. Moreover, for a 4-step ARD model the FID degradation from the teacher is $1.43=(4.32-2.89)$, which is 5$\times$ lower than that of step distillation $7.36=(10.25-2.89)$.

When using additional discriminator loss (R+D), the performance of both step distillation and ARD improves, with ARD outperforming step distillation and achieving FID of 1.84. \Cref{fig:6_d} shows that discriminator loss makes the samples sharper while maintaining the global structure from the coupling [$\rvx_{\tau_{S}}, \rvx_{\tau_{0}}$] provided by the teacher. This indicates the additional discriminator loss does not harm the diversity of the samples, and even improves the Recall metrics. The discriminator loss enables ARD to perform better than the teacher, and outperform the public few-step generative models in \Cref{tab:main2} and \cref{fig:2_d} in the low-step regime both in speed and quality. The speed in \cref{fig:2_d} was measured on an NVIDIA H100 with a batch size of 128. For the baselines, we used their official code and measured the speed under the same conditions.


\noindent\textbf{Ablation of the attention mask}
\Cref{tab:main1} shows the results for various attention masks introduced in \cref{fig:3_b} when $S=4$ and $N=6$. The default attention mask M4 exhibits the best FID of 4.32 due to its flexibility. The window attention in M2 and the retention of initial noise in M3 achieve FIDs of 4.75 and 4.45, respectively. These relaxed options, M2 and M3, also show significant gains over step distillation (10.25). Note that both M2 and M3 use two inputs at each step. M2 sets the window size to 2, so two recent denoised samples $[\rvx_{\tau_{s+1}}, \rvx_{\tau_{s}}]$ are used to predict $\rvx_{\tau_{s-1}}$. M3 uses $[\rvx_{\tau_{S}}, \rvx_{\tau_{s}}]$ as inputs. Using the initial noise $\rvx_{\tau_{S}}$ (M3) appears more beneficial as additional information, as it helps maintain a ground truth input signal at every step. The analysis (in \Cref{fig:4c,fig:4e,fig:4g}) of the attention scores with block-wise causal mask M4 show that the initial noise is the most activated among $\rvx_{\tau_{S}:\tau_{s+1}}$, demonstrating the importance of information in it.

\paragraph{Error accumulation ablation}
When we sample with a student model starting from ground truth points on the teacher's trajectory, we can analyze which steps accumulate more errors in the student models. As shown in \cref{fig:5_b}, if the first three steps are solved by the teacher ($75\%$), the performance gap between step distillation and ARD is small. However, if early steps are predicted by the student, the performance of step distillation drops significantly. This suggests that step distillation is more susceptible to exposure bias, whereas ARD is more robust.

\paragraph{Ablation of $N$}
For 4-step ARD model, optimal performance is achieved at $N=6$, as shown in \cref{fig:5_c}. While a large $N$ makes the student model flexible, a small $N$ provides an effective inductive bias. $N=6$  performs well in 2-step cases (see \Cref{tab:main1}) and with different attention masks too. For window attention M2, FID improves from 5.08 to 4.75. For M3, which keeps the initial noise, FID improves from 5.01 to 4.45 as $N$ decreases from 28 to 6. Additionally, smaller $N$ results in faster inference and requires less memory as we need to store kv-cache for fewer layers.

\paragraph{Efficiency analysis}
\Cref{tab:main1,fig:2_c} show the floating-point operations (FLOPs) in the denoising process during the inference phase, indicating the theoretical computational costs. The inference FLOPs for the backbone DiT architecture are 118.6 GFLOPs~\cite{peebles2023scalable}. The FLOPs for step distillation are proportional to the number of steps $S$, and the teacher requires twice the FLOPs due to classifier-free guidance~\cite{ho2022classifier}. ARD requires slightly more FLOPs due to the kv-cache, which increases the number of attended keys and the aggregation of the respective values.  The increased amount is proportional to $N$, which is the number of layers using the attention cache. ARD ($N=6$), which is our best model, only uses 1.1\% more FLOPs compared to the step distillation models. Latency shows a similar trend to FLOPs, except for the teacher.
The attention options M2 and M3 in \cref{fig:3_b} require slightly fewer FLOPs, at 477.1 GFLOPs in the 4-step cases, because the amount of kv-cache is smaller compared to the default attention. These relaxed options can significantly reduce the increase in kv-cache when S becomes larger.


\subsection{Text-conditional image generation}
\label{sec:4.2}

We use 1.7B Emu~\cite{dai2023emu} model with diffusion transformer architecture as a teacher. The teacher model was pre-trained for 1024p resolution on a large-scale internal dataset and fine-tuned on a small set of high-quality aesthetic images. We calculate the teacher ODE trajectories online and use 48 teacher steps by default. Since our distillation requires only the text prompt for training, we use a large-scale internal pre-training dataset for the distillation. For ARD student transformer we opt for a block-wise causal attention M4. We do 15k training iterations using only regression loss.

\paragraph{Evaluation metrics}
To evaluate sample fidelity and diversity, we compute zero-shot FID on the MS-COCO 2017 dataset~\cite{lin2014microsoft} using 5k random real samples and generated samples from 5k random prompts. To measure prompt alignment, we use CompBench~\cite{huang2023t2i}, which has six categories of prompts following the evaluation protocol of Imagine Flash~\cite{kohler2024imagine}. 

\paragraph{Text-Image alignment}
\Cref{tab:main3} shows the text-image alignment scores on CompBench compared to public high-res ($\geq$768 pix) distilled models. Our ARD based on Emu (7.5 CFG) outperforms all other public high-res distilled models in the average score. ARD surpasses all other 1024p distillation models in all six categories except DMD2~\cite{yin2024improved} which shows comparable performance, but uses 0.9B more parameters and more sampling steps. In ARD distilled from Emu (3.0 CFG), the average score gap between the student and the teacher is only 2.3, which is the smallest among all 3-step distillation models (see supplementary for each teacher's performance). ARD can also successfully generate images following long and detailed prompts (see \Cref{fig:7}).

\paragraph{Sample quality} \Cref{tab:main4} shows the FID comparison across $1024$p public distillation methods, which target the teacher's PF-ODE. While the absolute performance of the student (FID-S) is the second best among the models, the best model LCM-LoRA (2.8B) has 1.1B more parameters than ARD. Since the upper-bound performance of the distilled model is the teacher when the distillation targets PF-ODE, its performance highly depends on the teacher's performance (FID-T). \Cref{tab:main4} demonstrates a clear trend: teacher models with larger parameters achieve better performance in FID-T. To quantify the effectiveness of the distillation method, we measure the performance drop, which is the gap between the teacher and the student. ARD exhibits the smallest drop compared to the baselines. We trained step distillation from the same Emu teacher, and ARD still showed better performance, which further validates the benefit of using the trajectory history. 

\Cref{fig:1} shows the generated samples by ARD distilled from Emu (7.5 CFG). ARD generates high-quality images across various topics and styles. The left example in \cref{fig:7} compares samples from ARD and the target teacher with the same initial noise $\rvx_{\tau_{S}}$ and long detailed prompts with realistic contexts (left) and unrealistic contexts (right). The samples are not identical, but the image from ARD maintains high fidelity and effectively preserves text information. The detailed description of the subject and background is well captured in the images.
%
%
The right example in \cref{fig:7} compares samples generated by both teacher and student in 3 steps. Although the number of steps is the same, ARD uses only extra kv-cache, whereas the teacher requires twice more computations due to CFG.

\begin{table}[t]
 \vspace{-0mm}
    \centering
    \caption{Zero-shot FID 5k on MS-COCO 2017 for the public 1024p distillation models. FID-T is the teacher performance, and FID-S is the student performance. The drop indicates the performance gap between the students and the teachers. }
    \vspace{-1.5mm}
    \footnotesize
          \begin{tabular}{lccc|cc}
        \toprule
             Model & Params $\downarrow$ & Steps $\downarrow$ & FID-T & FID-S &  Drop $\downarrow$ \\
            \midrule
            \midrule
          Lightning~\cite{lin2024sdxl}   & 2.6B & 4&24.30&30.16&5.86 \\
          LCM-LoRA~\cite{luo2023lcm}&2.8B&3 &25.11&\textbf{27.77}&2.66 \\
          LCM-LoRA~\cite{luo2023lcm}&1.4B&3 &30.79&33.79&3.00 \\
          Step Distil.~\cite{meng2023distillation}&1.7B&3 &27.97&30.51&\underline{2.54} \\
          \rowcolor{gray!25}ARD&1.7B&3 &27.97&\underline{30.03}&\textbf{2.06} \\
        \bottomrule
    \end{tabular}
    \vspace{-4.5mm}
    \label{tab:main4}
\end{table}

\section{Conclusion}
This paper introduced a novel few-step distillation method for the diffusion transformers
that generalizes step distillation by leveraging the entire historical ODE trajectory in an autoregressive way. We also introduced a modified transformer architecture to support the autoregressive distillation design. The use of ODE trajectory  mitigates exposure bias by maintaining the ground truth input signal at every step. By analyzing the historical trajectory as a better source of coarse-grained information, ARD introduces an additional design choice of using historical trajectory only in lower layers based on attention weight analysis. The empirical results show that ARD outperforms step distillation and surpasses public few-step generative models.




\appendix

\newpage
{
    \small
    \bibliographystyle{ieeenat_fullname}
    \bibliography{main}
}
\clearpage


\onecolumn


\section{Related work}
\subsection{Diffusion acceleration models}

\paragraph{ODE-based acceleration}
A line of studies suggests methods to efficiently obtain the solution of a given PF-ODE by developing numerical solvers~\cite{lu2022dpm,karras2022elucidating,dockhorn2022genie,zhang2023fast} to reduce discretization error under few-step regimes. Another line of research proposes learning a continuous-flow ODE model by rectifying the curvature of the ODE trajectory~\cite{liu2023flow,lipman2023flow,lee2024improving,pmlr-v202-lee23j,liu2024instaflow}, which reduces the number of steps required for solving the ODE. Another approach is to distill a solution of the ODE into few-step student models by utilizing the deterministic nature of the given ODE trajectory, which is most aligned with our work. Some papers suggest learning a coupling of initial noise and samples by utilizing a regression loss between the student's prediction and the teacher's prediction~\cite{luhman2021knowledge,salimans2022progressive,zheng2023fast,kim2024pagoda,kohler2024imagine,meng2023distillation}. While others suggest utilizing a regression loss based on the student's self-consistency~\cite{song2023consistency,kim2024consistency,berthelot2023tract,gu2024datafree,luo2023latent}.

\paragraph{Distribution matching distillation}
This category of studies focuses on distribution matching between the few-step student and teacher distributions without relying on the ODE trajectory. Some methods estimate the score function of the student model during training and ensure that it becomes similar to that of the teacher~\cite{luo2024diff,Yin_2024_CVPR,yin2024improved,xie2024em,luo2024onestep}. Other methods utilize a discriminator and an adversarial loss~\cite{sauer2025adversarial,lin2024sdxl,sauer2024fasthighresolutionimagesynthesis,kang2024diffusion2gan}, but this usually requires an extensive hyperparameter search for stable training. Note that for practical performance gains, a distribution matching loss is often combined with a regression loss in ODE-based distillation methods as well~\cite{kim2024consistency,kohler2024imagine,Yin_2024_CVPR,kim2024pagoda}. These methods require the training of auxiliary neural networks (e.g., student score network, discriminator) for distillation, which demands extra computational resources and memory.

\subsection{Transformer-based visual generative models}
\paragraph{Autoregressive model} The visual autoregressive model was initially applied to pixel space using CNN or RNN architectures~\cite{van2016conditional,van2016pixel}. Subsequently, a series of studies proposed autoregressive modeling in the vector-quantized embedding space with an autoencoding module~\cite{razavi2019generating,esser2021taming,lee2022autoregressive,yu2022scaling}. These models conceptualize an image as a sequence of 1D discrete-value tokens and adopt transformer architectures similar to those used in language models. Further research has expanded autoregressive models to multimodal generation~\cite{sun2024emu,sun2024generative,wang2024emu3} by jointly estimating language and visual tokens – one token at a time. Recently, \cite{tian2024visual} proposed to predict a sequence of 2D token maps instead of the sequence of 1D tokens. This change significantly reduced the length of the predicted sequence, resulting in fewer calls to the model during inference. Our work also works on 2D token maps. However, unlike the approach in~\cite{tian2024visual}, which requires an ad-hoc autoencoder, our sequence of 2D token maps is obtained directly from any pre-trained diffusion transformer.

\paragraph{DART}
DART~\cite{gu2024dartdenoisingautoregressivetransformer} is a concurrent work that proposes an autoregressive model with a sequence of 2D token maps derived from the diffusion process. However, there are fundamental differences with our work. We propose a distillation method (ARD) that allows reducing the number of inference steps down to 3 or 4 following the backward ODE trajectories of a pre-trained diffusion model, in contrast, DART is a vanilla diffusion model that is very inefficient and requires a lot of sampling steps. This results in a significant performance gap in \Cref{tab:main5}, as our model is specifically designed to work in low-step regime. There is also a difference in how a sequence of 2D token maps is formed. In ARD, we directly use the teacher's backward ODE trajectory. While in DART, the trajectory is formed by applying the forward noising process to the ground truth data. 

\begin{table}[h]
 \vspace{-0mm}
    \centering
    \caption{Comparison of ARD and DART on ImageNet 256p.}
          \begin{tabular}{lcccc}
        \toprule
             Model&Steps $\downarrow$ & GFLOPs $\downarrow$ & FID $\downarrow$  \\
            \midrule
            DART~\cite{gu2024dartdenoisingautoregressivetransformer}&16&2157&3.98 \\
          \rowcolor{gray!25}ARD (Ours)&\textbf{4}&\textbf{479.9}&\textbf{1.84} \\
        \bottomrule
    \end{tabular}
    \label{tab:main5}
\end{table}

\paragraph{Masked generative models}
Using the vector-quantized embedding space~\cite{razavi2019generating}, masked prediction is another line of work for generation~\cite{chang2022maskgit,ni2024revisiting,li2023magemaskedgenerativeencoder} similar to BERT~\cite{devlin-etal-2019-bert} in language models. Unlike the autoregressive models, which predict the sequence of tokens in order, this approach starts with a full mask and predicts all masked tokens simultaneously. By repeating the process of re-masking and predicting, it eventually obtains the generated sample.

\section{Experimental details}
\subsection{Class conditional image generation}
We follow the teacher configuration~\cite{peebles2023scalable} for student training, except for gradient clipping and batch size. \Cref{tab:config1} presents the training configuration for the 4-step student model with regression loss. We use a batch size of 128 for the 2-step student model and 256 for the 1-step student model. The student model is initialized from the teacher's weights. By default, we set the prediction target at $s$ as $\mathbb{E}[\rvx_0|\rvx_{\tau_{s}}]$. We use 8 NVIDIA A100 GPUs for training, which takes approximately 2 days. As shown in the blue line of \cref{prediction:imagenet}, the FID almost converges within 16 hours (100k iterations). We applied the same settings to the baseline (step distillation~\cite{salimans2022progressive}) as well. Note that step distillation was not trained using the progressive algorithm proposed in the original paper~\cite{salimans2022progressive}; instead, it was learned directly from the teacher. Please refer to \Cref{sec:2.2} for the objective function.
\begin{table}[h]
    \centering
    \begin{tabular}{ |l|c| } 
    \toprule
    Configuration & Setting \\
    \midrule
    Learning Rate & $10^{-4}$ \\
    Weight Decay & $0.0$ \\
    Gradient Clipping & $1.0$ \\
    Batch Size & $64$\\
    Iterations & $300k$\\
    EMA Decay Rate & $0.9999$ \\
    \bottomrule
    \end{tabular}
    \caption{Details on class conditional generation.}
    \vspace{-3mm}
    \label{tab:config1}
\end{table}

When using an additional discriminator loss, we utilize the teacher network as a feature extractor similar to ~\cite{sauer2024fasthighresolutionimagesynthesis}, and train only the discriminator heads on top of the extracted features from each transformer block~\cite{sauer2021projected}. Discriminator heads predict logits token-wise. We use hinge loss, as described in~\cite{sauer2023stylegan}, and follow the discriminator head architecture proposed in the same work. The discriminator is trained on the final prediction of the student model $\hat{\rvx}_{\tau_{0}} = G_{\boldsymbol{\theta}}(\rvx_{\tau_{S}:\tau_{1}},1)$, and real data. We train it with a learning rate of $1e-3$ and no weight decay. We set adaptive balancing between the regression loss and the discriminator loss following~\cite{esser2021taming}. A batch size of 48 is used for both the student model and the discriminator. By adding a discriminator loss and further finetuning a student model pre-trained with regression loss, we achieve an improvement in FID from $4.32$ to $1.84$ within just 40k iterations.

\subsection{Text conditional image generation}
The Emu teacher model\cite{dai2023emu} has 1.7B parameters, consists of 24 DiT layers, and uses cross-attention layers for text conditioning. Emu is a latent diffusion model, which encodes 1024$\times$1024$\times$3 images in a 128$\times$128$\times$8 latent space. The distillation setup is similar to the procedure in the class-conditional case. We follow the same training configuration as the teacher, except that we adjust the batch size. The student model uses a block-wise causal mask (M4), and $N=1$ as it showed the best performance-quality trade-off. The prediction target is set to $\rvx_{\tau_{s}}$ for fast training. We use 32 H100 GPUs for the student model training. \Cref{tab:prompts_for_fig1} shows the prompts (from left to right, top to bottom) that we used for \cref{fig:1}.
\begin{table}[h!]
    \centering
    \begin{tabular}{|p{16.5cm}|}
        \toprule
        A majestic unicorn prancing through a vibrant field of rainbows and flowers, with its face visible and distinguishable. The unicorn's coat is a shimmering white, with a spiral horn protruding from its forehead and a flowing mane that shimmers in the sunlight. Its legs are slender and powerful, with hooves that barely touch the lush green grass. The rainbows in the field are multicolored, arching up from the ground and swirling around the unicorn's body, while the flowers are a variety of bright colors, blooming in every direction. The atmosphere is one of joy and wonder, as if the unicorn is dancing through a magical paradise.\\
        \midrule
        A scale-up 4k photo-realistic image of a cat wearing a golden crown, with its face visible and distinguishable. The cat is sitting on a chair made by avocado wood, with a soft and luxurious texture. The cat's fur is a vibrant shade of orange, with a few white stripes on its face. The crown is made of pure gold, with intricate carvings and a sparkling gemstone in the center. The cat's eyes are fixed intently on the camera, exuding a sense of confidence and royalty.\\
        \midrule
        A serene scene of a couple of sheep relaxing in a lush green field, with their face visible and distinguishable. The sheep are lying down, with one sheep resting its head on the other's back, conveying a sense of comfort and companionship. The sheep's wool is a soft, fluffy brown color, and their ears are slightly perked up, as if they are enjoying the peaceful atmosphere. The field is filled with wildflowers of various colors, adding a pop of vibrancy to the scene. The sun casts a warm glow over the entire scene, creating long shadows that stretch across the field.\\
        \midrule
        The image shows a corgi sitting on a white background, wearing a bright red bowtie around its neck and a vibrant purple party hat on its head. The bowtie is made of silk and has a subtle sheen to it, while the party hat is adorned with glittery decorations. The corgi's fur is a sandy brown color, and its ears are perked up, as if listening intently. Its big brown eyes are looking directly at the camera, and its tongue is slightly out, as if it's panting happily. The overall atmosphere is festive and playful, suggesting a celebratory occasion.\\
        \midrule
        A very bright and clear 4k close-up image of a tiger visible face and distinguishable features. The background is a mountain with a river flowing through it. The tiger is sitting on a rock, looking directly at the camera, with its arms relaxed and its paws clasped together. The tiger's fur is a soft, golden-brown color, and its eyes are a deep, dark brown. The mountain is a vibrant green, with a few snow-capped peaks visible in the distance. The river is a clear blue, with a few small fish swimming in the water. The overall atmosphere is serene and peaceful, as if the tiger is enjoying a quiet moment in nature.\\
        \midrule
        The image depicts a photo-realistic scene of a dog recklessly driving a go-kart on a winding road. The dog, a sleek black feline with bright eyes, is hunched over the steering wheel, its paws gripping the wheel tightly as it takes a sharp turn. The go-kart is a miniature version of a real race car, with shiny metal bodywork and large wheels. The road is lined with lush greenery and rocks, and the background shows a cloudy sky with a hint of sunlight peeking through. The dog's fur is ruffled by the wind, and its ears are flapping wildly as it speeds along.\\
        \midrule
        A serene landscape with a majestic mountain range in the background, its peaks covered in a blanket of snow. In the foreground, a vibrant meadow is filled with an array of colorful wildflowers, swaying gently in the breeze. A tranquil lake reflects the beauty of the surroundings, its crystal-clear waters glistening in the sunlight. The atmosphere is peaceful, with a sense of harmony between the natural elements, as if nature is offering a sacrifice of its own beauty.\\
        \midrule
        The image shows a vibrant and colorful spread of traditional Chinese cuisine. A steaming plate of dumplings sits alongside a plate of savory Peking duck, complete with crispy skin and tender meat. A bowl of steaming hot noodles is placed nearby, garnished with sliced green onions and a sprinkle of sesame seeds. A small dish of egg rolls and a plate of fresh fruit complete the spread. The food is arranged on a red tablecloth, and the background is a warm and inviting Chinese-style restaurant setting with intricately carved wooden panels and ornate lanterns.\\
        \midrule
        A burger on a wooden board with fries and ketchup. The burger has a sesame seed bun, lettuce, tomato slices, and a beef patty. There are two skewers sticking out of the top of the burger. On the left side of the burger, there is a small white bowl filled with ketchup. To the right of the burger, there are french fries on a wooden board. In front of the burger, there is a knife with a wooden handle laying on its side. Behind the burger, there is a small white bowl filled with mayonnaise. In the background, there is a wooden cutting board with a bowl of cherry tomatoes, a green chili pepper, and a brown cloth.\\
        \bottomrule
    \end{tabular}
    \vspace{-3mm}
    \caption{The prompts used to generate the images for \cref{fig:1}.}
    \label{tab:prompts_for_fig1}
\end{table}

\newpage
\newpage

\section{Additional experimental results}
\subsection{More ablations on the ImageNet 256p}
\begin{figure}[h]
    \centering
    \begin{subfigure}[b]{0.3\textwidth}
     \includegraphics[width=\textwidth, height=1.4in]{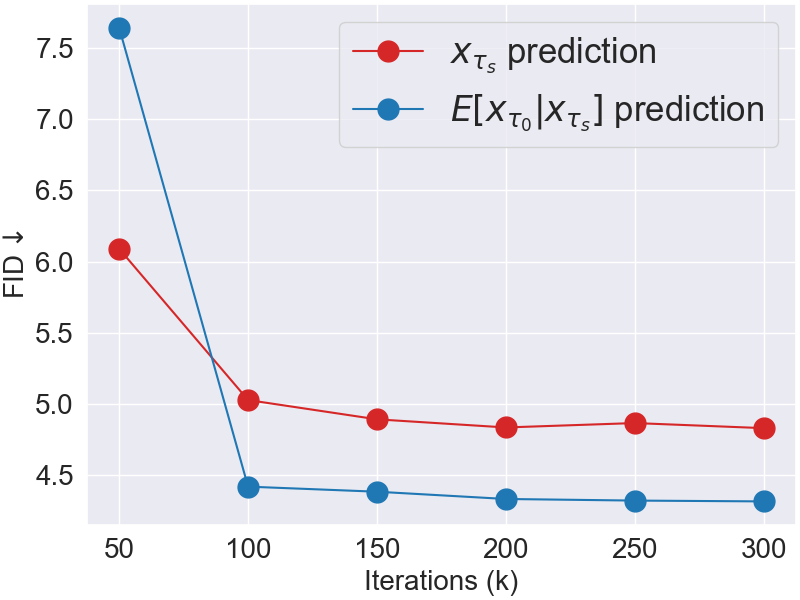}
     \caption{Class-conditional experiments}
     \label{prediction:imagenet}
    \end{subfigure}
    \begin{subfigure}[b]{0.3\textwidth}
     \includegraphics[width=\textwidth, height=1.4in]{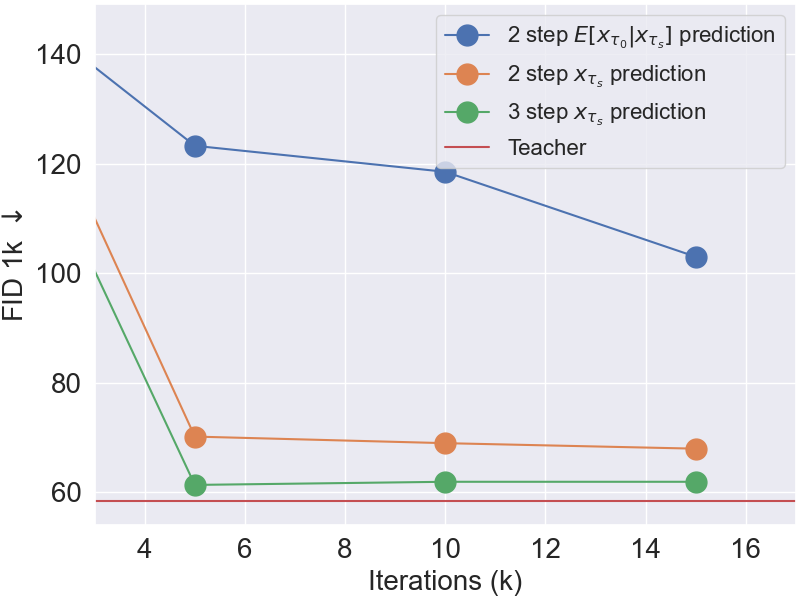}
     \caption{Text-conditional experiments}
     \label{prediction:t2i}
    \end{subfigure}
    \caption{Ablation on the prediction targets.} 
    \label{fig:prediction_abl}
\end{figure}

\paragraph{Prediction targets}
The teacher model provides the ODE path $\rvx_{\tau_{s}}\in$ $[\rvx_{\tau_{S}}, ... ,\rvx_{\tau_{0}}]$. When we solve the teacher ODE, the teacher also provides the equivalent target $\mathbb{E}[\rvx_{\tau_{0}}|\rvx_{\tau_{s}}]$, which is known as ``predicted $\rvx_{\tau_{0}}$"~\cite{songdenoising} at each step $s$. \Cref{prediction:imagenet} shows the ablation of two prediction targets in class-conditional experiments. The convergence of the $\rvx_{\tau_{s}}$ prediction is faster during the first 50k iterations, but with the target $\mathbb{E}[\rvx_{\tau_{0}}|\rvx_{\tau_{s}}]$ the model finally converges to a better local optimum. We suspect that $\mathbb{E}[\rvx_{\tau_{0}}|\rvx_{\tau_{s}}]$ has an advantage in providing fine-grained information as an input because unnecessary noise is removed. The difference in learning speed between the two estimation targets is more pronounced in the text-conditional experiments, as shown in \cref{prediction:t2i}. In our early work, we observed that the $\rvx_{\tau_{s}}$ prediction quickly converged with satisfactory performance in the 2-step experiment, so we chose the $\rvx_{\tau_{s}}$ prediction for the remaining experiments.

\begin{figure*}[h]
    \centering
     \begin{subfigure}[b]{0.24\textwidth}
         \centering
         \includegraphics[width=\textwidth, height=1.1in]{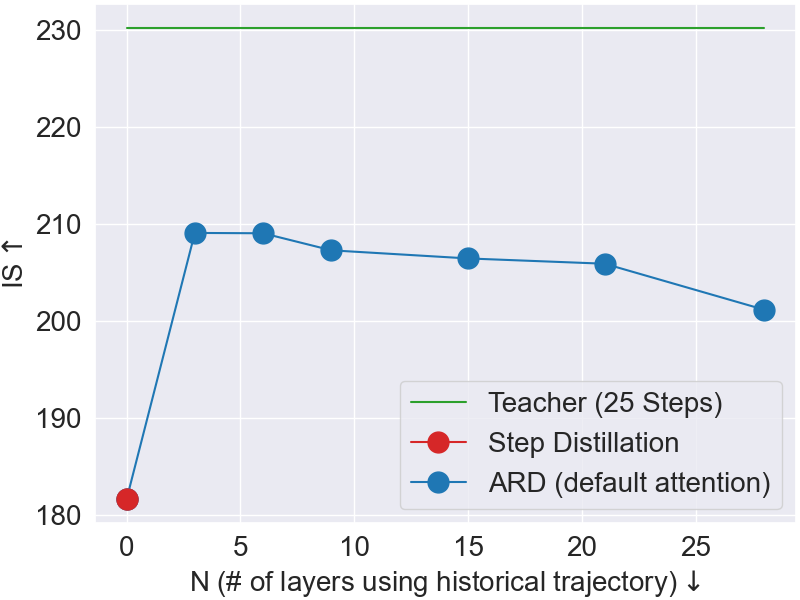}
         \caption{Inception Score}
         \label{fig:more_n_abl_1}
     \end{subfigure}
     \begin{subfigure}[b]{0.24\textwidth}
         \centering
         \includegraphics[width=\textwidth, height=1.1in]{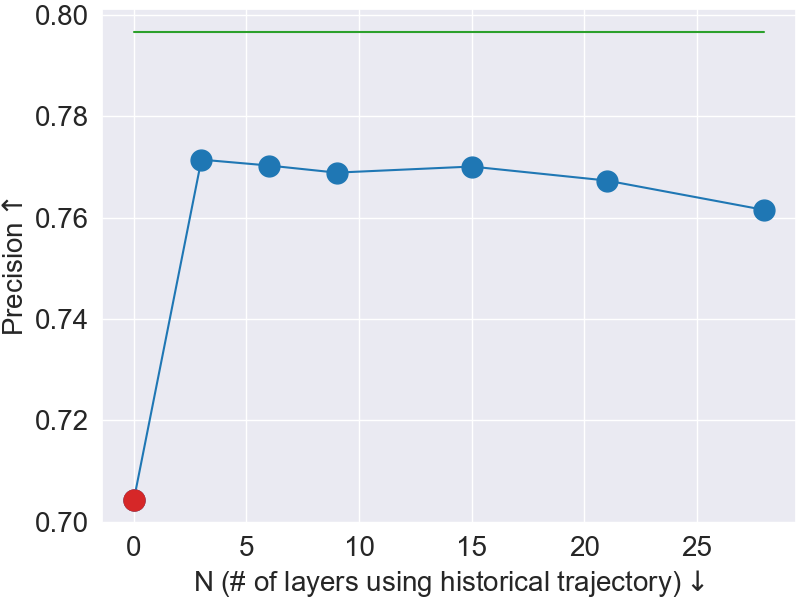}
         \caption{Precision}
         \label{fig:more_n_abl_2}
     \end{subfigure}
    \begin{subfigure}[b]{0.24\textwidth}
         \centering
         \includegraphics[width=\textwidth, height=1.1in]{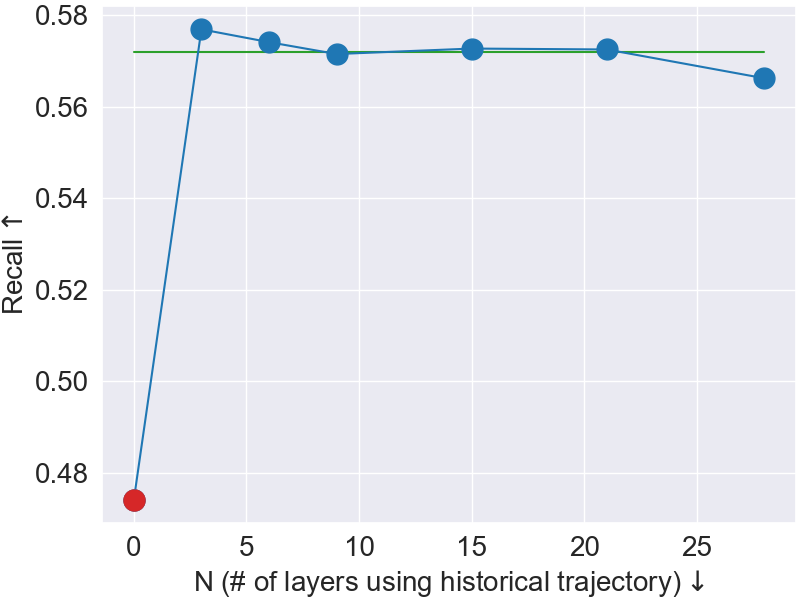}
         \caption{Recall}
         \label{fig:more_n_abl_3}
     \end{subfigure}
     \begin{subfigure}[b]{0.24\textwidth}
         \centering
         \includegraphics[width=\textwidth, height=1.1in]{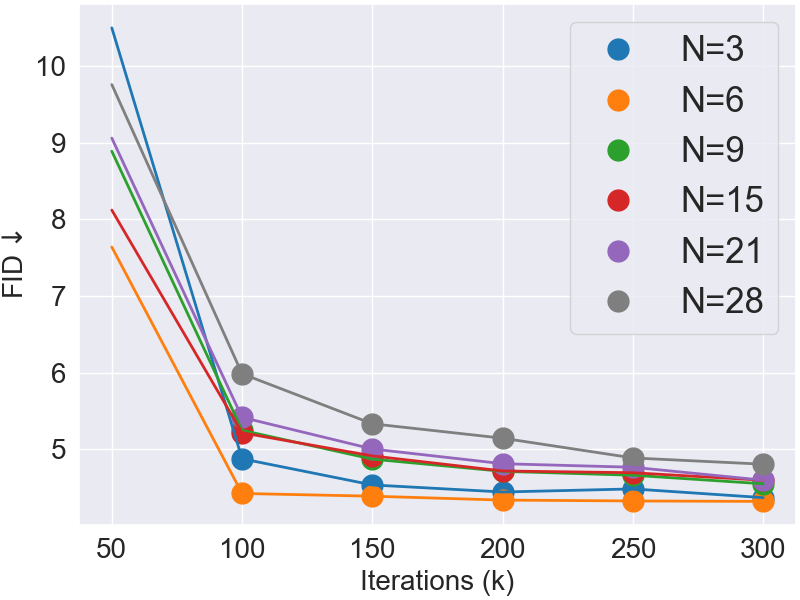}
         \caption{FID per itertions}
         \label{fig:more_n_abl_4}
     \end{subfigure}
    \caption{More evaluation results on $N$ ablations.} 
    \label{fig:more_n_abl}
\end{figure*}

\paragraph{Ablation of the number of layers using historical trajectory ($N$)}
We provide more evaluation results by ablating $N$ for our class-conditional ARD model, where $N$ is the number of layers that use the historical trajectory. \Cref{fig:more_n_abl_1,fig:more_n_abl_2,fig:more_n_abl_3} show the performance of our method for different values of $N$. Each metric shows a similar trend as the FID, as shown in \cref{fig:5_c}. \Cref{fig:more_n_abl_4} shows the training curves for various values of $N$. We find that $N=6$ is the optimal value. The performance advantage is maintained after 100k iterations and $N=6$ not only has a better convergence point but also learns faster.

\newpage
\subsection{Image manipulation}
\begin{figure*}[h]
\centering
     \centering
         \begin{subfigure}{\textwidth}
             \centering
             \includegraphics[width=0.88\textwidth]{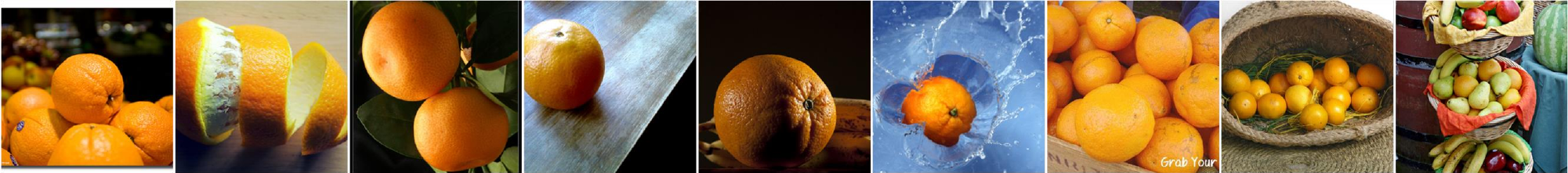}
             \caption{Source images from class \textit{orange}.}
         \end{subfigure}
         \begin{subfigure}{\textwidth}
             \centering
             \includegraphics[width=0.88\textwidth]{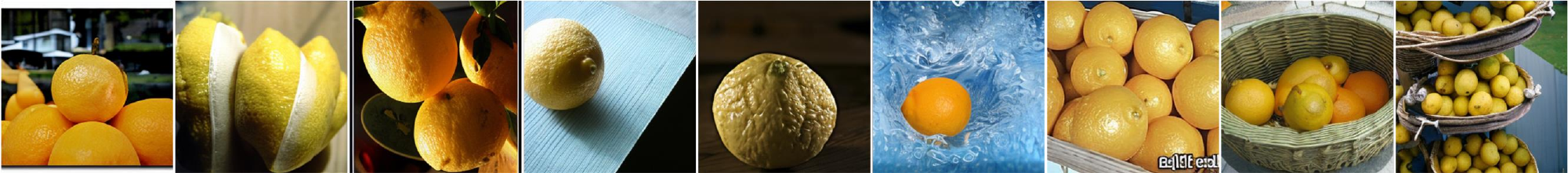}
             \caption{Translated images to class \textit{lemon}.}
         \end{subfigure}
         \begin{subfigure}{\textwidth}
             \centering
             \includegraphics[width=0.88\textwidth]{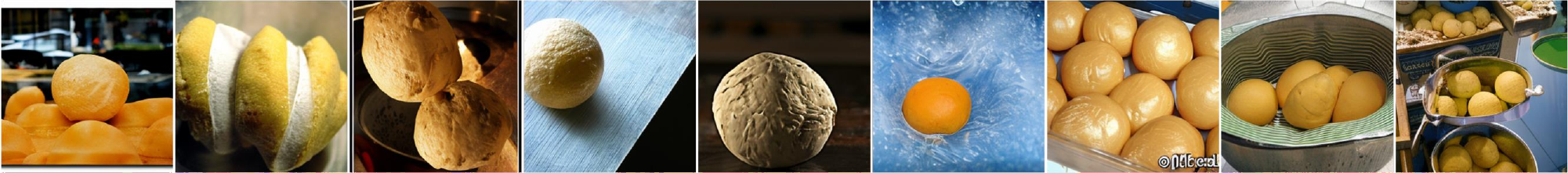}
             \caption{Translated images to class \textit{dough}.}
         \end{subfigure}
          \begin{subfigure}{\textwidth}
             \centering
             \includegraphics[width=0.88\textwidth]{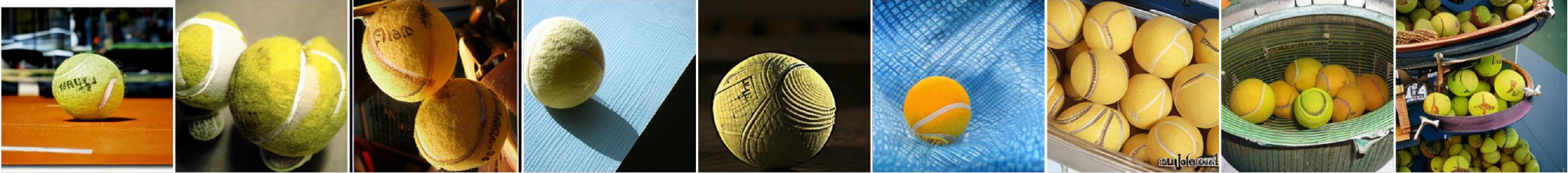}
             \caption{Translated images to class \textit{tennis}.}
         \end{subfigure}
    \caption{Image translation with ARD (R+D) 4-step model.}
    \label{fig:translation}
\end{figure*}

ARD provides image manipulation capability similar to~\cite{meng2022sdedit}. Starting sampling from initial noise $\rvx_{\tau_{S}}$, ARD denoises it to the target class by using the source image $\rvx^{\textrm{src}}$ as input instead of the prediction $\hat{\rvx}_{\tau_{s}}$ at a certain time step $s$ (i.e., $\hat{\rvx}_{\tau_{s-1}}=G_{\boldsymbol{\theta}}([\hat{\rvx}_{\tau_{S}:\tau_{s+1}},\rvx^{\textrm{src}}],s)$). The subsequent sampling process remains the same as before. \Cref{fig:translation} shows the image translation results using 4-step ARD. The prediction of the first step is replaced with the source images and is fed into the ARD model.

\subsection{Reference teacher performance for T2I Image-Text alignment}
\Cref{tab:corresponding_teacher} shows the performance of the corresponding teacher of the 3-step distillation models listed in \Cref{tab:main3}. ARD (distilled from Emu (3.0 CFG)) has a gap of $2.3$ ($=58.5-56.2$) in average score, which is the smallest among all the competitors. Pixart-delta~\cite{chenpixart} (distilled from Pixart-alpha) shows a gap of 4.2, and LCM-LoRA~\cite{luo2023lcm} (distilled from SSD and LDM-XL) shows gaps of 8.0 and 3.3, respectively. All the 768-resolution student models are distilled from Emu (2.7B), and the best model Imagine Flash~\cite{kohler2024imagine} has a gap of $4.9$.

\begin{table*}[h]
 \vspace{-0mm}
    \footnotesize
    \centering
    \caption{The performance on CompBench for the target teachers for each 3-step T2I distillation model.}
    \vspace{-3mm}
          \begin{tabular}{lccc|cccccc|c}
        \toprule
         &  &    & &\multicolumn{4}{c}{\textbf{\quad \quad \quad \quad \quad \quad \quad \quad \quad CompBench $\uparrow$ (\%)}}&& \\
             
             Teacher & Params $\downarrow$ & Res.&CFG & Color &Shape&Texture&Spatial&Non-spatial&Complex & AVG $\uparrow$ \\
            \midrule
          Emu~\cite{dai2023emu}   & 2.7B &768& 6&51.8&39.8&53.5&60.3&67.9&46.0&53.2 \\
          Pixart-alpha~\cite{chen2024pixart}&0.6B &1024&4.5&41.7&39.1&45.9&60.7&62.7&43.0&48.9 \\
          SSD~\cite{SSD}&1.3B &1024&7.5&60.3&44.5&51.4&61.7&65.4&45.3&54.8 \\
         LDM-XL~\cite{podell2024sdxl}&2.6B &1024&7.5&62.8&51.2&54.9&61.3&61.9&43.3&55.9 \\
          \rowcolor{gray!25}Emu~\cite{dai2023emu} (Our teacher) &1.7B &1024&3.0&67.9&50.7&64.5&63.0&59.6&45.4&58.5 \\
          \rowcolor{gray!25}Emu~\cite{dai2023emu} (Our teacher)&1.7B &1024&7.5&72.6&55.4&70.4&69.6&64.4&49.9&63.7 \\
        \bottomrule
    \end{tabular}
    \label{tab:corresponding_teacher}
\end{table*}

\subsection{Additional samples}
\begin{figure*}[h]
\centering
     \centering
         \begin{subfigure}{\textwidth}
             \centering
             \includegraphics[width=0.88\textwidth]{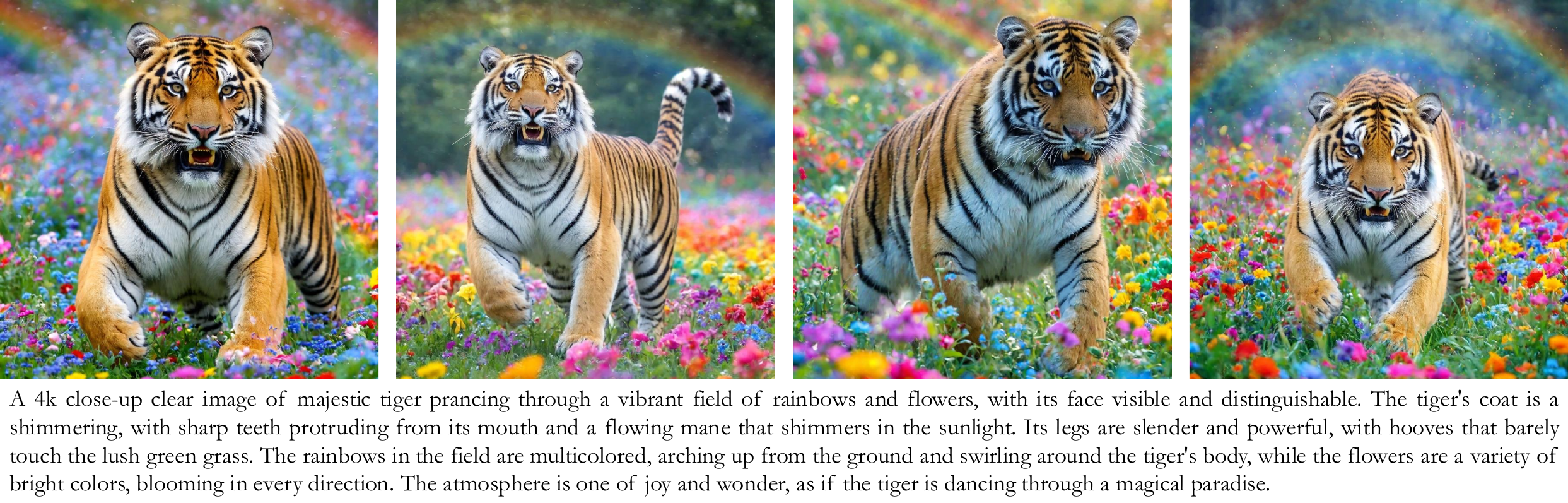}
         \end{subfigure}
         \vspace{3mm}
         \begin{subfigure}{\textwidth}
             \centering
             \includegraphics[width=0.88\textwidth]{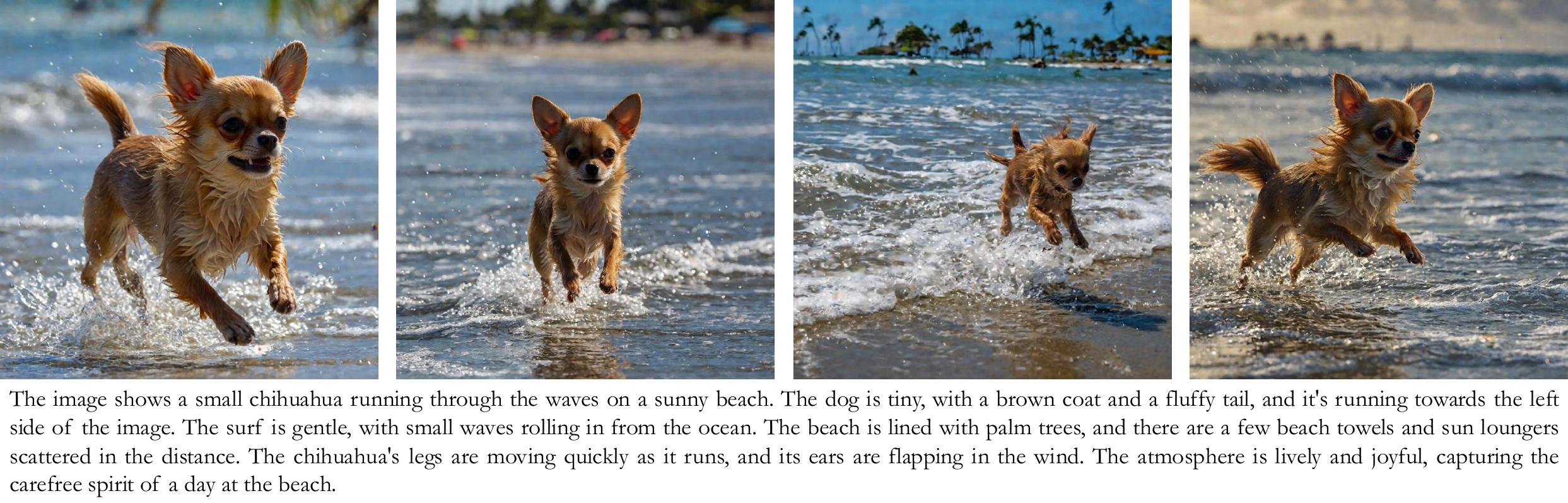}
         \end{subfigure}
         \vspace{3mm}
         \begin{subfigure}{\textwidth}
             \centering
             \includegraphics[width=0.88\textwidth]{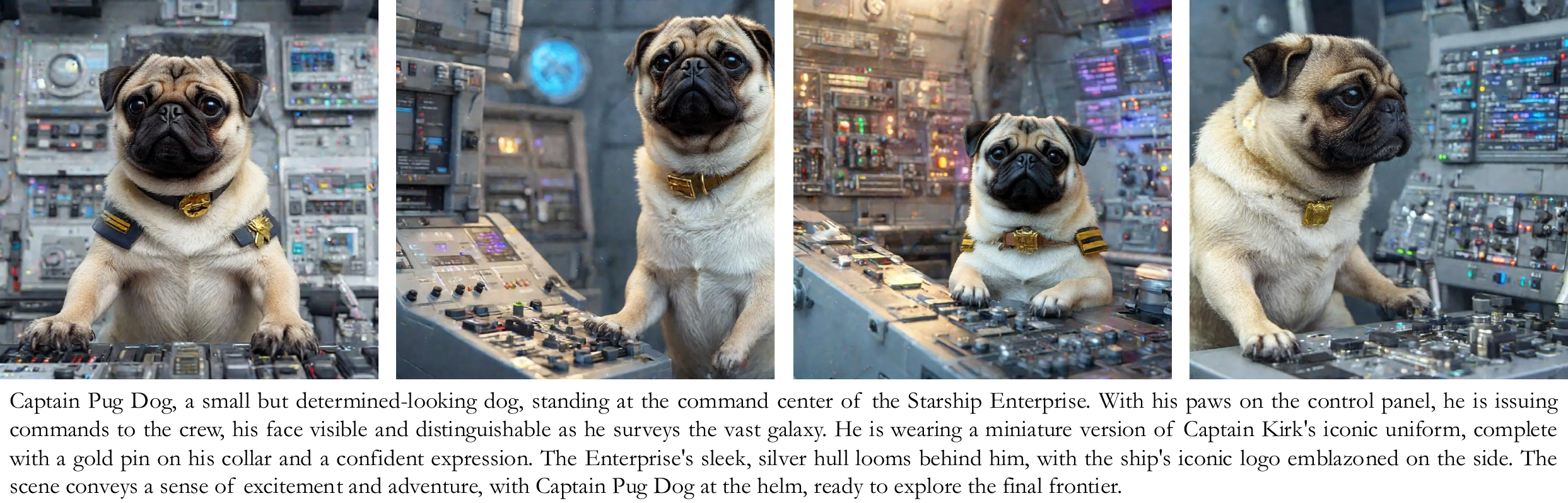}
         \end{subfigure}
        \vspace{3mm}
          \begin{subfigure}{\textwidth}
             \centering
             \includegraphics[width=0.88\textwidth]{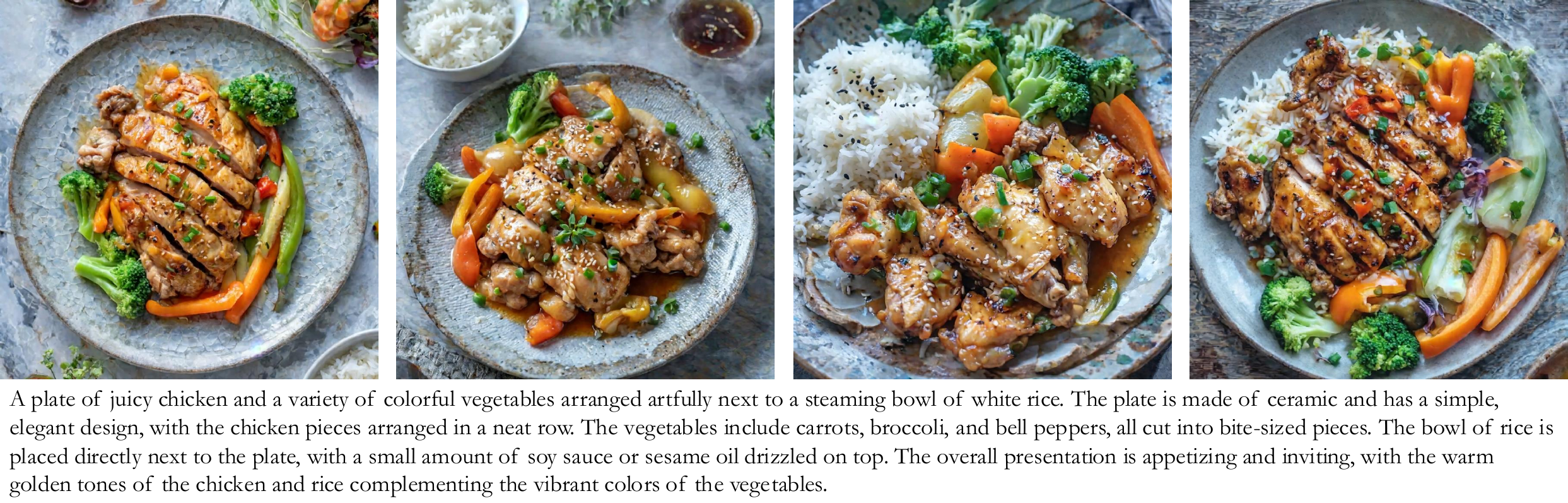}
         \end{subfigure}
    \caption{Samples generated by our 3-step ARD model, distilled from a 1.7B Emu.}
    \label{fig:various_rand}
            \vspace{-0mm}
\end{figure*}

\begin{figure*}[h]
\centering
     \centering
         \begin{subfigure}{\textwidth}
             \centering
             \includegraphics[width=1.0\textwidth]{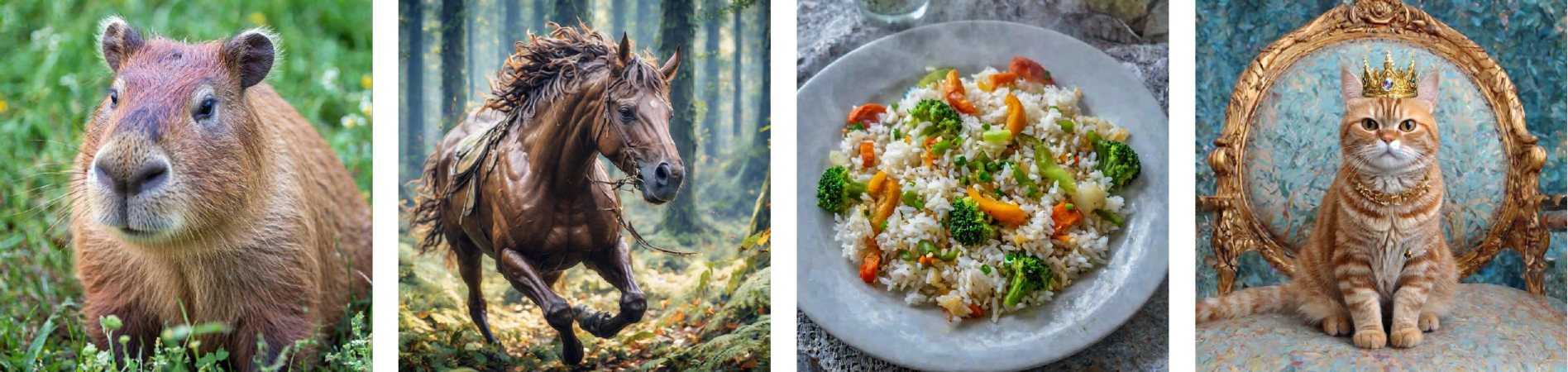}
             \caption{Samples from 3-step ARD}
         \end{subfigure}
         \begin{subfigure}{\textwidth}
             \centering
             \includegraphics[width=1.0\textwidth]{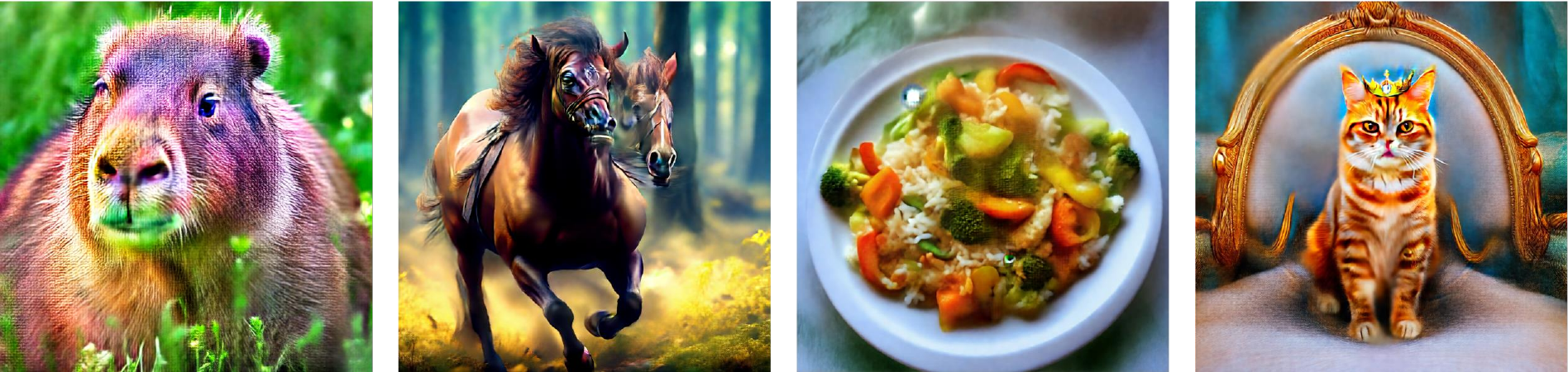}
             \caption{Samples from the Emu teacher with 3-step}
         \end{subfigure}
    \caption{Comparison of generations between ARD and the Emu teacher.}
    \label{fig:more_compare}
            \vspace{-0mm}
\end{figure*}

\begin{figure*}[h]
\centering
     \begin{minipage}[h]{\textwidth}
     \centering
         \begin{subfigure}{\textwidth}
             \centering
             \includegraphics[width=0.88\textwidth]{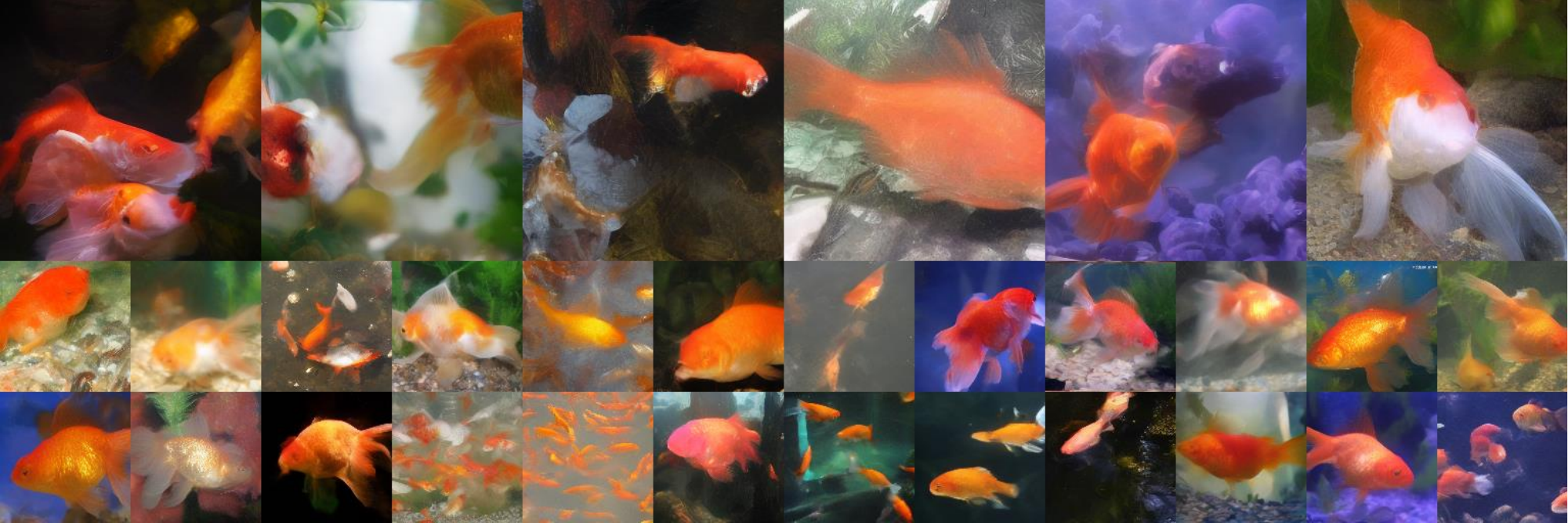}
             \caption{Step Distillation (R) / FID: 10.25}
         \end{subfigure}
    \end{minipage}
    \begin{minipage}[h]{\textwidth}
         \begin{subfigure}{\textwidth}
             \centering
             \includegraphics[width=0.88\textwidth]{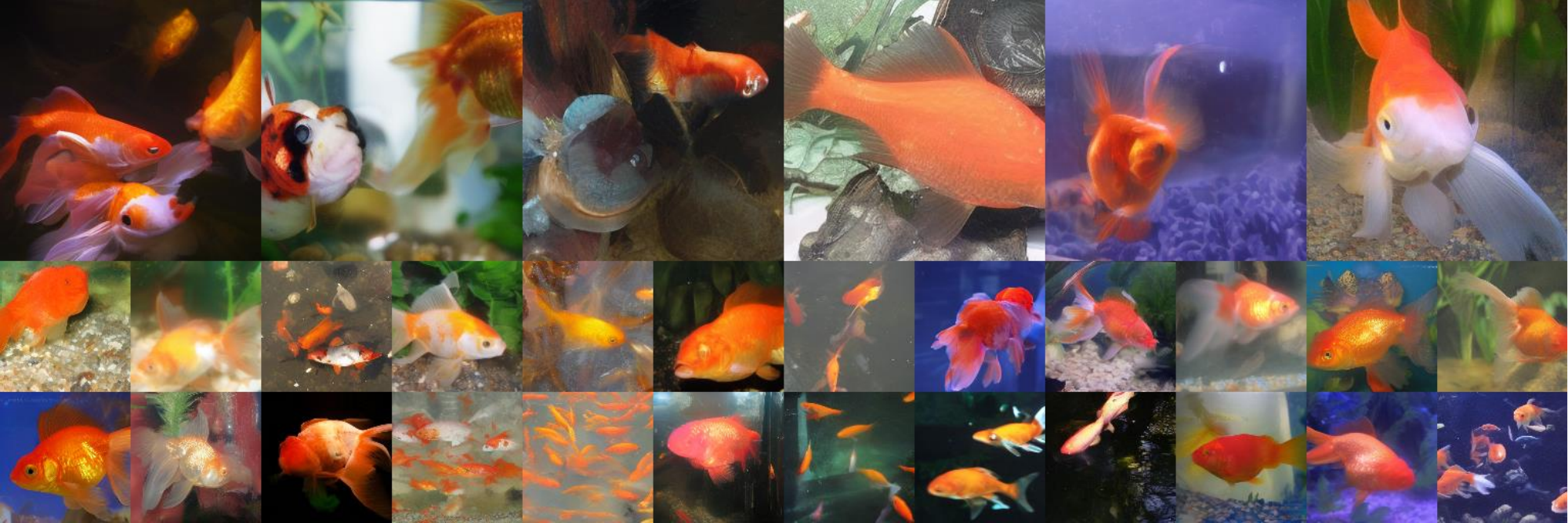}
             \caption{ARD (R) / FID: 4.32} 
         \end{subfigure}
    \end{minipage}
    \begin{minipage}[h]{\textwidth}
        \begin{subfigure}{\textwidth}
             \centering
             \includegraphics[width=0.88\textwidth]{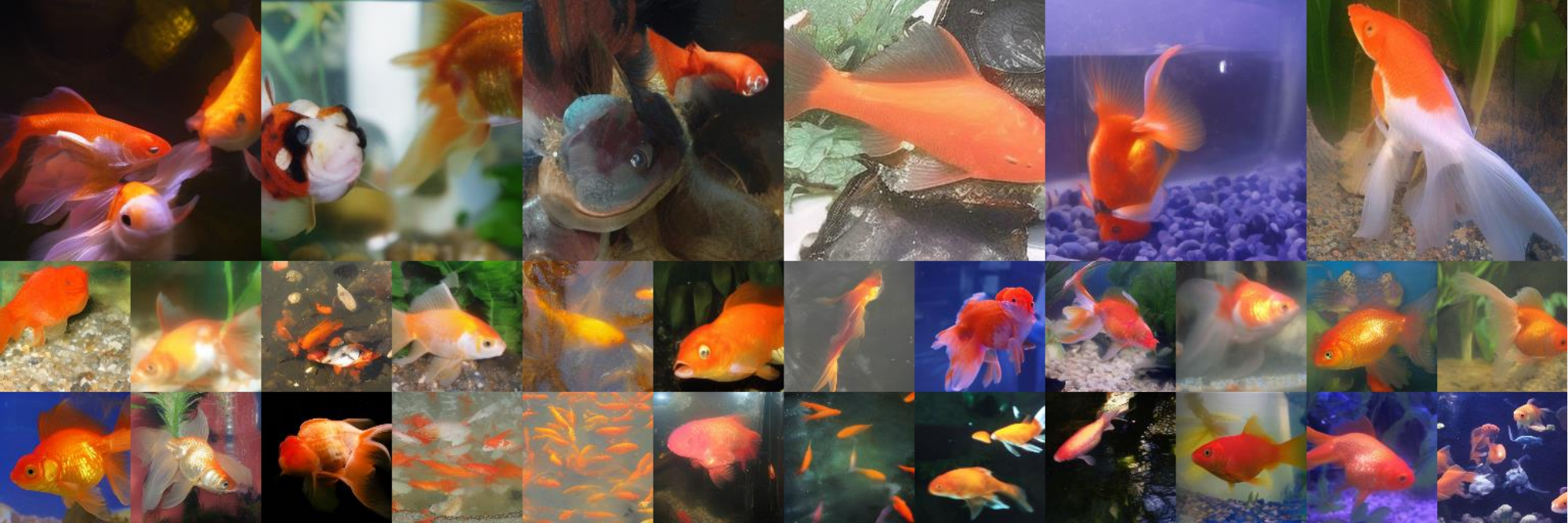}
             \caption{Teacher (25 steps) / FID: 2.89}
         \end{subfigure}
        \end{minipage}
    \begin{minipage}[h]{\textwidth}
         \begin{subfigure}{\textwidth}
             \centering
             \includegraphics[width=0.88\textwidth]{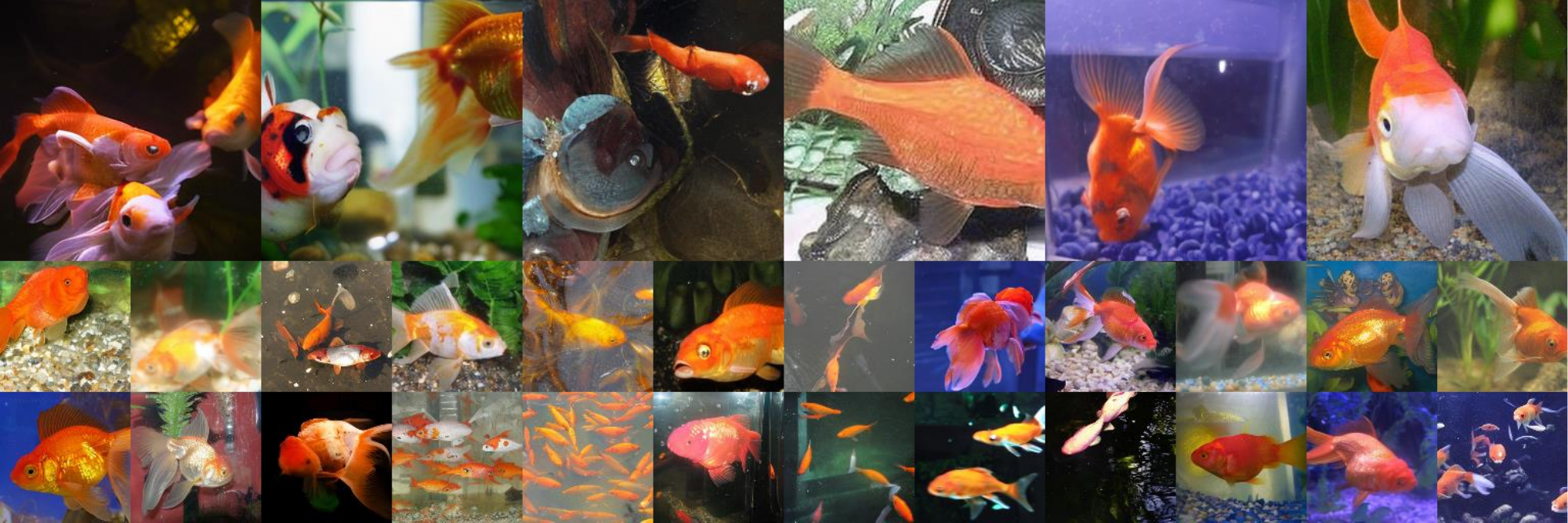}
             \caption{ARD (R+D) / FID: 1.84} 
         \end{subfigure}
    \end{minipage}
    \caption{Randomly generated ImageNet 256p samples for class \textit{goldfish}. All distilled models are 4-step models.}
    \label{fig:img_1}
            \vspace{-0mm}
\end{figure*}

\begin{figure*}[h]
\centering
     \begin{minipage}[h]{\textwidth}
     \centering
         \begin{subfigure}{\textwidth}
             \centering
             \includegraphics[width=0.88\textwidth]{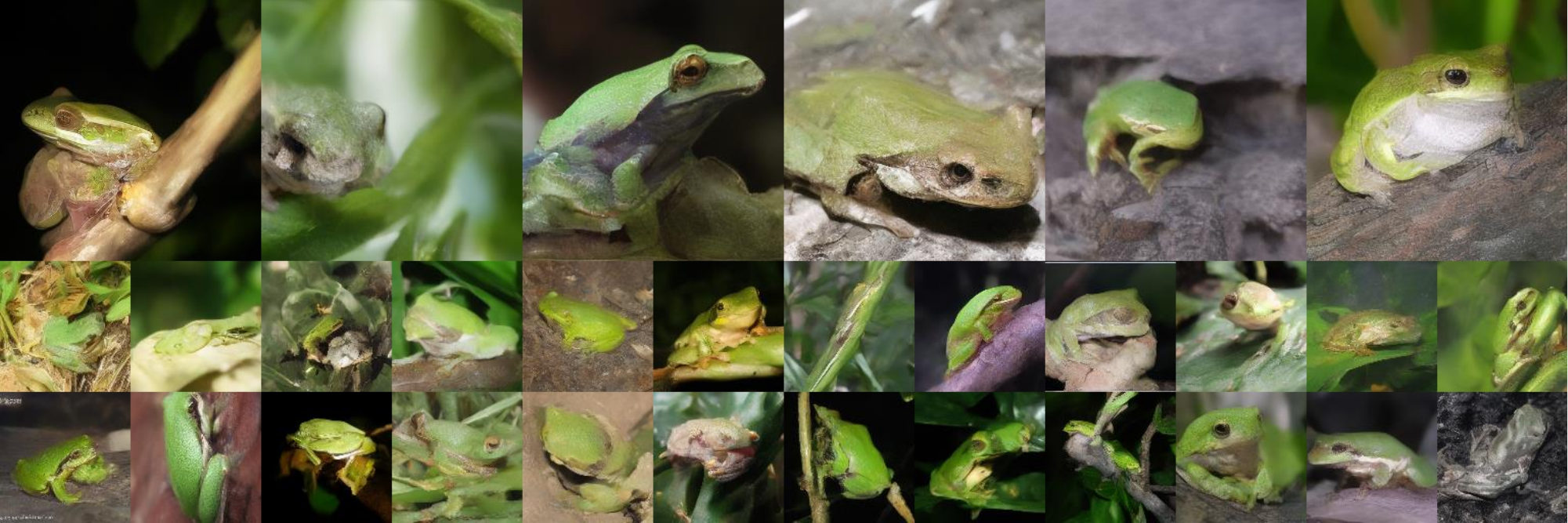}
             \caption{Step Distillation (R) / FID: 10.25}
         \end{subfigure}
         \begin{subfigure}{\textwidth}
             \centering
             \includegraphics[width=0.88\textwidth]{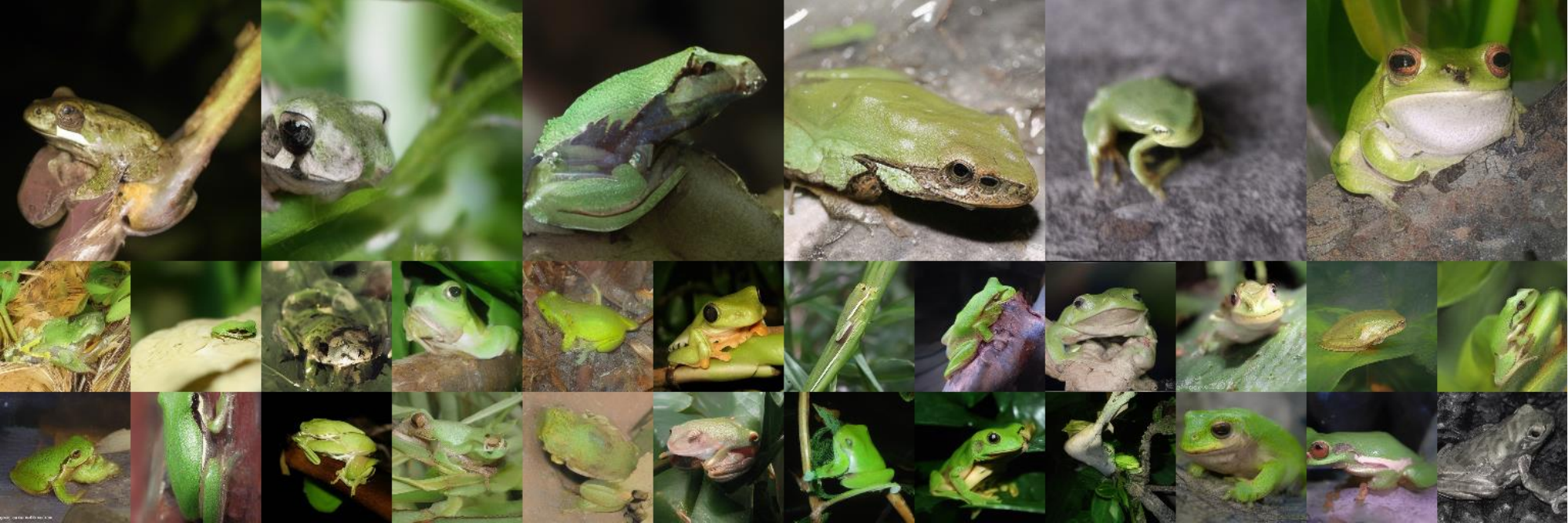}
             \caption{ARD (R) / FID: 4.32} 
         \end{subfigure}
    \end{minipage}
    \begin{minipage}[h]{\textwidth}
        \begin{subfigure}{\textwidth}
             \centering
             \includegraphics[width=0.88\textwidth]{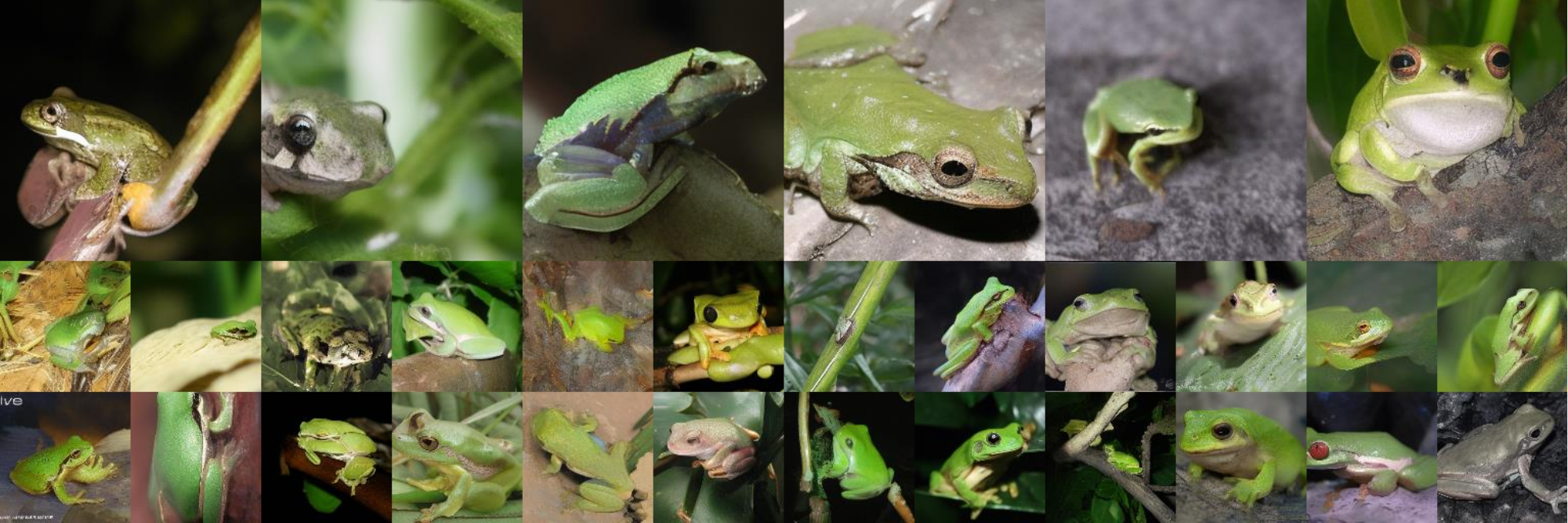}
             \caption{Teacher (25 steps) / FID: 2.89}
         \end{subfigure}
         \begin{subfigure}{\textwidth}
             \centering
             \includegraphics[width=0.88\textwidth]{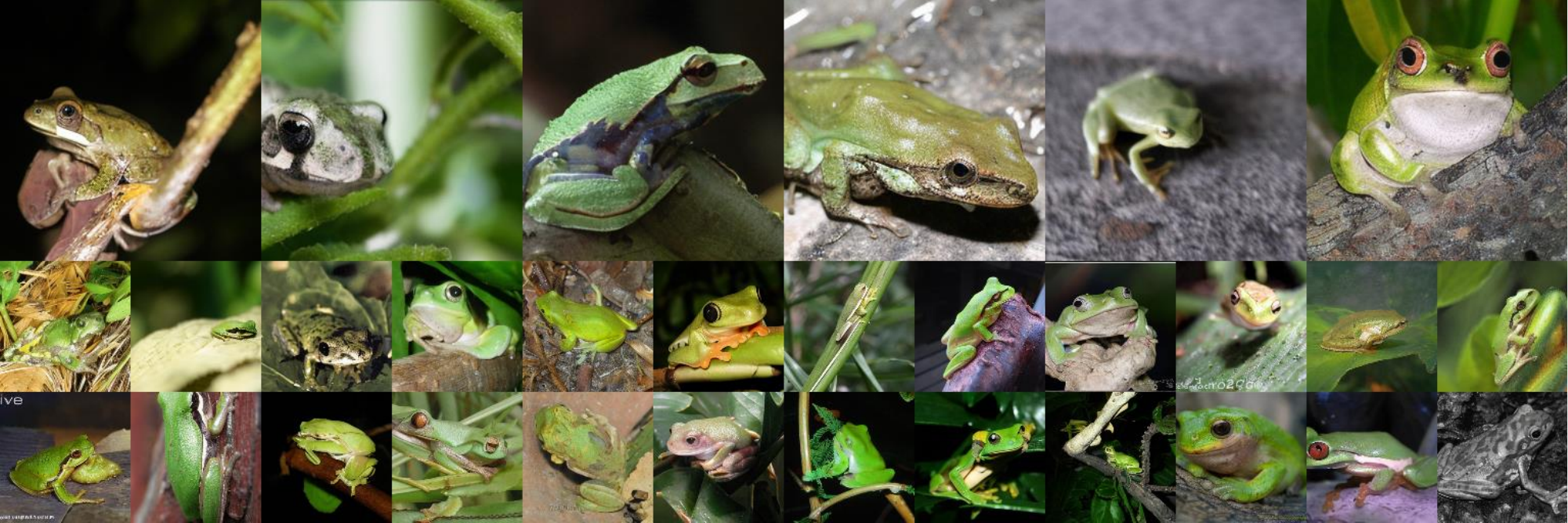}
             \caption{ARD (R+D) / FID: 1.84} 
         \end{subfigure}
    \end{minipage}
    \caption{Randomly generated ImageNet 256p samples for class \textit{tree frog}. All distilled models are 4-step models.}
    \label{fig:img_2}
            \vspace{-0mm}
\end{figure*}

\begin{figure*}[h]
\centering
     \begin{minipage}[h]{\textwidth}
     \centering
         \begin{subfigure}{\textwidth}
             \centering
             \includegraphics[width=0.88\textwidth]{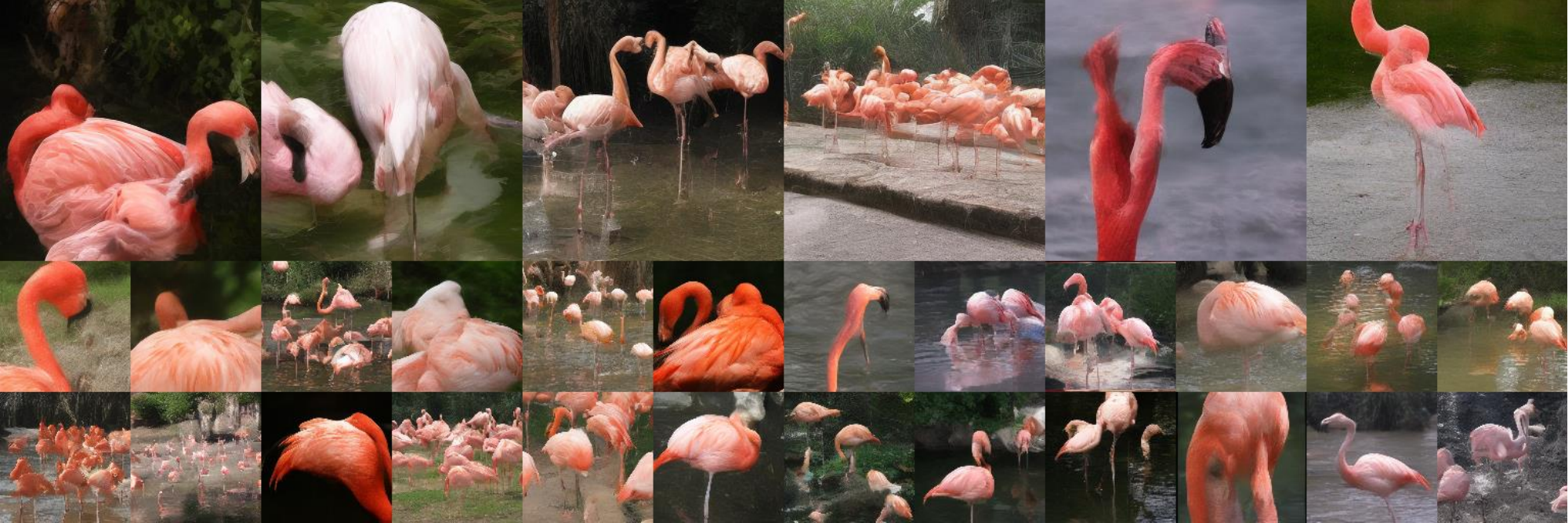}
             \caption{Step Distillation (R) / FID: 10.25}
         \end{subfigure}
         \begin{subfigure}{\textwidth}
             \centering
             \includegraphics[width=0.88\textwidth]{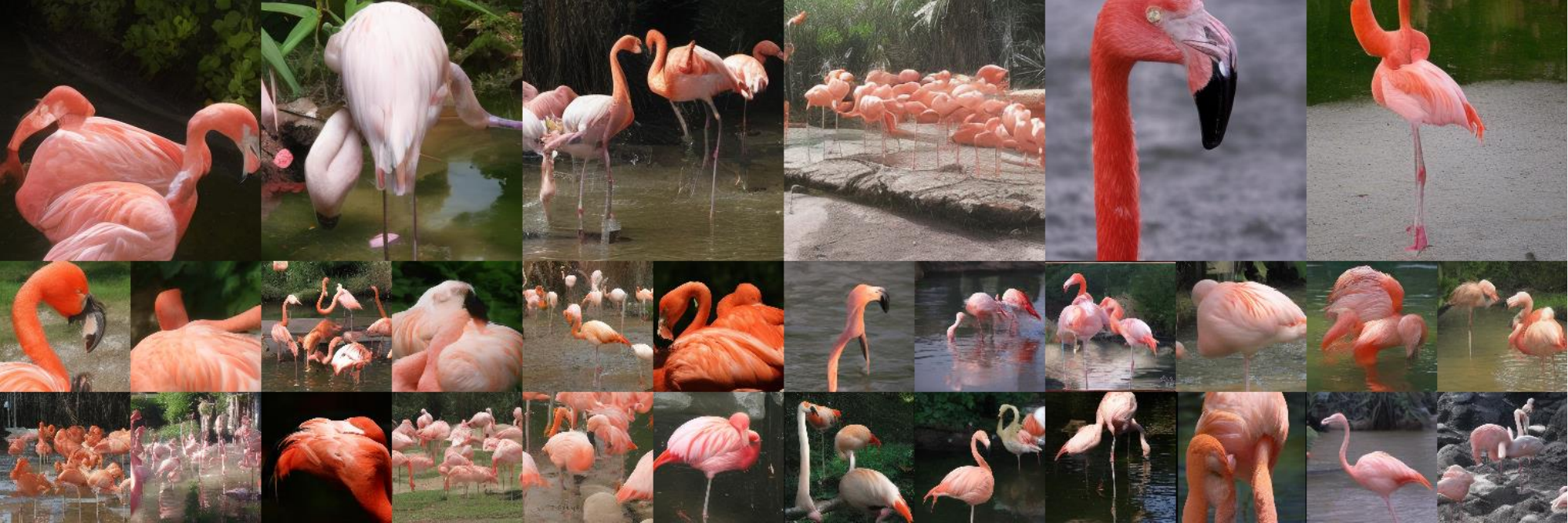}
             \caption{ARD (R) / FID: 4.32} 
         \end{subfigure}
    \end{minipage}
    \begin{minipage}[h]{\textwidth}
        \begin{subfigure}{\textwidth}
             \centering
             \includegraphics[width=0.88\textwidth]{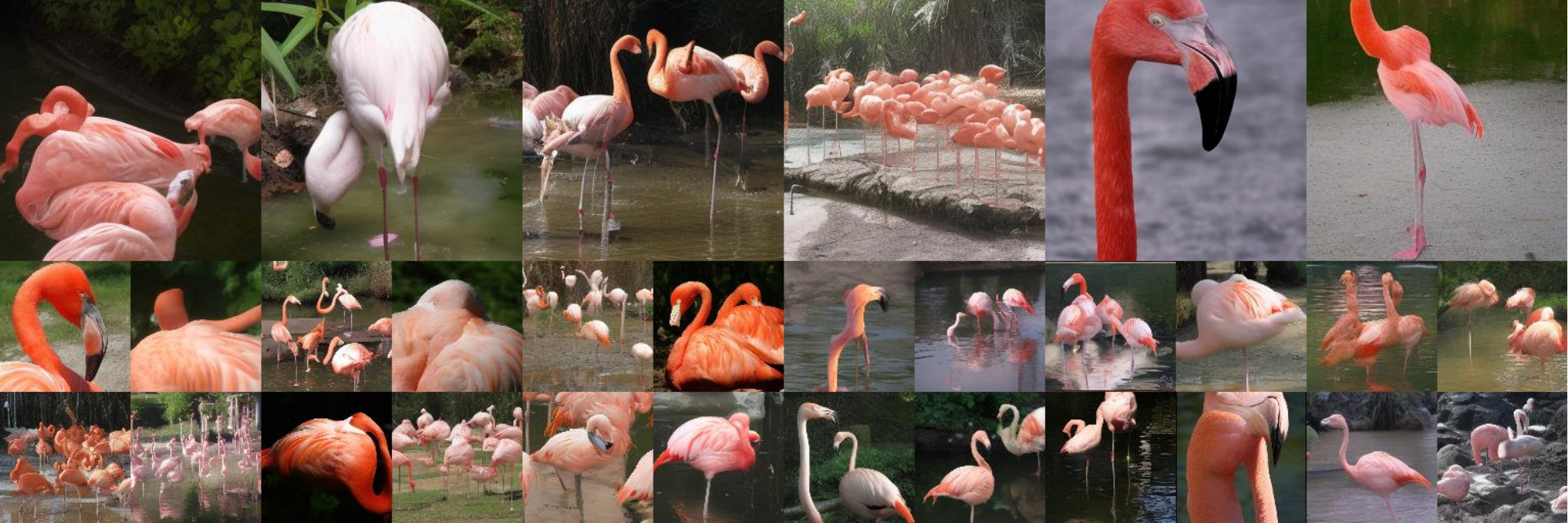}
             \caption{Teacher (25 steps) / FID: 2.89}
         \end{subfigure}
         \begin{subfigure}{\textwidth}
             \centering
             \includegraphics[width=0.88\textwidth]{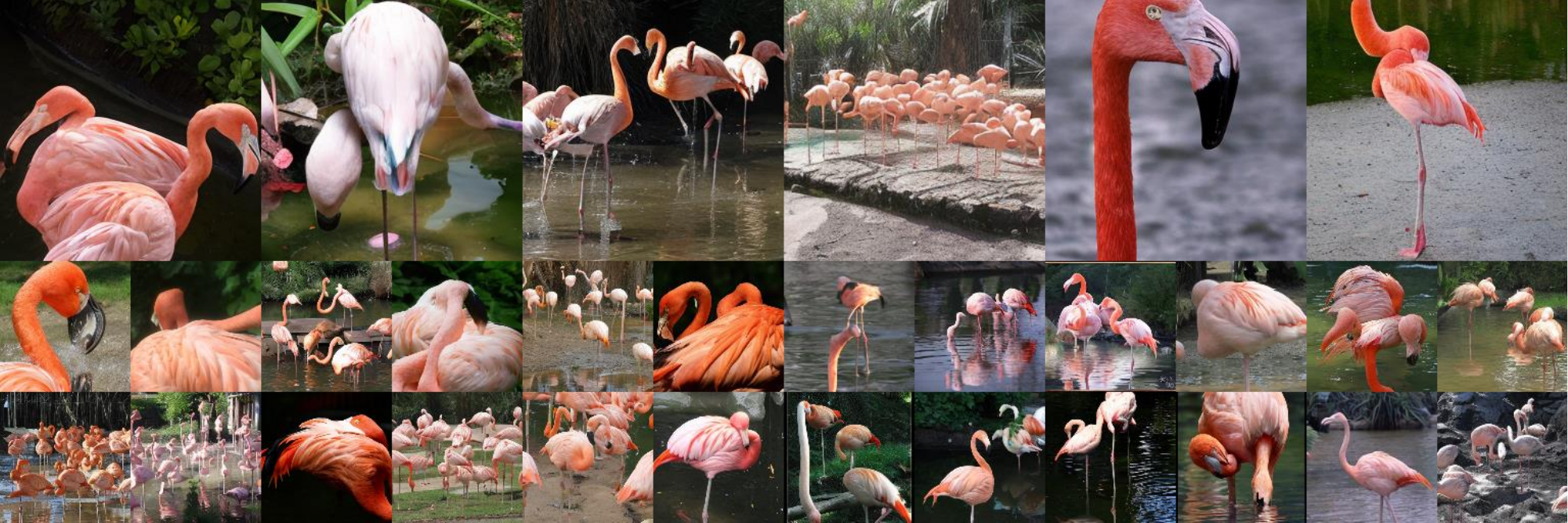}
             \caption{ARD (R+D)  / FID: 1.84} 
         \end{subfigure}
    \end{minipage}
    \caption{Randomly generated ImageNet 256p samples for class \textit{flamingo}. All distilled models are 4-step models.}
    \label{fig:img_3}
            \vspace{-0mm} 
\end{figure*}

\begin{figure*}[h]
\centering
     \begin{minipage}[h]{\textwidth}
     \centering
         \begin{subfigure}{\textwidth}
             \centering
             \includegraphics[width=0.88\textwidth]{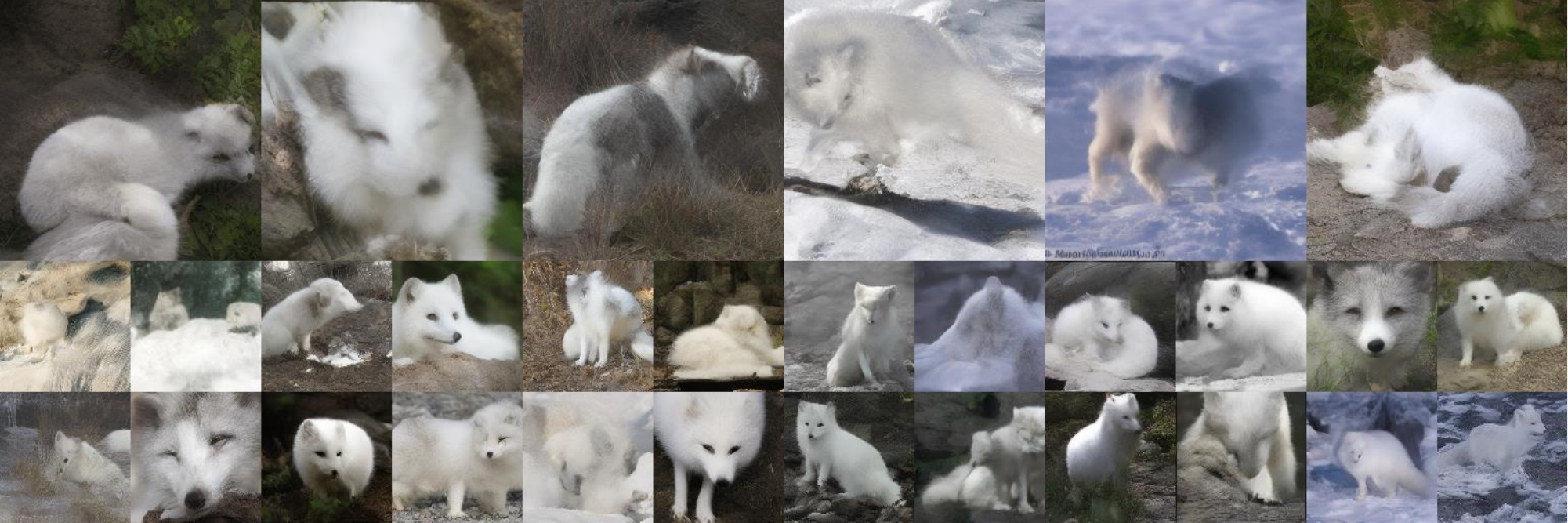}
             \caption{Step Distillation (R) / FID: 10.25}
         \end{subfigure}
         \begin{subfigure}{\textwidth}
             \centering
             \includegraphics[width=0.88\textwidth]{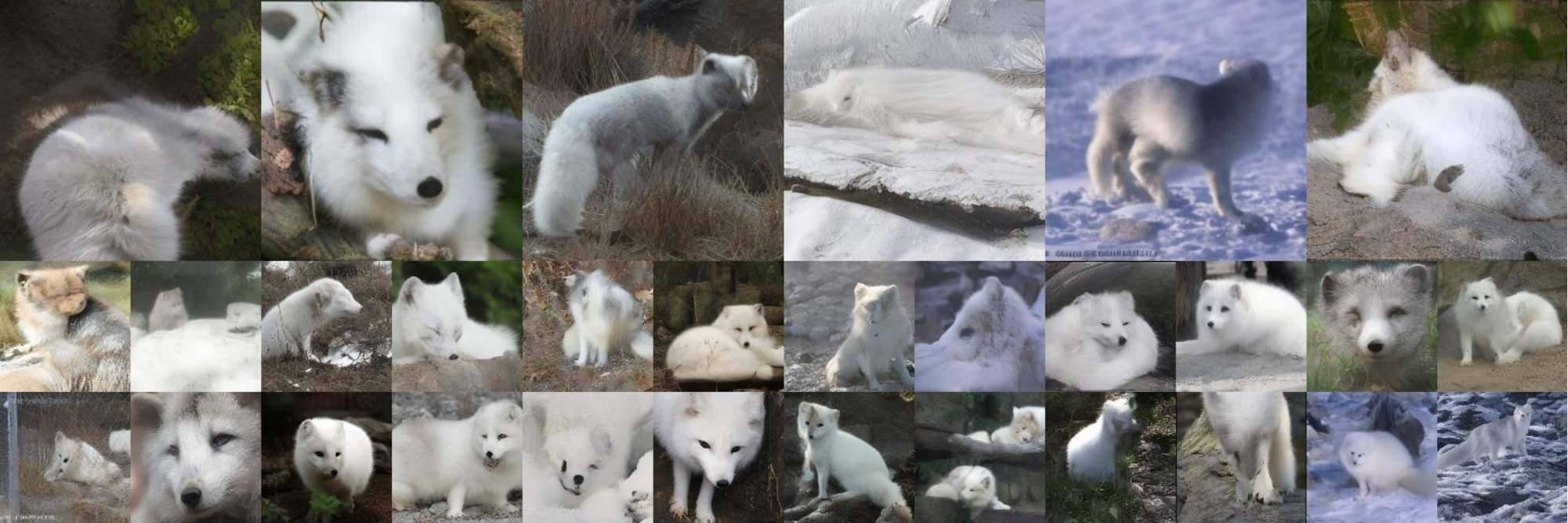}
             \caption{ARD (R) / FID: 4.32} 
         \end{subfigure}
    \end{minipage}
    \begin{minipage}[h]{\textwidth}
        \begin{subfigure}{\textwidth}
             \centering
             \includegraphics[width=0.88\textwidth]{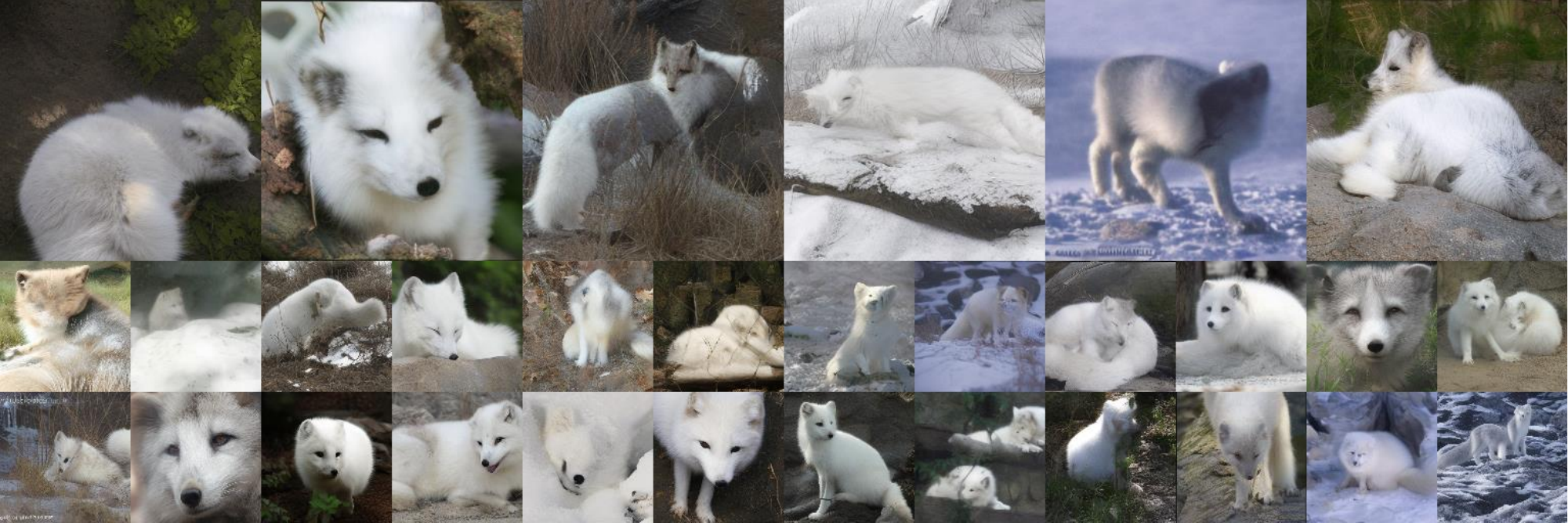}
             \caption{Teacher (25 steps) / FID: 2.89}
         \end{subfigure}
         \begin{subfigure}{\textwidth}
             \centering
             \includegraphics[width=0.88\textwidth]{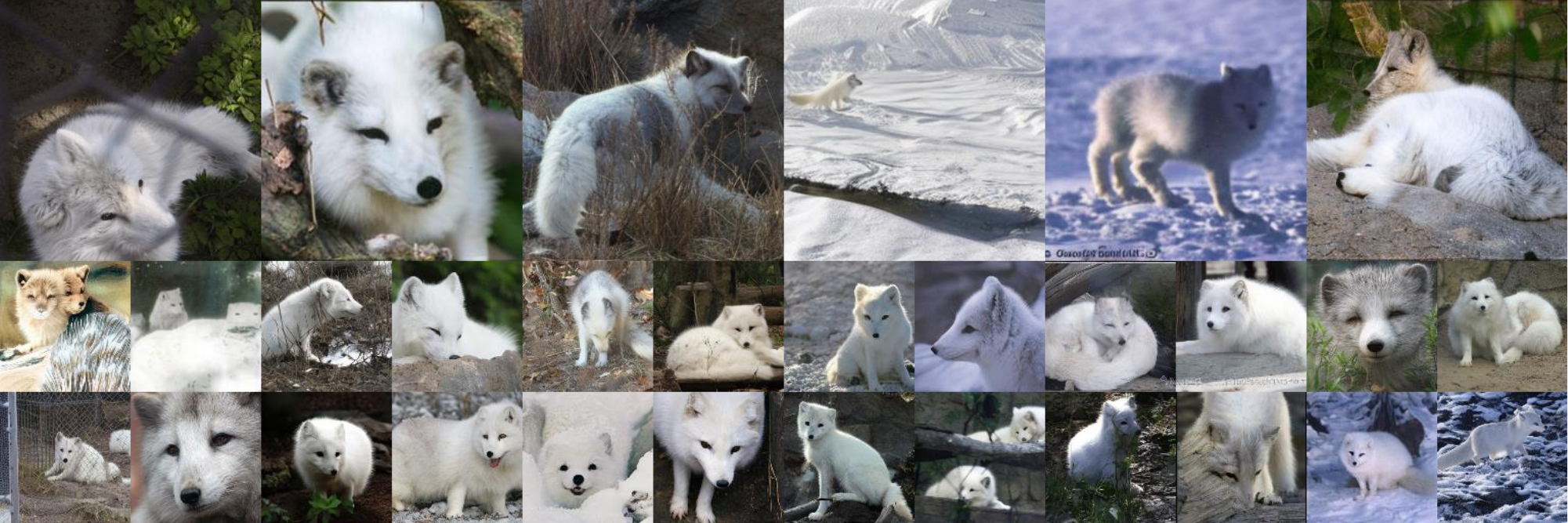}
             \caption{ARD (R+D) / FID: 1.84} 
         \end{subfigure}
    \end{minipage}
    \caption{Randomly generated ImageNet 256p samples for class \textit{Arctic fox}. All distilled models are 4-step models.}
    \label{fig:img_4}
            \vspace{-0mm}
\end{figure*}

\begin{figure*}[h]
\centering
     \begin{minipage}[h]{\textwidth}
     \centering
         \begin{subfigure}{\textwidth}
             \centering
             \includegraphics[width=0.88\textwidth]{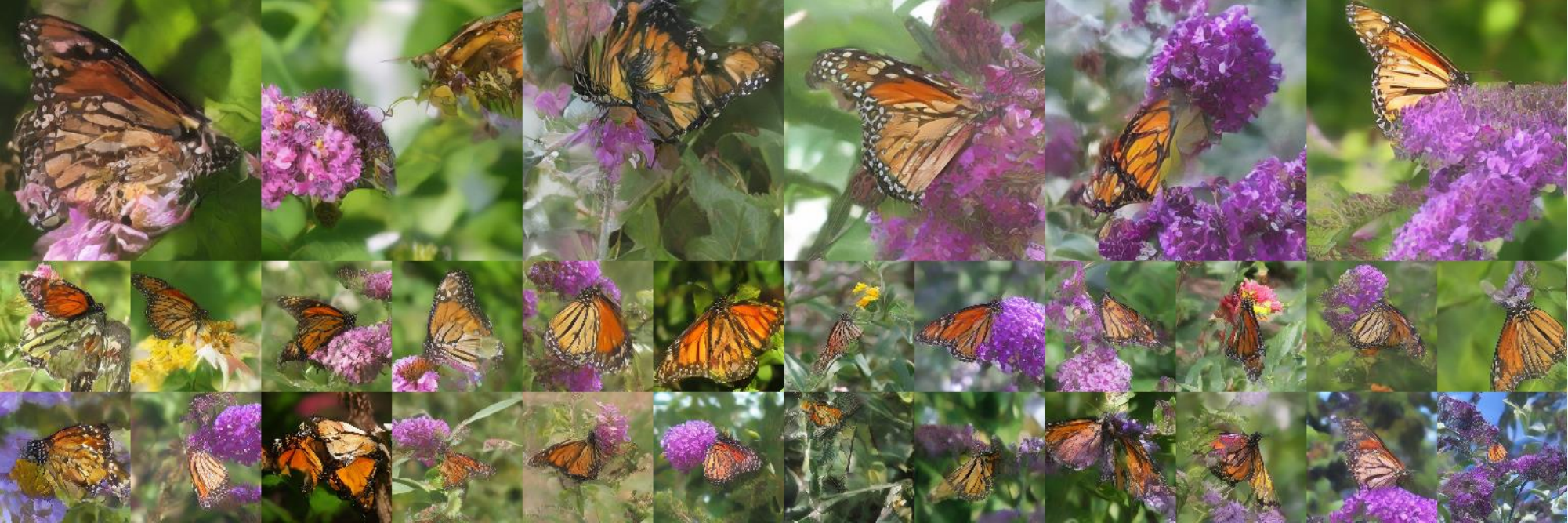}
             \caption{Step Distillation (R) / FID: 10.25}
         \end{subfigure}
         \begin{subfigure}{\textwidth}
             \centering
             \includegraphics[width=0.88\textwidth]{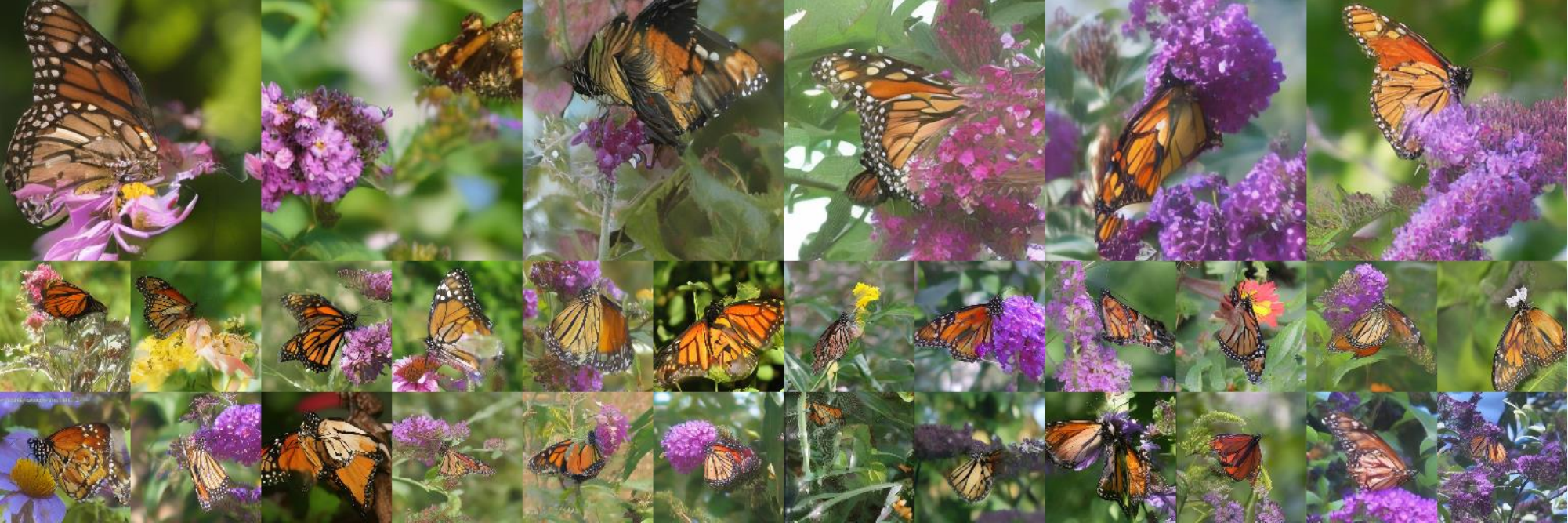}
             \caption{ARD (R) / FID: 4.32} 
         \end{subfigure}
    \end{minipage}
    \begin{minipage}[h]{\textwidth}
        \begin{subfigure}{\textwidth}
             \centering
             \includegraphics[width=0.88\textwidth]{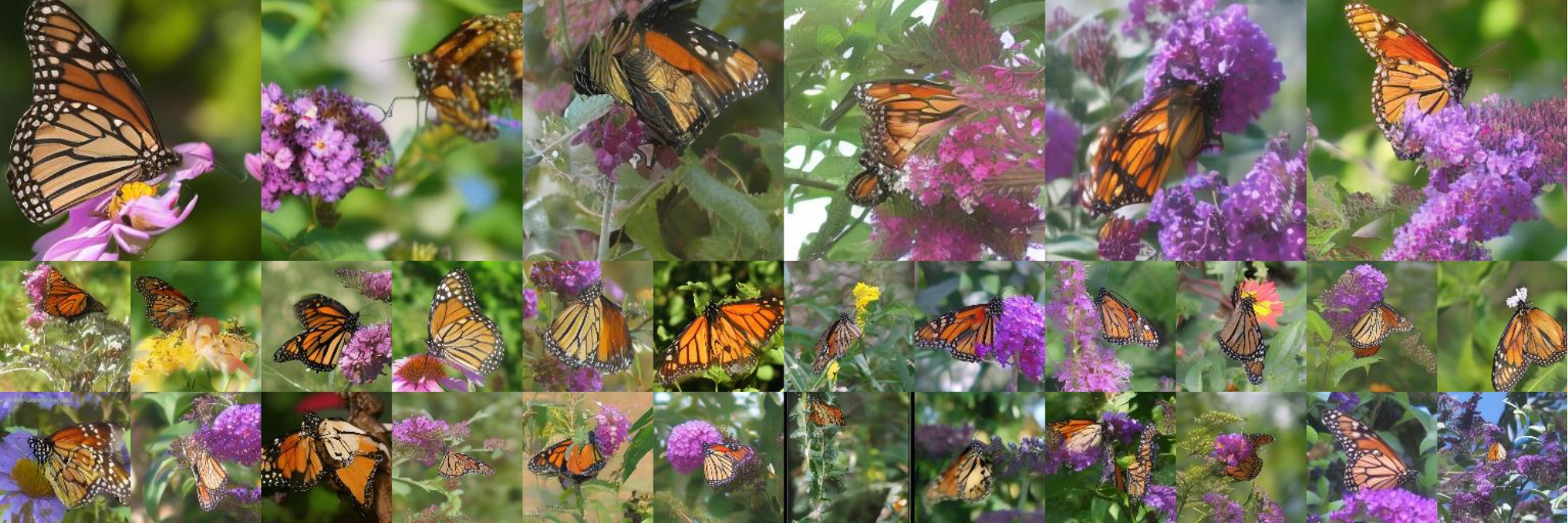}
             \caption{Teacher (25 steps) / FID: 2.89}
         \end{subfigure}
         \begin{subfigure}{\textwidth}
             \centering
             \includegraphics[width=0.88\textwidth]{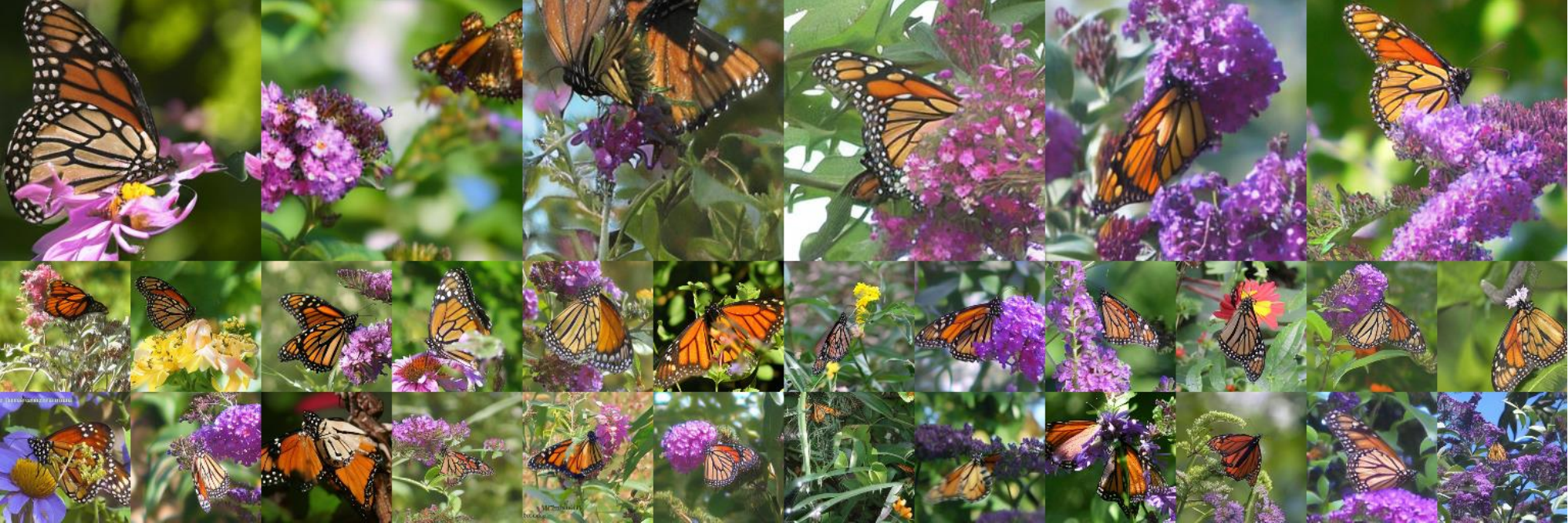}
             \caption{ARD (R+D) / FID: 1.84} 
         \end{subfigure}
    \end{minipage}
    \caption{Randomly generated ImageNet 256p samples for class \textit{monarch butterfly}. All distilled models are 4-step models.}
    \label{fig:img_5}
            \vspace{-0mm}
\end{figure*}

\begin{figure*}[h]
\centering
     \begin{minipage}[h]{\textwidth}
     \centering
         \begin{subfigure}{\textwidth}
             \centering
             \includegraphics[width=0.88\textwidth]{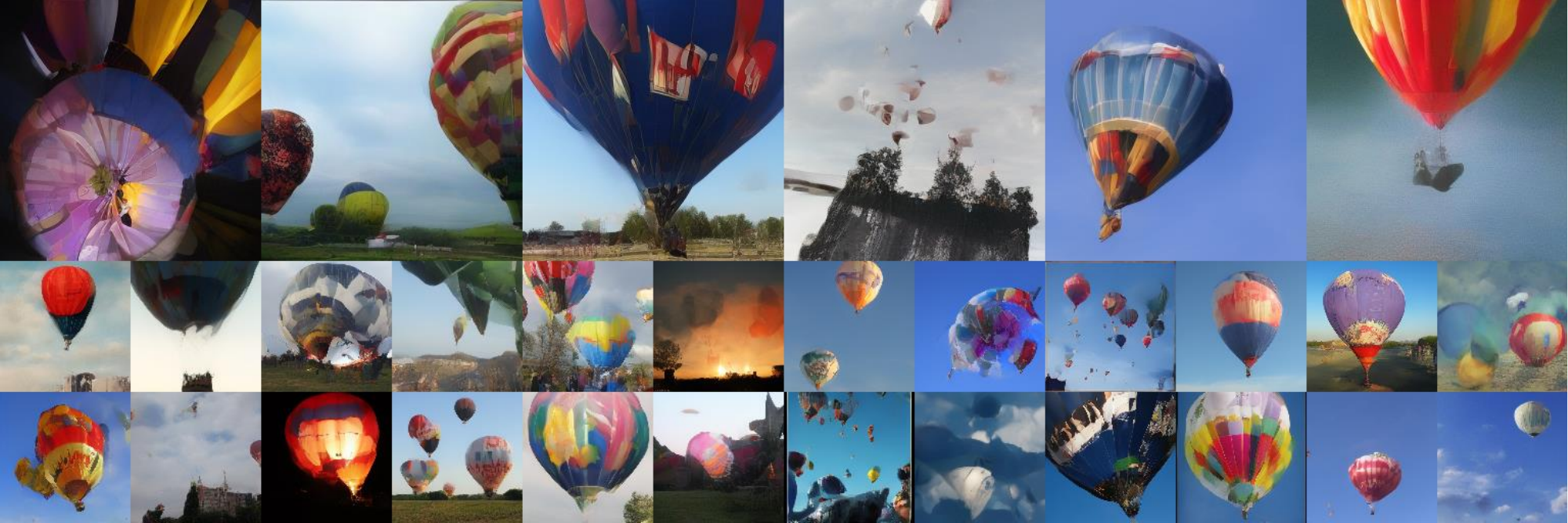}
             \caption{Step Distillation (R) / FID: 10.25}
         \end{subfigure}
         \begin{subfigure}{\textwidth}
             \centering
             \includegraphics[width=0.88\textwidth]{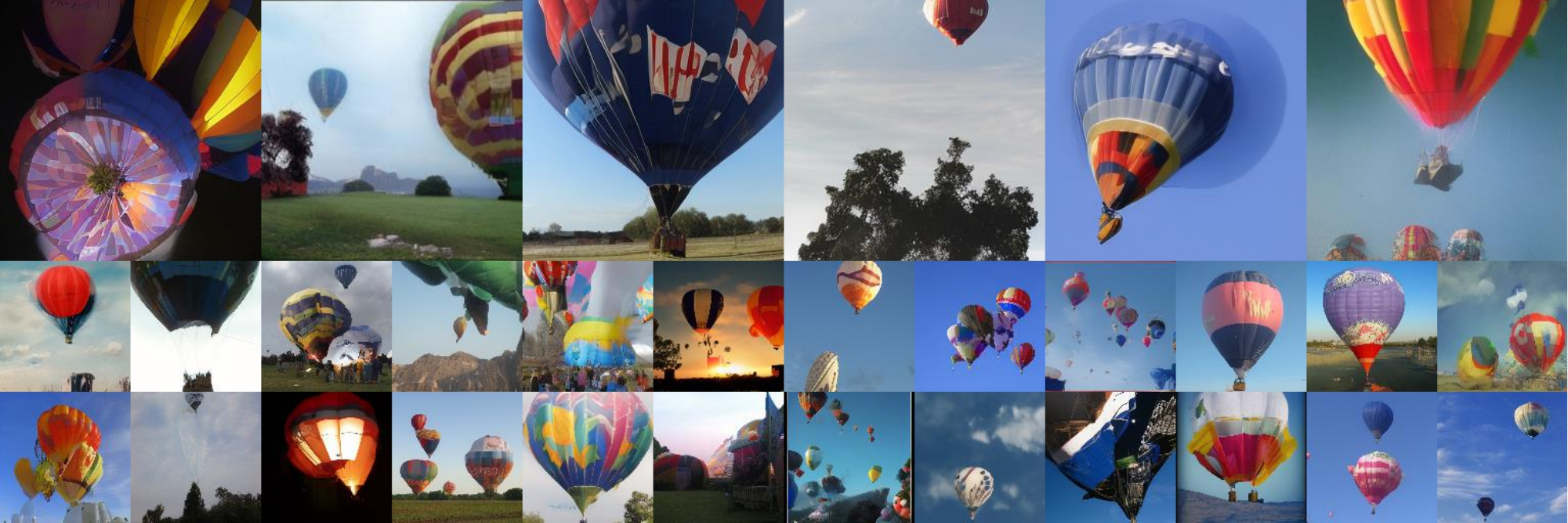}
             \caption{ARD (R) / FID: 4.32} 
         \end{subfigure}
    \end{minipage}
    \begin{minipage}[h]{\textwidth}
        \begin{subfigure}{\textwidth}
             \centering
             \includegraphics[width=0.88\textwidth]{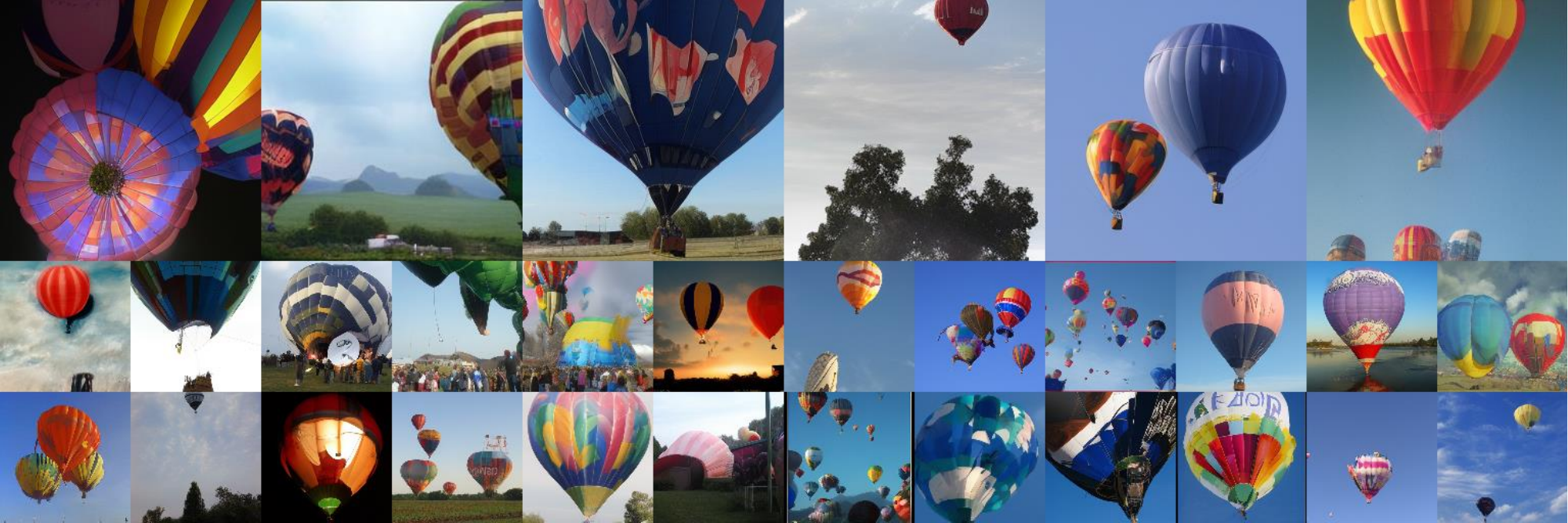}
             \caption{Teacher (25 steps) / FID: 2.89}
         \end{subfigure}
         \begin{subfigure}{\textwidth}
             \centering
             \includegraphics[width=0.88\textwidth]{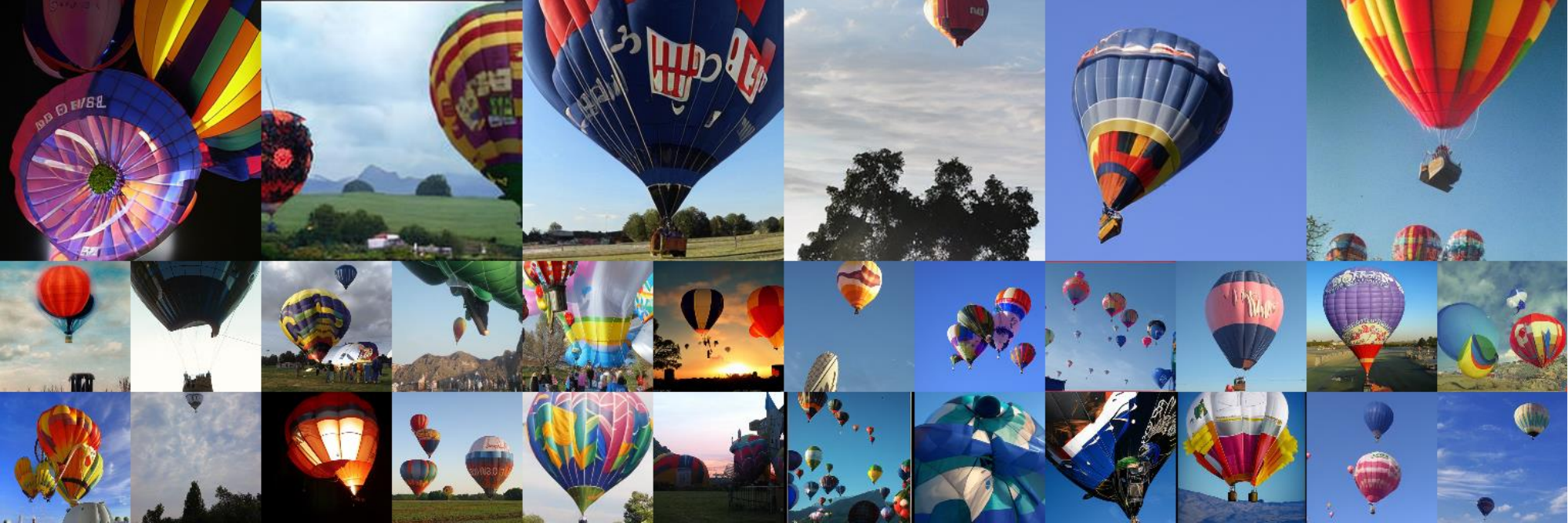}
             \caption{ARD (R+D) / FID: 1.84} 
         \end{subfigure}
    \end{minipage}
    \caption{Randomly generated ImageNet 256p samples for class \textit{balloon}. All distilled models are 4-step models.}
    \label{fig:img_6}
            \vspace{-0mm}
\end{figure*}

\begin{figure*}[h]
\centering
     \begin{minipage}[h]{\textwidth}
     \centering
         \begin{subfigure}{\textwidth}
             \centering
             \includegraphics[width=0.88\textwidth]{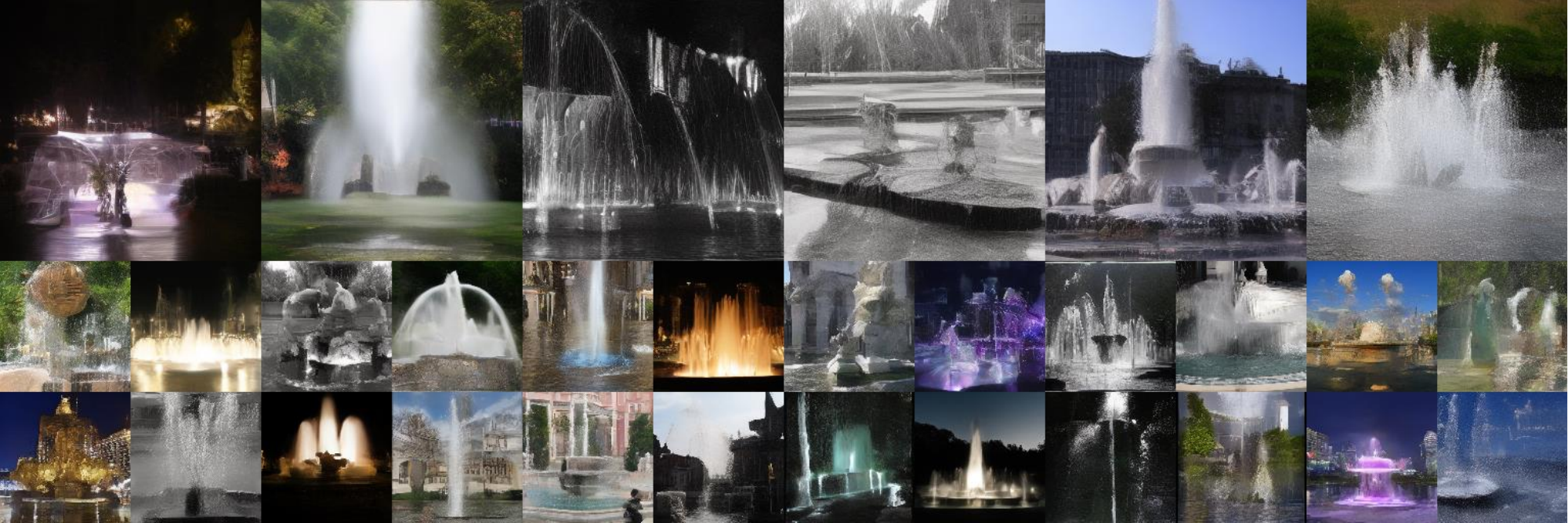}
             \caption{Step Distillation (R) / FID: 10.25}
         \end{subfigure}
         \begin{subfigure}{\textwidth}
             \centering
             \includegraphics[width=0.88\textwidth]{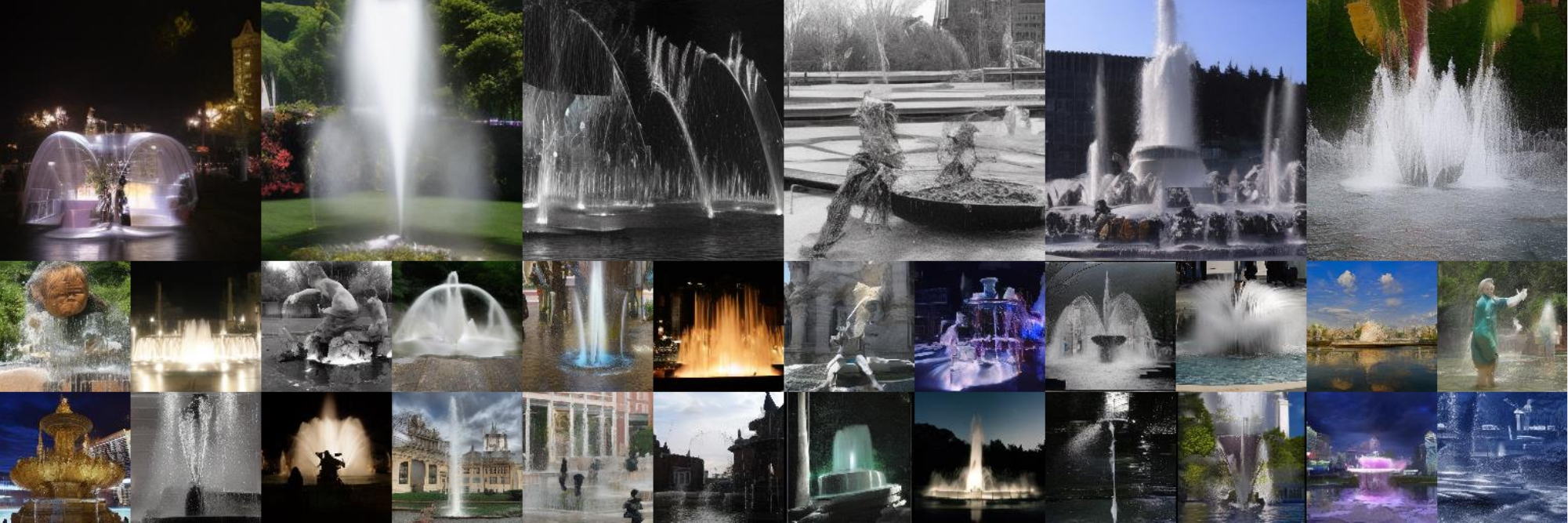}
             \caption{ARD (R) / FID: 4.32} 
         \end{subfigure}
    \end{minipage}
    \begin{minipage}[h]{\textwidth}
        \begin{subfigure}{\textwidth}
             \centering
             \includegraphics[width=0.88\textwidth]{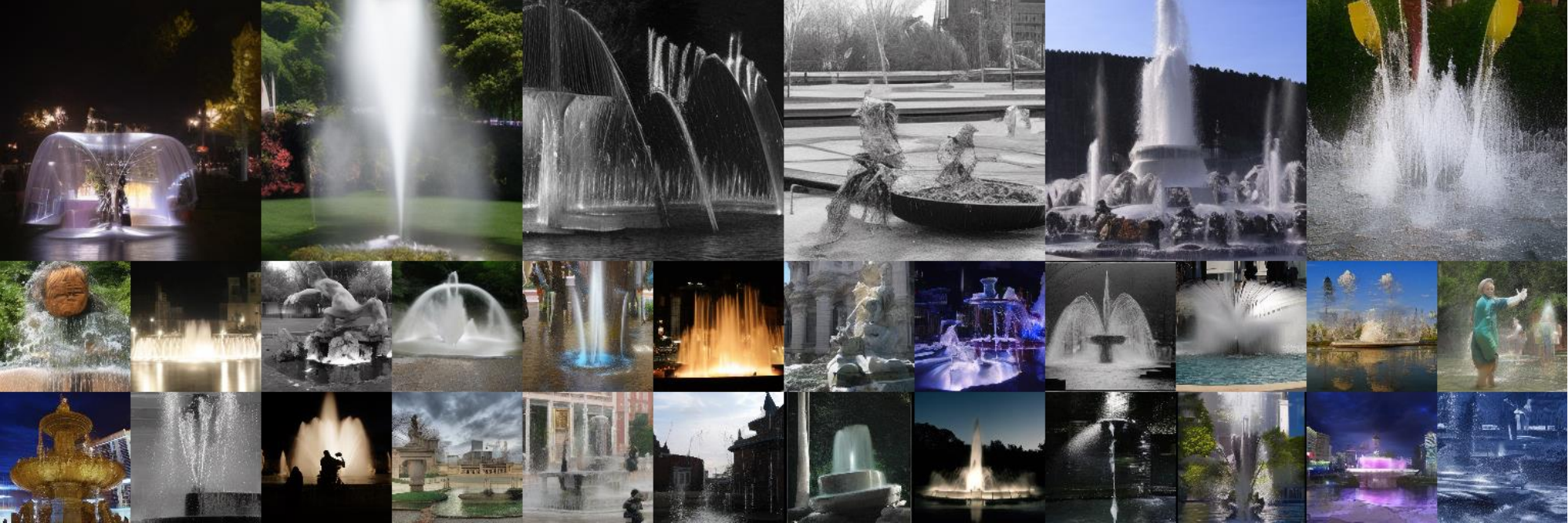}
             \caption{Teacher (25 steps) / FID: 2.89}
         \end{subfigure}
         \begin{subfigure}{\textwidth}
             \centering
             \includegraphics[width=0.88\textwidth]{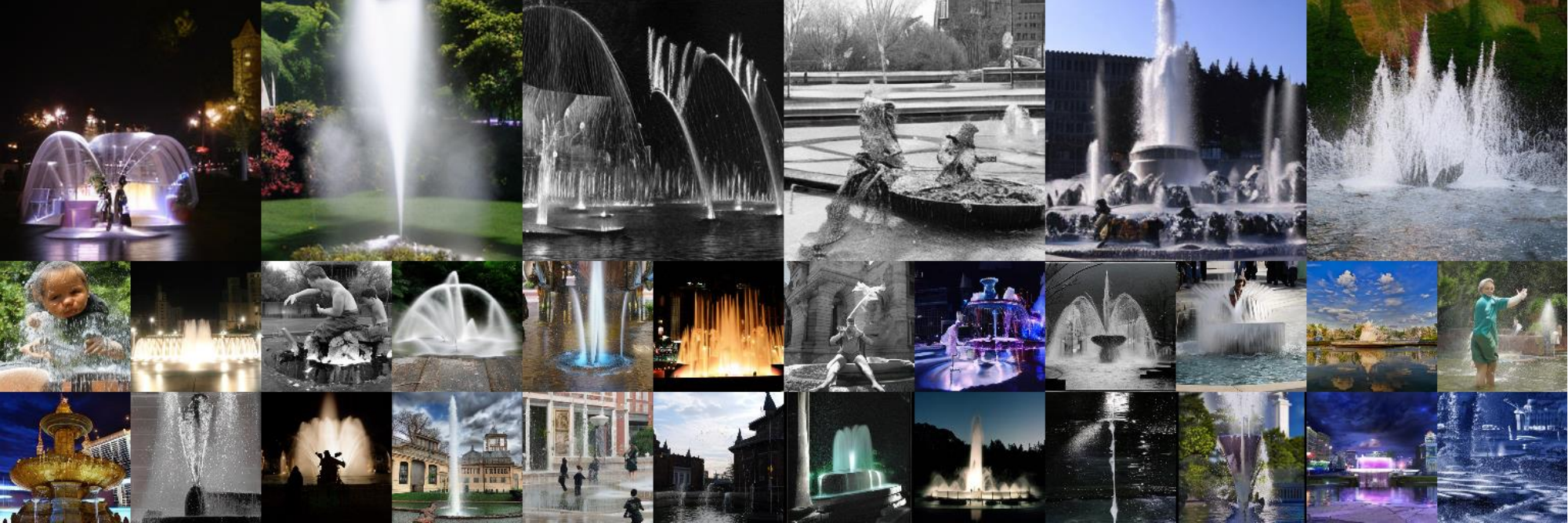}
             \caption{ARD (R+D) / FID: 1.84} 
         \end{subfigure}
    \end{minipage}
    \caption{Randomly generated ImageNet 256p samples for class \textit{fountain}. All distilled models are 4-step models. }
    \label{fig:img_7}
            \vspace{-0mm}
\end{figure*}

\begin{figure*}[h]
\centering
     \begin{minipage}[h]{\textwidth}
     \centering
         \begin{subfigure}{\textwidth}
             \centering
             \includegraphics[width=0.88\textwidth]{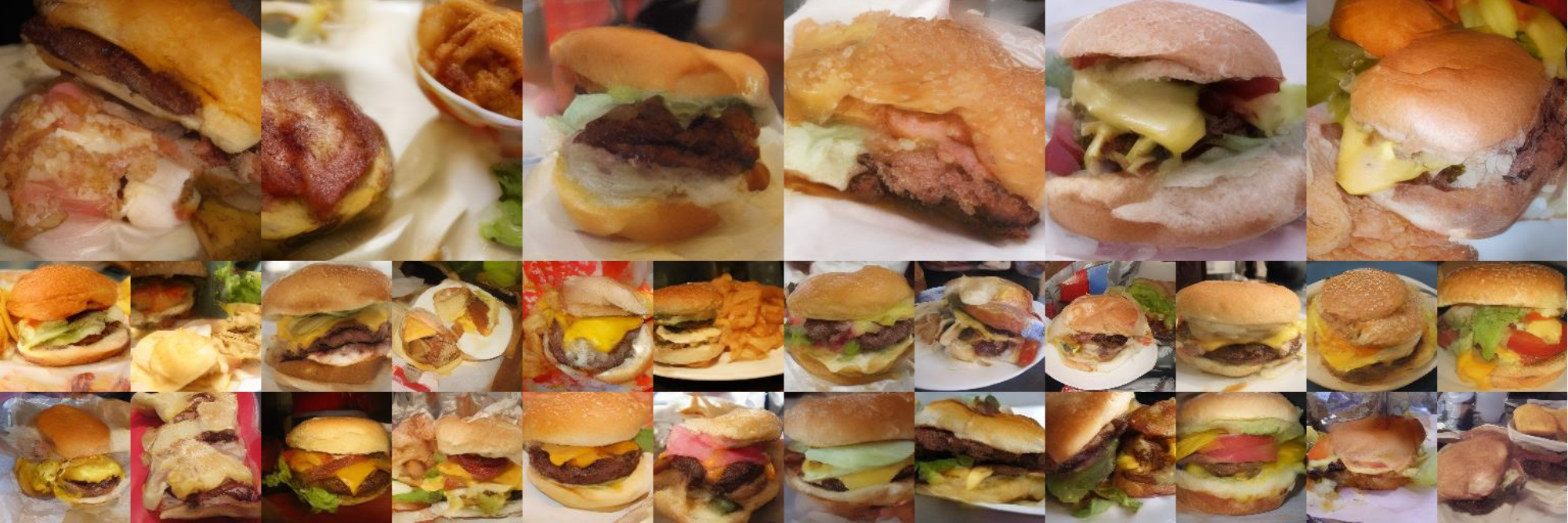}
             \caption{Step Distillation (R) / FID: 10.25}
         \end{subfigure}
         \begin{subfigure}{\textwidth}
             \centering
             \includegraphics[width=0.88\textwidth]{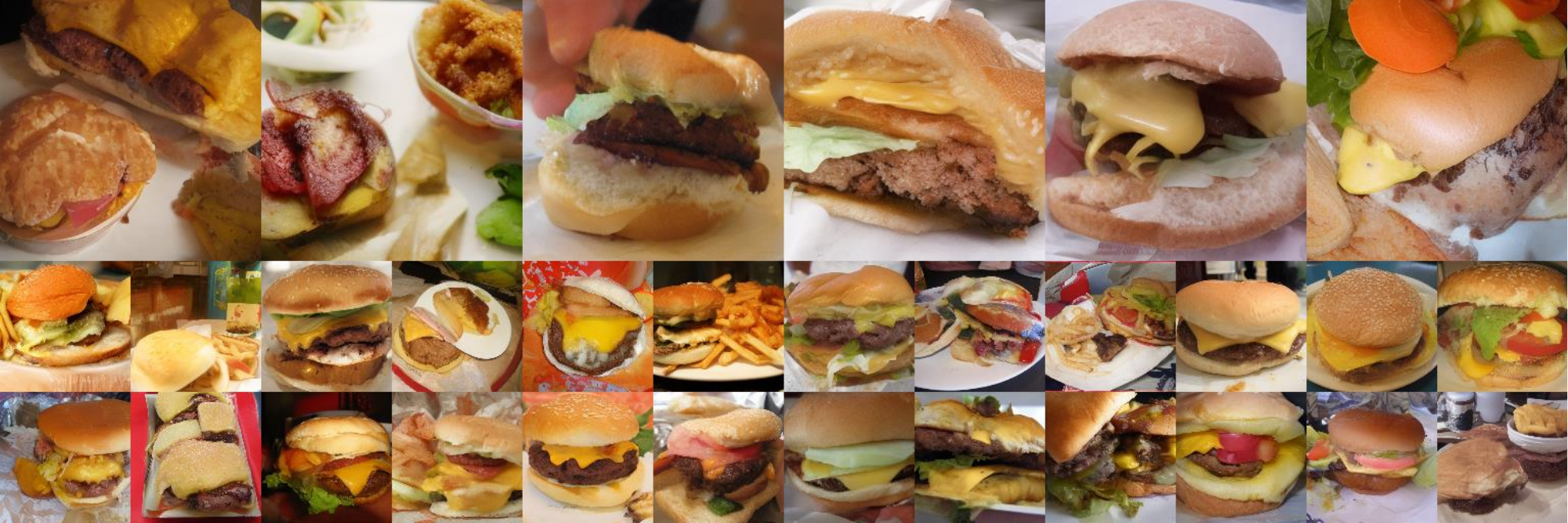}
             \caption{ARD (R) / FID: 4.32} 
         \end{subfigure}
    \end{minipage}
    \begin{minipage}[h]{\textwidth}
        \begin{subfigure}{\textwidth}
             \centering
             \includegraphics[width=0.88\textwidth]{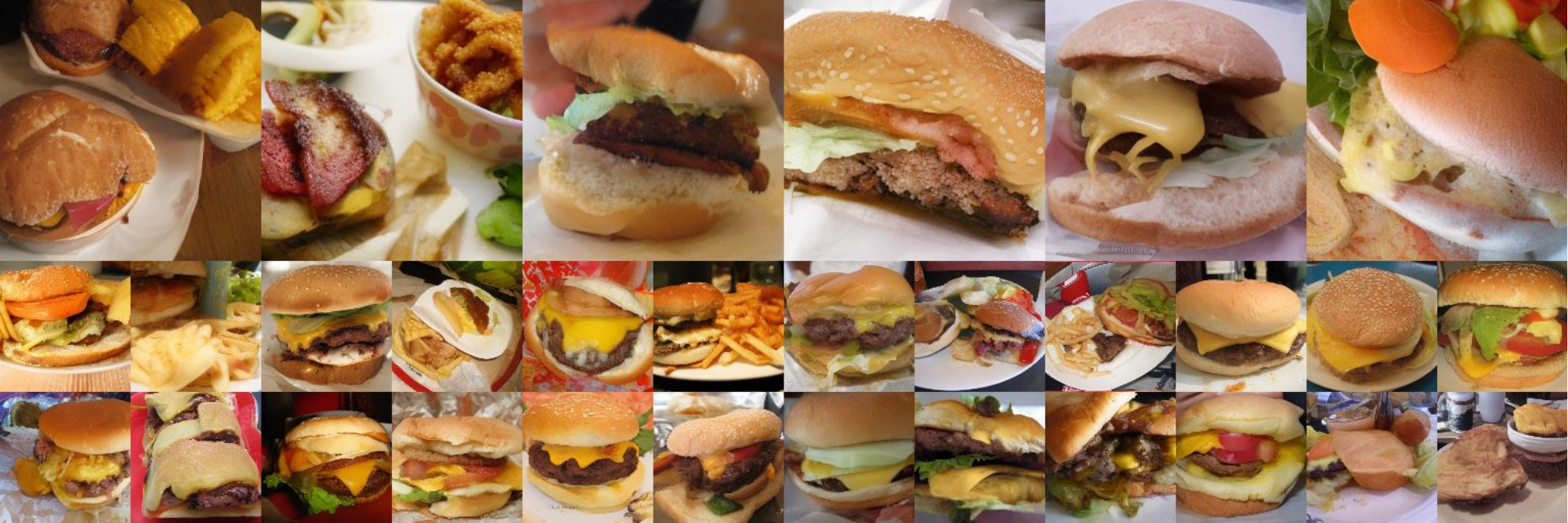}
             \caption{Teacher (25 steps) / FID: 2.89}
         \end{subfigure}
         \begin{subfigure}{\textwidth}
             \centering
             \includegraphics[width=0.88\textwidth]{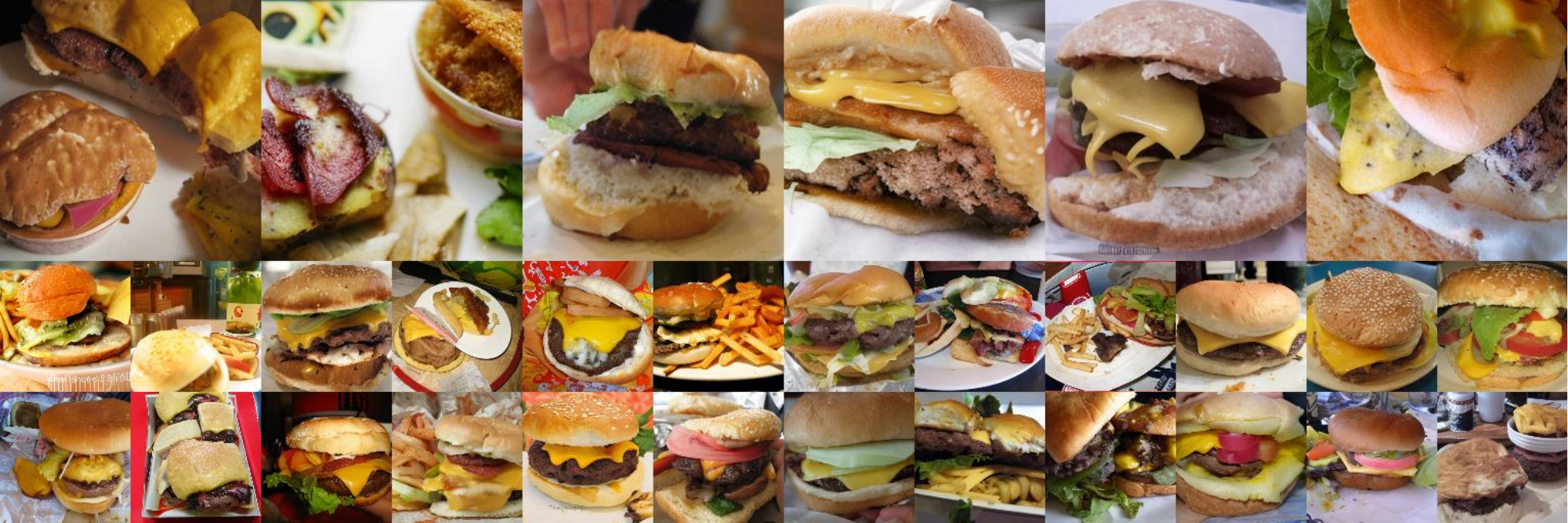}
             \caption{ARD (R+D) / FID: 1.84} 
         \end{subfigure}
    \end{minipage}
    \caption{Randomly generated ImageNet 256p samples for class \textit{cheeseburger}. All distilled models are 4-step models.}
    \label{fig:img_8}
            \vspace{-0mm}
\end{figure*}

This section presents additional generated samples. \Cref{fig:various_rand} illustrates the text-conditional samples with various initial noise $\rvx_{\tau_{S}}$. \Cref{fig:more_compare} shows the comparison between ARD and the Emu teacher with the same number of sampling steps. \Cref{fig:img_1,fig:img_2,fig:img_3,fig:img_4,fig:img_5,fig:img_6,fig:img_7,fig:img_8} show the class-conditioned generations across various classes.




\end{document}